\documentclass[a4paper,12pt]{article}
\pdfoutput=1
\usepackage{amsmath}
\usepackage{url}
\usepackage{amssymb}
\usepackage{graphicx}
\usepackage{setspace, enumitem,titlesec}
\usepackage{calc}
\usepackage{mathtools}
\usepackage{mathrsfs }

\usepackage{algorithm}
\usepackage{algorithmic}
\usepackage{fancybox}
\usepackage{array, makecell} %
\usepackage{amsmath}
\usepackage{amssymb}
\usepackage{amsthm}
\usepackage{cite}
\usepackage{authblk}
\usepackage{multirow}% http://ctan.org/pkg/multirow
\usepackage{hhline}% http://ctan.org/pkg/hhline

\renewenvironment{abstract}
 {\small
  \begin{center}
  \bfseries \abstractname\vspace{-.5em}\vspace{0pt}
  \end{center}
  \list{}{
    \setlength{\leftmargin}{.5cm}%
    \setlength{\rightmargin}{\leftmargin}%
  }%
  \item\relax}
 {\endlist}

\title {A survey of Object Classification and Detection based on 2D/3D data }
\author{Xiaoke Shen}
\affil{The Graduate Center, City University of New York}
\date{}
\begin{document}
\maketitle
\begin{abstract}
Recently, by using deep neural network based algorithms, object classification, detection and semantic segmentation solutions are significantly improved. However, one challenge for 2D image-based systems is that they cannot provide accurate 3D location information. This is critical for location sensitive applications such as autonomous driving and robot navigation. On the other hand, 3D methods, such as RGB-D and RGB-LiDAR based systems, can provide solutions that significantly improve the RGB only approaches. That is why this is an interesting research area for both industry and academia.\\

Compared with 2D image-based systems, 3D-based systems are more complicated due to the following five reasons:  1) Data representation itself is more complicated. 3D images can be represented by point clouds, meshes, volumes. 2D images have pixel grid representations. 2) The computation and memory resource requirement is higher as an extra dimension is added. 3) Different distribution of the objects and difference in scene areas between indoor and outdoor make one unified framework hard to achieve. 4) 3D data, especially for the outdoor scenario, is sparse compared with the dense 2D images which makes the detection task more challenging. Finally, large size labelled datasets, which are extremely important for supervised based algorithms, are still under construction compared with well-built 2D datasets such as ImageNet.\\

Based on those challenges listed above, the described systems are organized by application scenarios, data representation methods and main tasks addressed. At the same time, critical 2D based systems which greatly influence the 3D ones are also introduced to show the connection between them.\\
\end{abstract}

\section{Introduction}

Currently, great achievements have been shown in 2D-based systems by using deep neural networks. Actually, the neural network is not a new construct. It was first introduced in 1950s. In 1958, Rosenblatt\cite{Rosenblatt} created the perceptron, an algorithm for pattern classification.  The perceptron algorithm's first implementation, in custom hardware, was one of the first artificial neural networks to be produced. Although the perceptron initially seemed promising, Neural network research stagnated after machine learning research by Minsky and Papert (1969)\cite{Minsky}, who discovered two key issues with the computational machines that processed neural networks. The first was that basic perceptrons were incapable of processing the exclusive-or circuit. The second was that computers did not have enough processing power to effectively handle the work required by large neural networks
\cite{ann}. The first issue was solved by introducing more layers of networks and the second issue by both reducing the complexity of the algorithms and introducing more powerful computing hardware such as GPUs.\\

By using deep neural network based on algorithms, especially convolutional neural networks based algorithms, computer vision systems based on 2D images have been making great achievements in image classification, object detection and semantic segmentation since the year 2012. For some scenarios, deep neural network based algorithms can achieve a similar or even better performance on 2D image classification/detection and semantic segmentation than the human expert, but 3D-based systems are not as developed yet. \\

The survey is organized as follows: in the second section, 2D-based systems are introduced and in the third section the 3D-based systems are introduced.\\

\section{2D-image based systems}
2D-image based systems are introduced in this section based on the high-level tasks addressed.\\
\subsection{High-level main tasks}

  \begin{figure}[H]
  \begin{center}
  %0.43
      \includegraphics[width=1.0\linewidth]{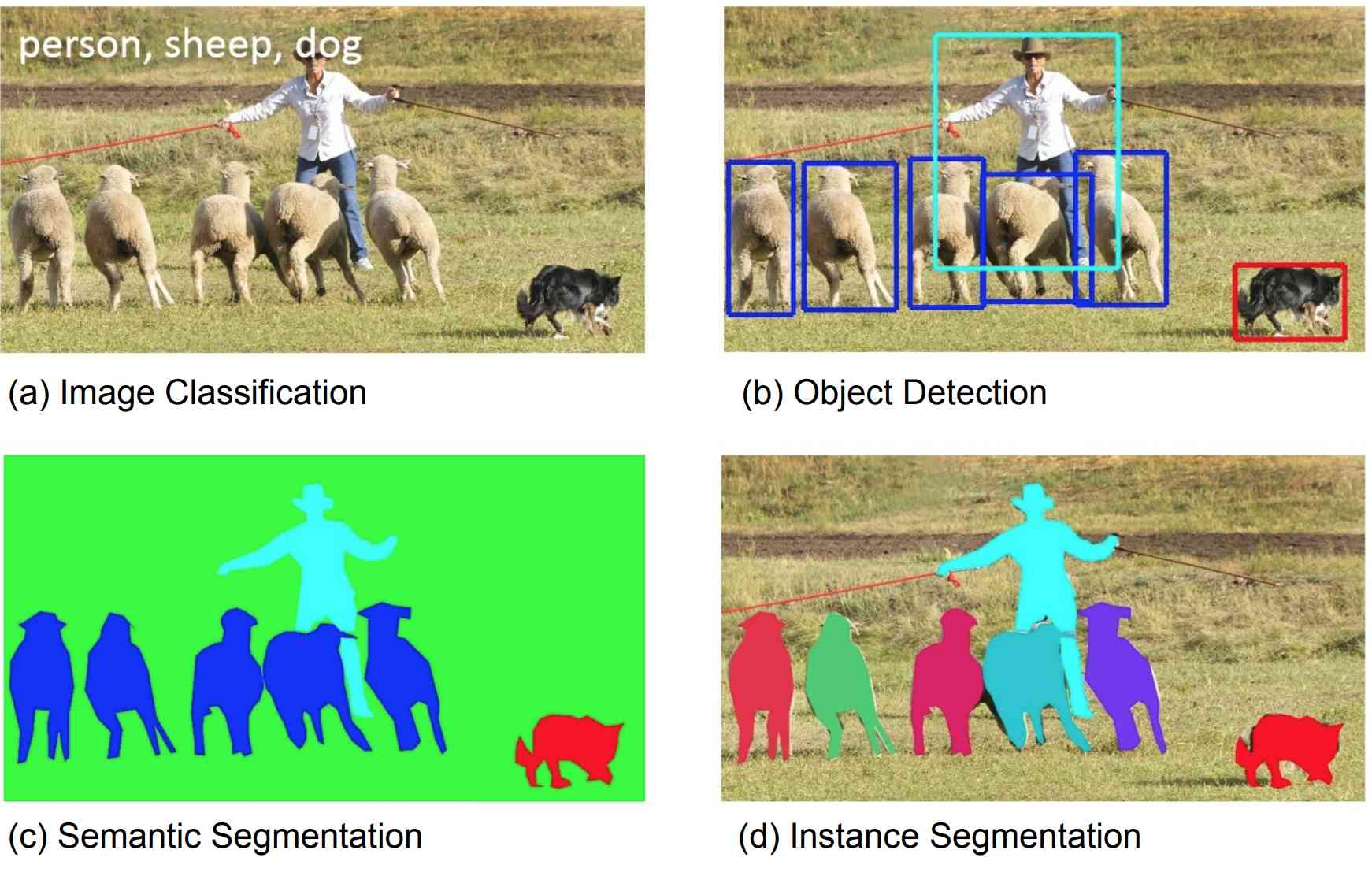}
\end{center}
\caption{One example of different 2D computer vision tasks: a) Image classification b) Object Detection c) Semantic Segmentation d) Instance Segmentation. Figure is adjusted from \cite{DBLP:journals/corr/LinMBHPRDZ14}.}
 \label{fig:differnence_object_detection_segmentation}
 \end{figure}

Figure \ref{fig:differnence_object_detection_segmentation} from \cite{DBLP:journals/corr/LinMBHPRDZ14} demonstrates the tasks of Image classification, Object Detection, Semantic Segmentation and Instance Segmentation. Image classification is recognizing the interesting objects in each image and output the objects categories. Object Detection will not only output the objects categories but also the location of those objects with bounding boxes. Both semantic segmentation and instance segmentation will output the pixel level location of each object. The difference between those two tasks is that semantic segmentation does not distinguish the instance in the same category while the instance segmentation does.\\
\subsection{Main Networks used for Image Classification }
\begin{figure}[H]
  \begin{center}
  %0.8
      \includegraphics[width=0.85\linewidth]{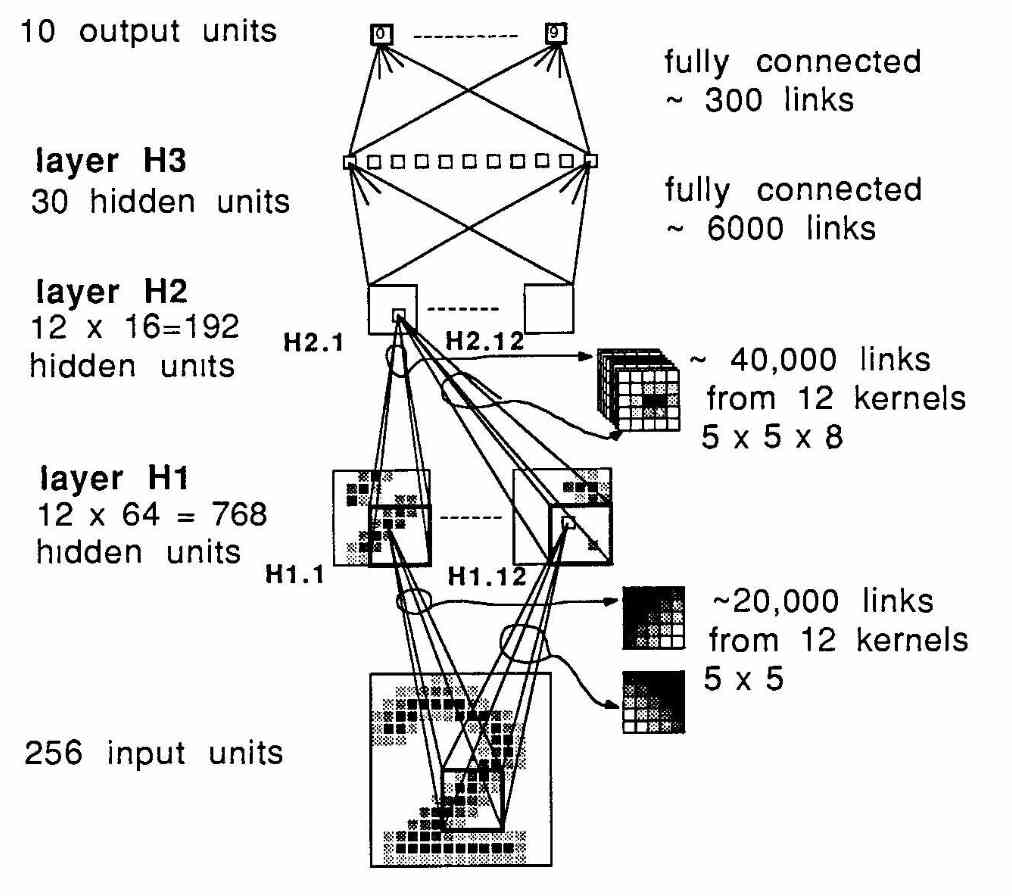}
\end{center}
\caption{An early Neural Network used in \cite{doi:10.1162/neco.1989.1.4.541}. }
 \label{fig:bpzip}
 \end{figure}
Some traditional algorithms used for the image classification are nearest neighbor and SVM. The features are the flattened pixel values. In the year 1989, the first important application\cite{doi:10.1162/neco.1989.1.4.541} of using BP(Back Propagation) appeared. From this paper, a basic structure is shown in Figure \ref{fig:bpzip}. Similar structures are used in the modern neural networks such as AlexNet\cite{NIPS2012_4824}, VGG 16  \cite{SimonyanZ14a} and ResNet \cite{DBLP:journals/corr/HeZRS15}. The basic idea of the CNN was also introduced in \cite{doi:10.1162/neco.1989.1.4.541}.

One important reason of the success of deep learning algorithms in the computer vision area is the invention of CNN. Another important reason is the availabIlity of large labelled datasets. As we know, in the machine learning research area, two kinds of learning approaches can be done: supervised learning and unsupervised learning. For the supervised learning algorithms, the labeled data is required to train the algorithm. So the availability of the labeled data is very important to the development of the supervised learning based algorithms. The ImageNet dataset \cite{imagenet_cvpr09} provides 1.2 million high-resolution labeled images of 1000 categories. This dataset became one of the most important datasets related to the object classification.\\

 \begin{figure}[H]
  \begin{center}
  %0.55
      \includegraphics[width=1.0\linewidth]{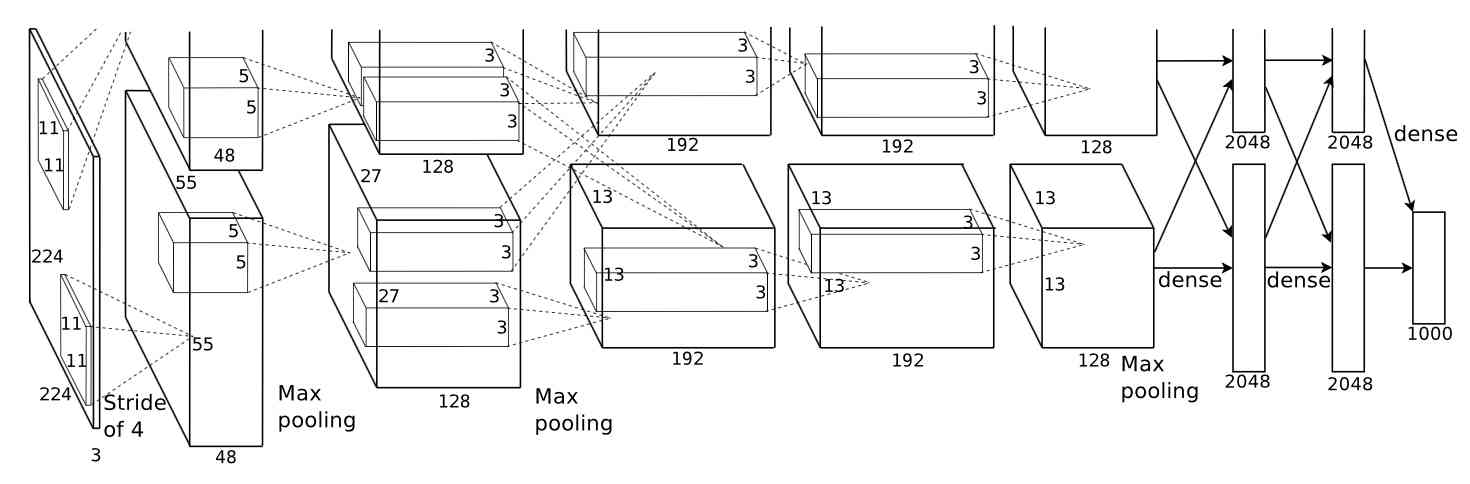}
\end{center}
\caption{An illustration of the architecture of AlexNet\cite{NIPS2012_4824}. Figure is from\cite{NIPS2012_4824}. }
 \label{fig:alexnet}
 \end{figure}
In the ILSVRC-2012 competition, the method of \cite{NIPS2012_4824} achieved a top-5 test error rate of $15.3\%$, compared to $26.2\%$ achieved by the second-best entry. The outstanding performance of deep neural network used in this paper brought the focus back to the neural network research again. the CNN network used in  \cite{NIPS2012_4824}  is  shown in Figure \ref{fig:alexnet}. Two GPUs were used to speed up the calculation. Dropout \cite{DBLP:journals/corr/abs-1207-0580} was used here and was proved to be effective to reduce the overfitting problem.  This structure is called AlexNet to emphasize the unique contribution of the author of this paper. Negatives of this network are it is a bit complicated and the structure is not so elegant.  This was addressed by the future works.\\

  \begin{figure}[H]
  \begin{center}
  %0.75
      \includegraphics[width=0.65\linewidth]{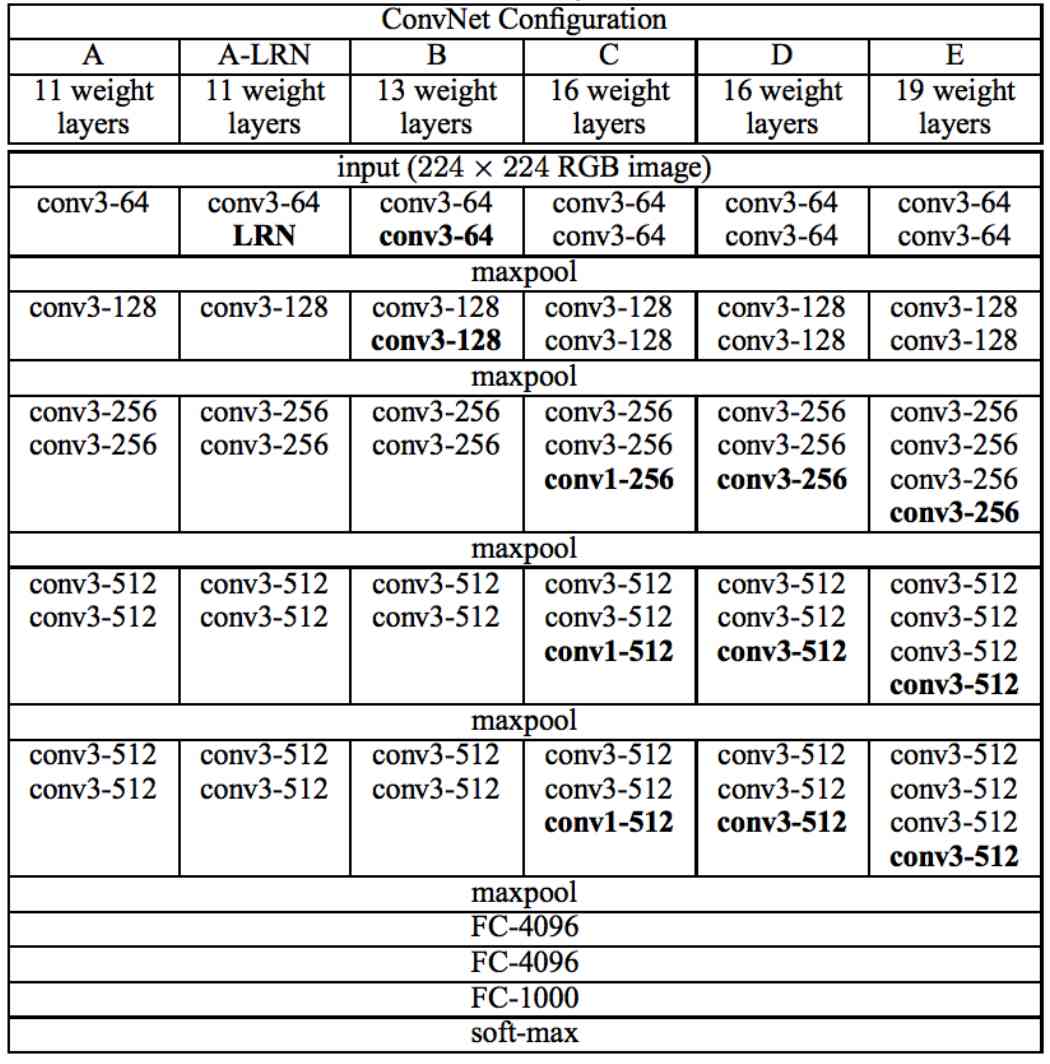}
\end{center}
\caption{An illustration of the VGG network structure. Figure is from \cite{NIPS2012_4824}.}
 \label{fig:vgg16}
 \end{figure}

Two years after the AlexNet, a new network called VGG was proposed in \cite{NIPS2012_4824}. The structure of this network is very tidy and elegant. In contrast to the AlexNet\cite{NIPS2012_4824} that used different size of the convolutional kernels, in the VGG CNN network\cite{SimonyanZ14a} , only  3 by 3 kernels were used for the whole network. At the same time, the trained network weights based on the ImageNet\cite{imagenet_cvpr09} dataset were shared with the public. The negative of this model is that there are too many parameters and thus the computation is slow. \\
%This elegant design can also be used in the 3D CNN network to address the 3D object classification problem which will be shown later.\\

 \begin{figure}[H]
  \begin{center}
  %0.5
      \includegraphics[width=0.75\linewidth]{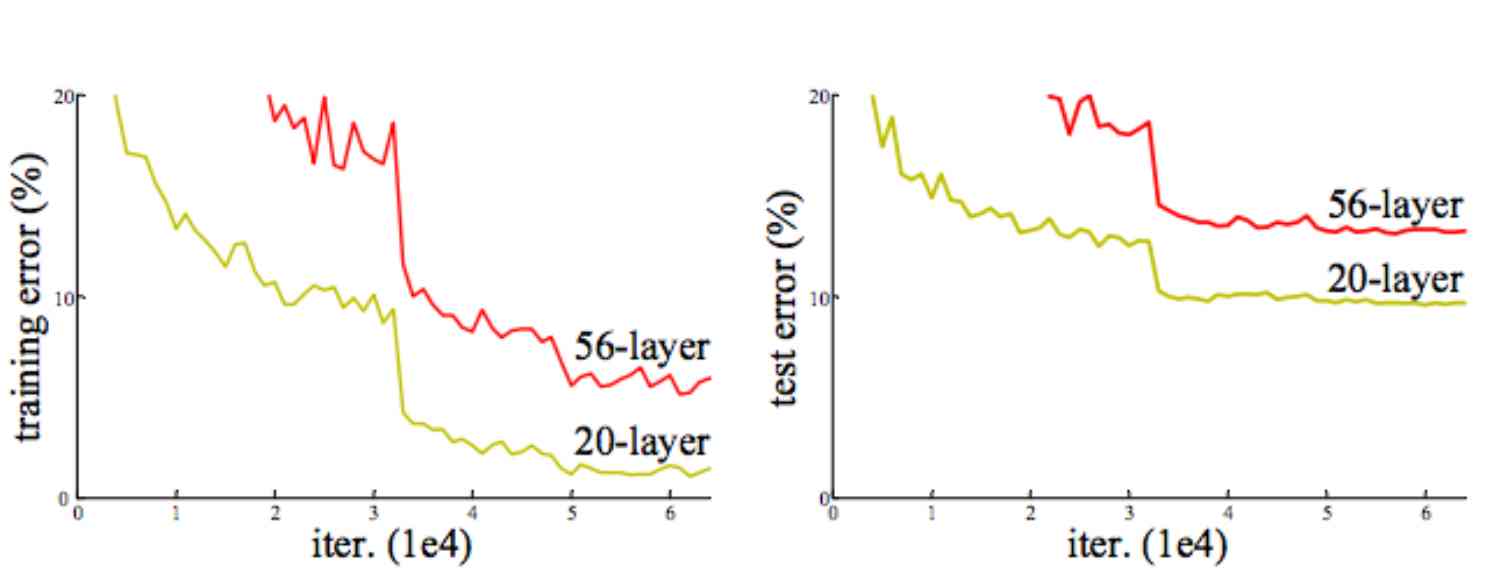}
\end{center}
\caption{An illustration of deep neural networks fails by using the traditional network structure. Figure is from \cite{DBLP:journals/corr/HeZRS15}.}
 \label{fig:ResNet}
 \end{figure}

%\subsection{Deep Residual Learning for Image Classification\cite{DBLP:journals/corr/HeZRS15}}
In the year of 2015, a new structure called ResNet\cite{DBLP:journals/corr/HeZRS15} was proposed and this new structure increased the layers of the network to more than 100 with less parameters than the VGG16 model and better performance. The motivation of this work was to try to find a deeper neural network structure to achieve better performance as the deepest network prior this work was around 20 to 30 layers. The authors had a basic assumption that if more layers are added, the performance should be at least the same as the smaller networks. However, the traditional network's performance decreases when the layers increase\cite{DBLP:journals/corr/HeZRS15} as shown in Figure \ref{fig:ResNet}. In order to address this problem, the authors of \cite{DBLP:journals/corr/HeZRS15} designed a deep residual network by adding a short cut between every other layer and finally show that this network can achieve a better performance. In\cite{DBLP:journals/corr/HeZRS15}, layers of the neural network goes up to 101 layer and it has a better performance than the previous state-of-the-art neural network such as VGG16\cite{SimonyanZ14a}, which only has 16 layers. However, as mentioned in \cite{DBLP:journals/corr/HeZRS15}, the parameters used for the 101-layer ResNet are even less than the 16-layer VGG16 network. This seems amazing. Indeed, the main contribution of the reduction of the parameters was using the convolutional network layer instead of the fully connected layer.\\

\begin{figure}[H]
  \begin{center}
  %0.6
      \includegraphics[width=1.0\linewidth]{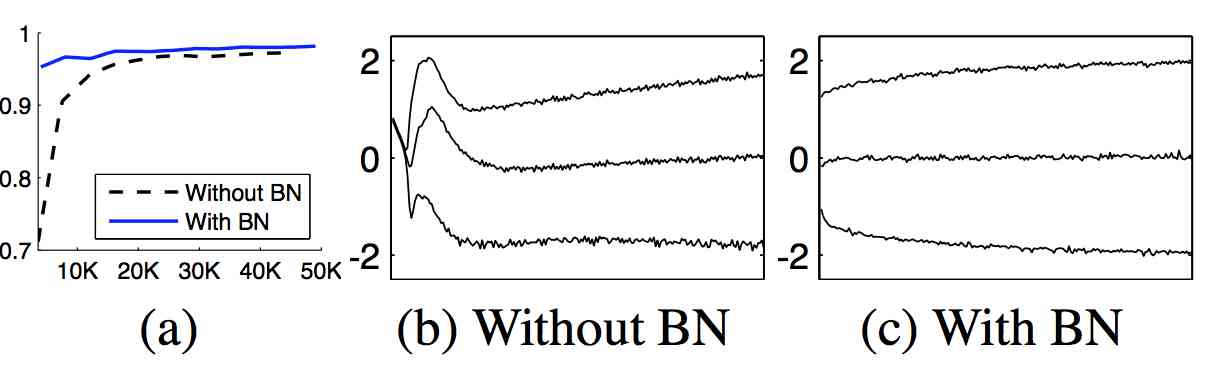}
\end{center}
\caption{(a) The test accuracy of the MNIST network trained with and without Batch Normalization, vs. the number of training steps. Batch Normalization helps the network train faster and achieve higher accuracy\cite{DBLP:journals/corr/IoffeS15}. (b, c) The evolution of input distributions to a typical sigmoid, over the course of training, shown as  15, 50, 85th percentiles. Batch Normalization makes the distribution more stable and reduces the internal covariate shift\cite{DBLP:journals/corr/IoffeS15}. Figure and Caption are from original paper.}
 \label{fig:batch_norm}
 \end{figure}

Alongside of the changing of the structure of deep neural networks\cite{DBLP:journals/corr/HeZRS15}, another algorithm called BN(Batch Normalization)\cite{DBLP:journals/corr/IoffeS15} was proposed to speedup the convergence rate of the training process. The main contribution of this paper as shown in Figure \ref{fig:batch_norm} is that it greatly reduced the convergence time for the training process. This is why BN became one of the standard training steps for deep neural networks.\\

There are lots of other new deep neural network models introduced to the community in the past few years. In \cite{DBLP:journals/corr/CanzianiPC16}, an analysis of deep neural network models are given based on the ImageNet classification challenge\cite{NIPS2012_4824}. The network modes analyzed in \cite{DBLP:journals/corr/CanzianiPC16} include: AlexNet
\cite{NIPS2012_4824} , batch normalised AlexNet\cite{bnalex}, batch normalised Network
In Network (NIN) \cite{DBLP:journals/corr/LinCY13}, ENet \cite{DBLP:journals/corr/PaszkeCKC16},
GoogLeNet \cite{DBLP:journals/corr/SzegedyLJSRAEVR14}, VGG-16 and -19 \cite{SimonyanZ14a}, ResNet-18,
-34, -50, -101 and -152  \cite{DBLP:journals/corr/HeZRS15}, Inception-v3 \cite{DBLP:journals/corr/SzegedyVISW15} and Inception-v4
\cite{DBLP:journals/corr/SzegedyIV16} .\\

\begin{figure}[H]
  \begin{center}
  %0.75
      \includegraphics[width=1.0\linewidth]{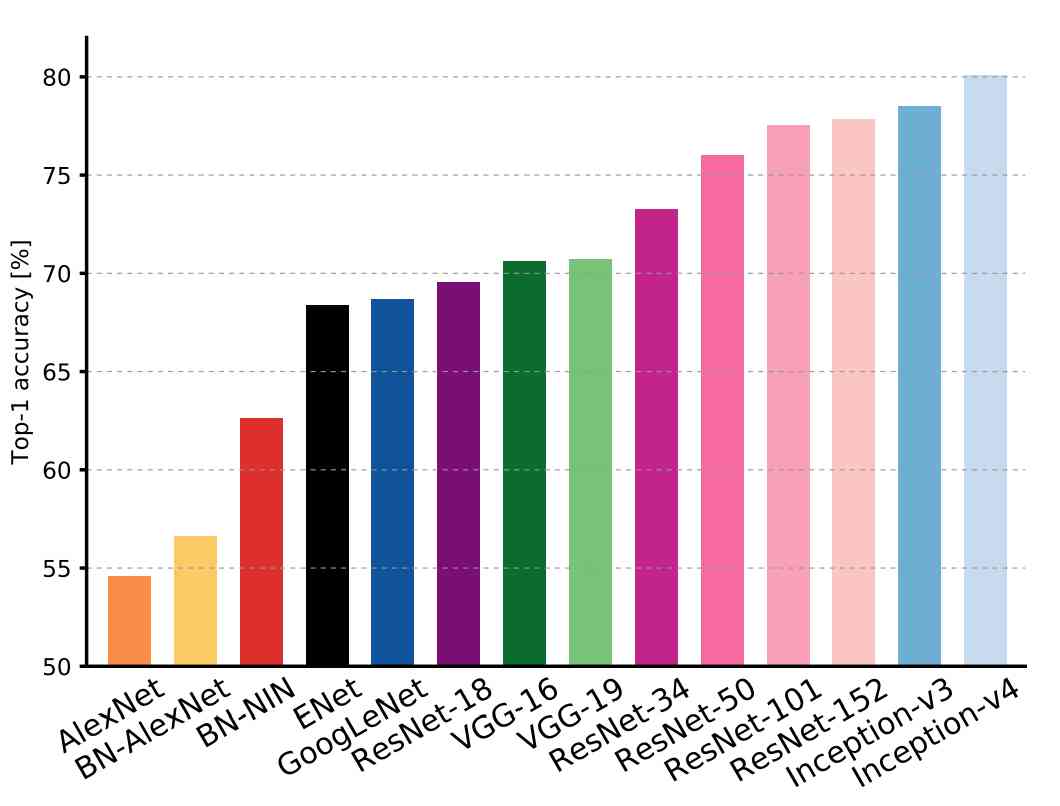}
\end{center}
\caption{\textbf{Top1 vs. network}. Top-1 validation
accuracies for top scoring single-model architectures. Notice that
networks of the same group share the same hue, for
example ResNet are all variations of pink.Figure and Caption are adjusted from \cite{DBLP:journals/corr/CanzianiPC16}}
 \label{fig:network1}
 \end{figure}

\begin{figure}[H]
  \begin{center}
  %0.75
      \includegraphics[width=1.0\linewidth]{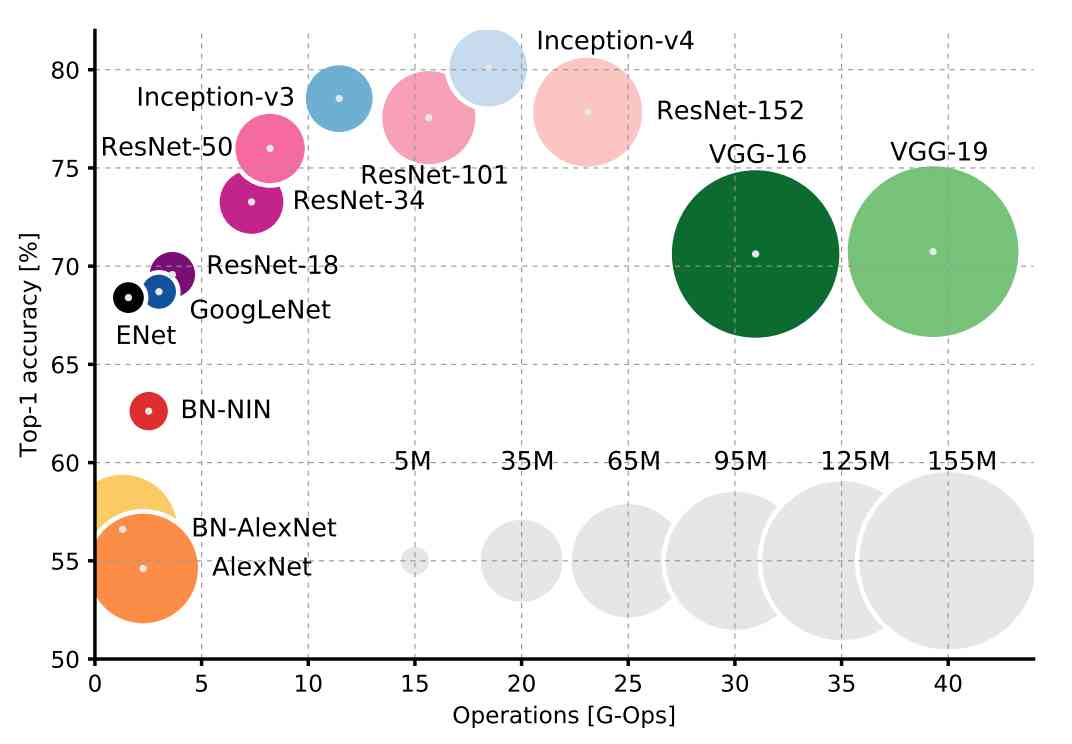}
\end{center}
\caption{\textbf{Top1 vs. operations, size $\propto$ parameters}.
Top-1 one-crop accuracy versus amount of operations
required for a single forward pass. The size of the
blobs is proportional to the number of network parameters;
a legend is reported in the bottom right corner,
spanning from 5×106
to 155×106
params. Both
these figures share the same y-axis, and the grey dots
highlight the centre of the blobs. Figure and Caption are from \cite{DBLP:journals/corr/CanzianiPC16}}
 \label{fig:network2}
 \end{figure}

\begin{figure}[H]
  \begin{center}
  %0.35
      \includegraphics[width=1.0\linewidth]{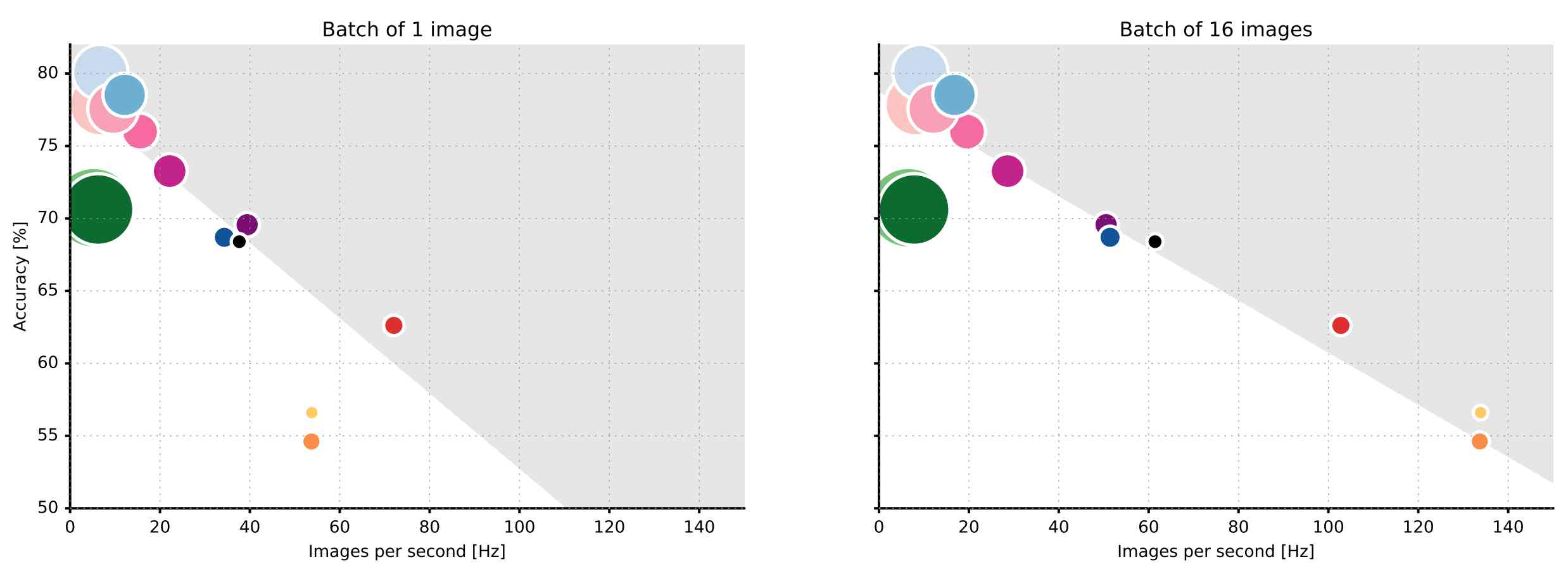}
\end{center}
\caption{\textbf{Accuracy vs. inferences per second, size  $\propto$ operations}. Non trivial linear upper bound is shown
in these scatter plots, illustrating the relationship between prediction accuracy and throughput of all examined
architectures. These are the first charts in which the area of the blobs is proportional to the amount of operations,
instead of the parameters count. We can notice that larger blobs are concentrated on the left side of the charts,
in correspondence of low throughput, i.e. longer inference times. Most of the architectures lay on the linear
interface between the grey and white areas. If a network falls in the shaded area, it means it achieves exceptional
accuracy or inference speed. The white area indicates a suboptimal region. E.g. both AlexNet architectures
improve processing speed as larger batches are adopted, gaining 80 Hz. Figure and Caption are from \cite{DBLP:journals/corr/CanzianiPC16}.}
 \label{fig:network3}
 \end{figure}

Top-1 validation accuracies\footnote{For top-1 validation, if the top predicted class is the same as the target label, then the prediction is correct. In the case of top-5 validation, if the target label is one of the top 5 predictions, then the prediction is correct. In most work for the ImageNet classification challenge\cite{NIPS2012_4824}, the top-5 validation accuracies were provided. In \cite{DBLP:journals/corr/CanzianiPC16}, the comparison is done based on top-1 validation accuracies for all works to make sure it is fair.} for top scoring single-model architectures is shown in Figure \ref{fig:network1}. Top1 validation accuracies vs. operations and also size of the parameters is shown in Figure \ref{fig:network2}. The accuracy vs. inferences per second and the parameter size are shown in Figure \ref{fig:network3}. From those visualization, we can have a clear picture about the parameter size and the performance from both the accuracy and inference time's perspective. Meanwhile, we can see the number of operations is a reliable estimate of the inference time \cite{DBLP:journals/corr/CanzianiPC16} which is consistent with the intuition.\\
%In this section, several main deep neural network models will be described. The analysis here will try to cover all deep neural network models which inspired or be used directly in those 3D deep learning algorithms introduced in the future section of this article.

\begin{figure}[H]
  \begin{center}
  %0.55
      \includegraphics[width=1.0\linewidth]{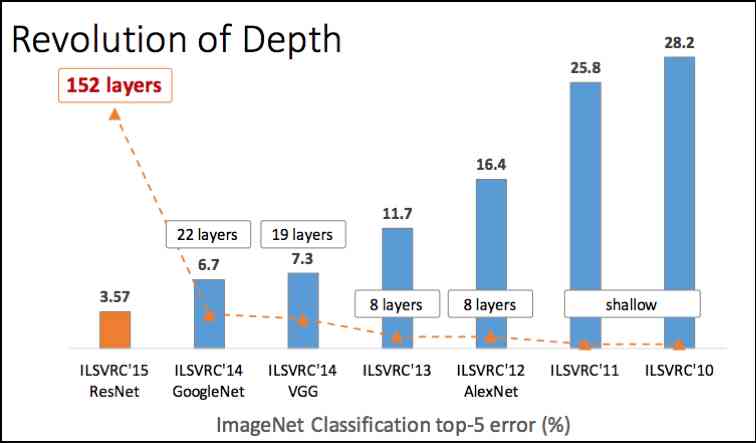}
\end{center}
\caption{Revolution of Depth of DNN on ImageNet Classification. Figure is from \cite{depthevolution}}
 \label{fig:depth}
 \end{figure}

Meanwhile, Figure \ref{fig:depth} from \cite{depthevolution} we can see the revolution of the depth of DNN on the ImageNet classification.\\

%%%%%%%%%%%%%%%%%%%%%%%%%%%%%%

%\cite{Schmidhuber201585}

 %%%%%%%%%%%%%%%%%%%%%%%%%%%%%%%%%%%%%%%%%%%%%%

\subsection{Object detection}

In the object detection research area, just like in the image classification where the ImageNet dataset is available to train the algorithm, in the object detection, COCO(Common Objects in Context)\cite{DBLP:journals/corr/LinMBHPRDZ14} and VOC dataset can be used to train the algorithms. For the COCO dataset, the bounding box and the mask of the objects are provided. The first important contribution of using deep neural networks to solve the object detection problem is the R-CNN\cite{DBLP:journals/corr/GirshickDDM13}. The region of interest(ROI) of an image is proposed by the selective search(SS)\cite{Uijlings:2013:SSO:2509349.2509382} algorithm, and the ROI is cropped to feed a CNN to do the detection. However, as every proposed interesting area will be calculated to predict whether that specified region is an object, the speed of this algorithm is very slow. In order to address this problem, a fast R-CNN algorithm is proposed in \cite{DBLP:conf/iccv/Girshick15} by improving the feature map generation efficiency. In another paper which is short for faster RCNN\cite{DBLP:conf/nips/RenHGS15}, instead of using the SS algorithm to generate ROI, the ROI is proposed by using deep neural network structure. By the combination, the performance of the algorithm can also be improved itself.\\

\subsubsection{Bounding box encoding method }
\begin{figure}[H]
  \begin{center}
     \includegraphics[width=0.55\linewidth]{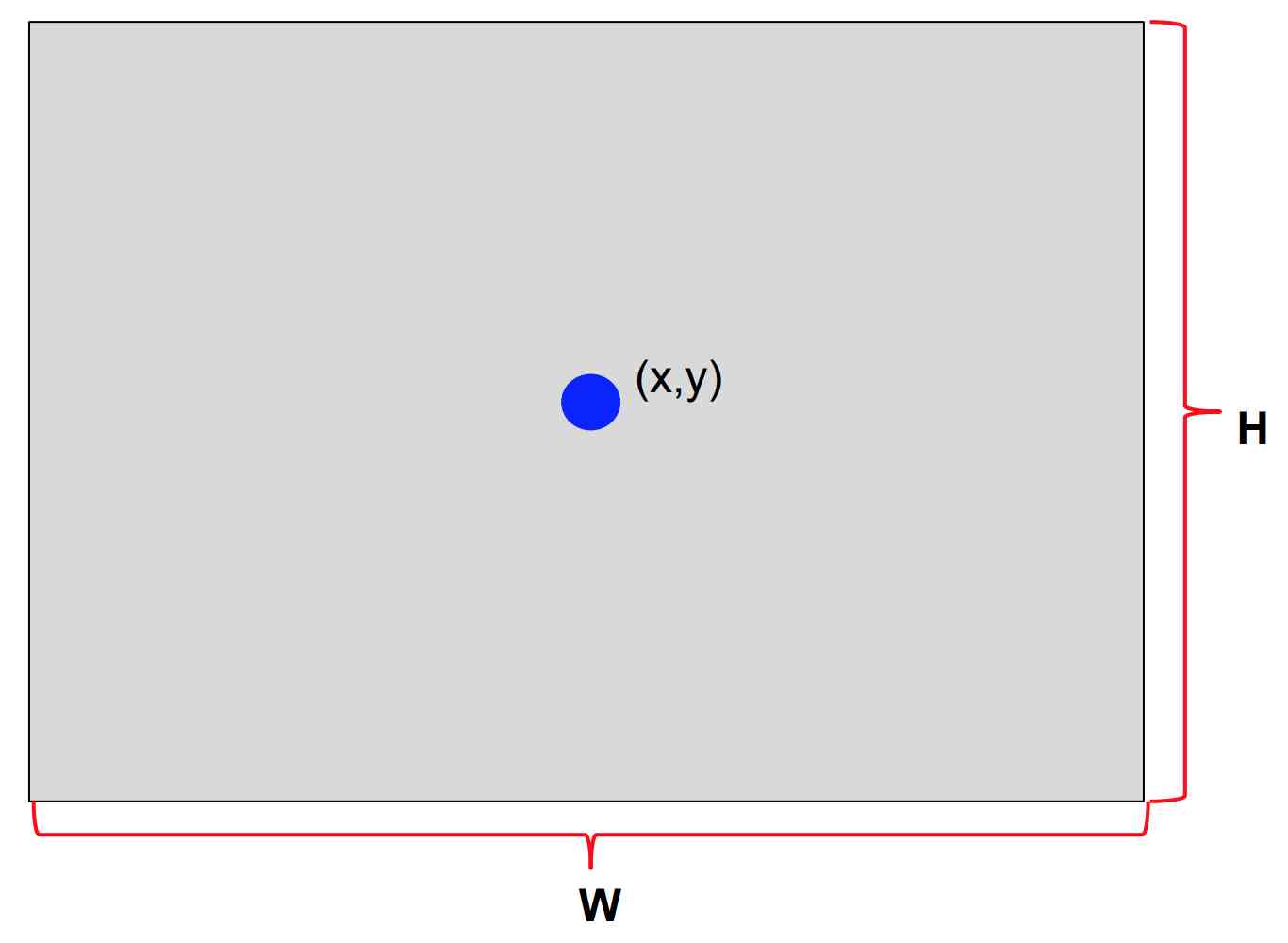}
\end{center}
\caption{The 2D center offset box encoding method.}
 \label{fig:bbox}
 \end{figure}

In the object detection task, one import part is estimating the bounding box of the object. The 2D bounding box estimation only focus on axis-aligned boxes. The box encoding method is simply based on a simple center coordinates of the bounding box $(x,y)$ and the offset of the bounding box: width: $W$ and height: $H$. We are calling this method as 2D center offset box encoding method and it is shown in Figure \ref{fig:bbox}. The RCNN, Fast RCNN, Faster RCNN, YOLO, YOLOv2 and Mask R-CNN are using this kind of bounding box encoding with a slightly difference on the loss calculation.\\

\subsubsection{Two-stage systems: RCNN\cite{DBLP:journals/corr/GirshickDDM13}, Fast RCNN\cite{DBLP:conf/iccv/Girshick15}  and Faster RCNN}
\begin{figure}[H]
  \begin{center}
     \includegraphics[width=1.0\linewidth]{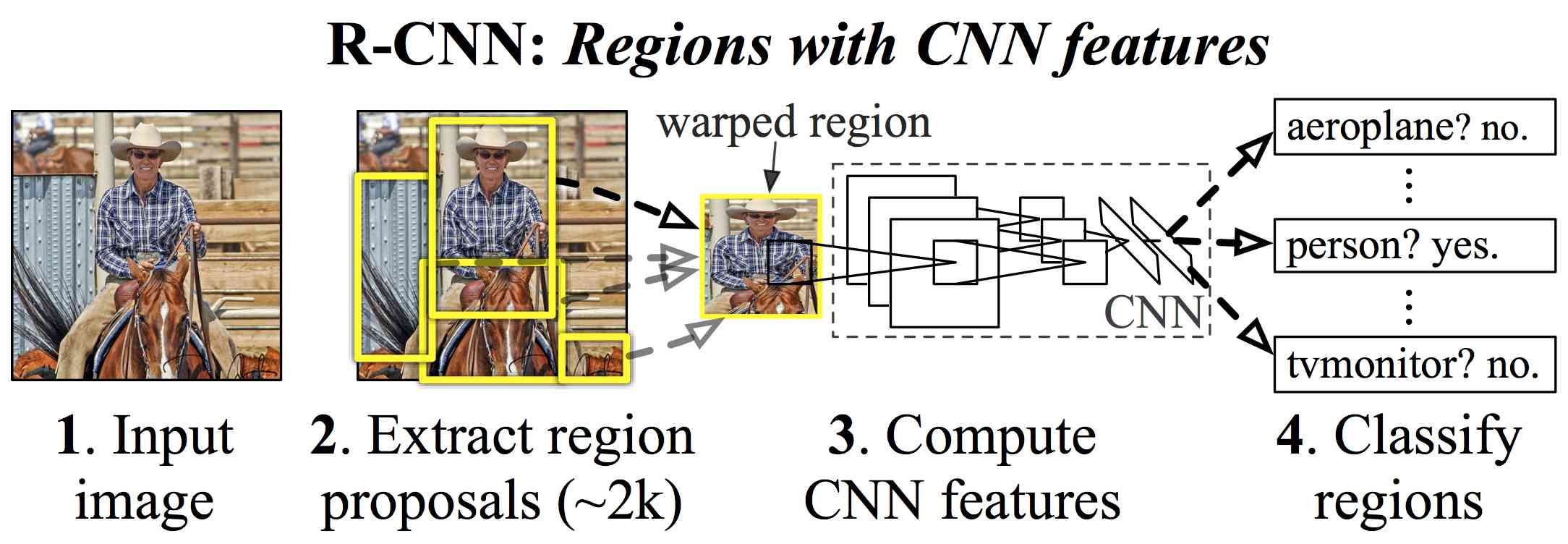}
\end{center}
\caption{The RCNN\cite{DBLP:journals/corr/GirshickDDM13} system object detection system overview: (1)
takes an input image, (2) extracts around 2000 bottom-up region
proposals, (3) computes features for each proposal using a large
convolutional neural network (CNN), and then (4) classifies each
region using class-specific linear SVMs. Figure and Caption are from original paper.}
 \label{fig:rcnn}
 \end{figure}
RCNN is an important framework in 2D image detection task which mainly uses the CNN to extract the features for each cropped interesting region. After that the extracted features are fed to a SVM to do the classification and a bounding box regression is followed to improve the bounding box prediction based on the method from \cite{Felzenszwalb:2010:ODD:1850486.1850574}.
It mainly has two stages: region proposal and detection. As the detection process is computation expensive, the region proposals can make the detection step mainly focus on limited interesting regions (about 2000 regions for a typical image) which greatly reduces the complexity of the whole system and achieves a good performance on the 2D image detection task. This two-stage detection framework is becoming a classical model in both 2D image based object detection and 3D image based object systems. The frame work of RCNN is shown in Figure \ref{fig:rcnn}.\\

\begin{figure}[H]
\begin{center}
%\fbox{\rule{0pt}{2in} \rule{.9\linewidth}{0pt}}
\includegraphics[width=1.0\linewidth]{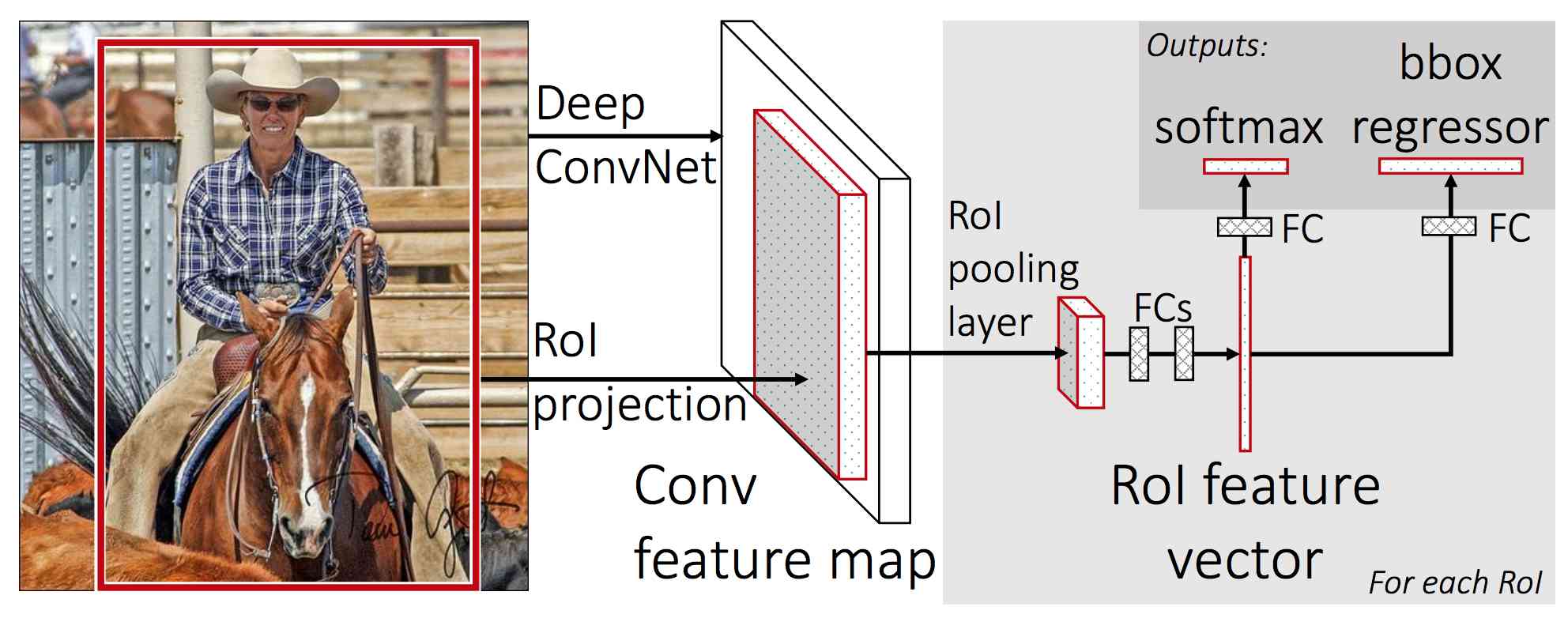}

\end{center}
   \caption{Fast R-CNN\cite{DBLP:conf/iccv/Girshick15}  architecture. An input image and multiple
regions of interest (RoIs) are input into a fully convolutional
network. Each RoI is pooled into a fixed-size feature map and
then mapped to a feature vector by fully connected layers (FCs).
The network has two output vectors per RoI: softmax probabilities
and per-class bounding-box regression offsets. The architecture is
trained end-to-end with a multi-task loss. Figure and Caption are from original paper.}
\label{fig:411}
\end{figure}

\begin{figure}[H]
\begin{center}
%\fbox{\rule{0pt}{2in} \rule{.9\linewidth}{0pt}}
\includegraphics[width=1.0\linewidth]{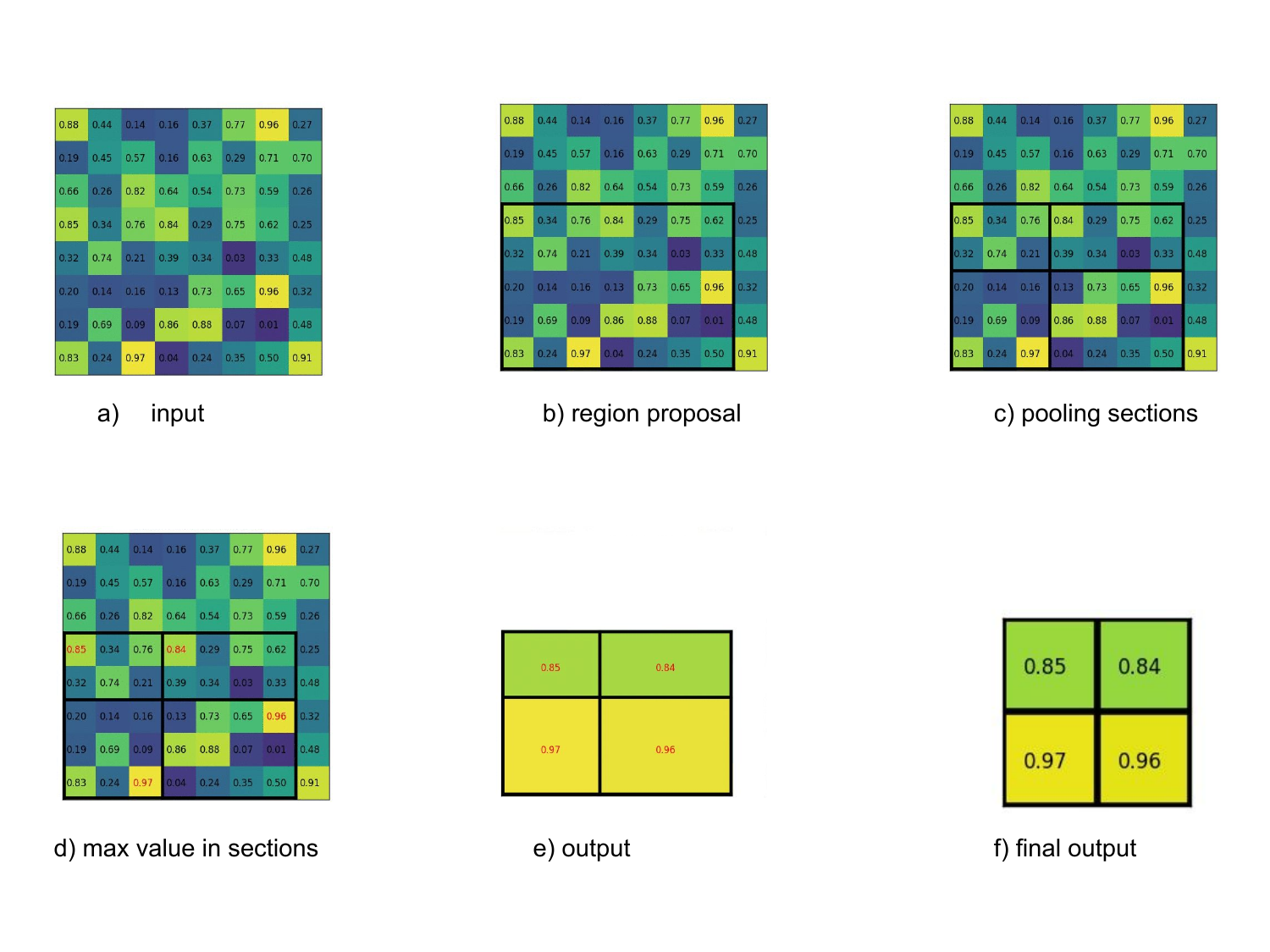}

\end{center}
   \caption{An example of the ROI pooling operation. The max pooling is used.  ROI pooling is based on a single 8×8 feature map, one region of interest and an output size of 2 by 2.  The figure is adjusted from \cite{roi_explained}.}
\label{fig:roi}
\end{figure}

Fast R-CNN improves the RCNN mainly with respect to three aspects: First, instead of doing convolution operations separately for each proposed region, the Fast RCNN does the convolution operations for the whole image firstly and then uses region proposals from the feature map directly to do the further detection. The feature map level region proposals are projected from the region proposals based on the original image. Second, using the softmax layer to replace the SVM classifier to make the detection under one deep learning framework. Finally, Fast R-CNN is using the Multi-task loss to do the object classification and the bounding box regression. \\
The Fast RCNN framework is shown in Figure \ref{fig:411}. In order to have a same size of feature vectors from different size proposed regions, the ROI pooling is used and the ROI pooling is demonstrated in Figure \ref{fig:roi}.

\begin{figure}[H]
\begin{center}
%\fbox{\rule{0pt}{2in} \rule{.9\linewidth}{0pt}}
\includegraphics[width=1.0\linewidth]{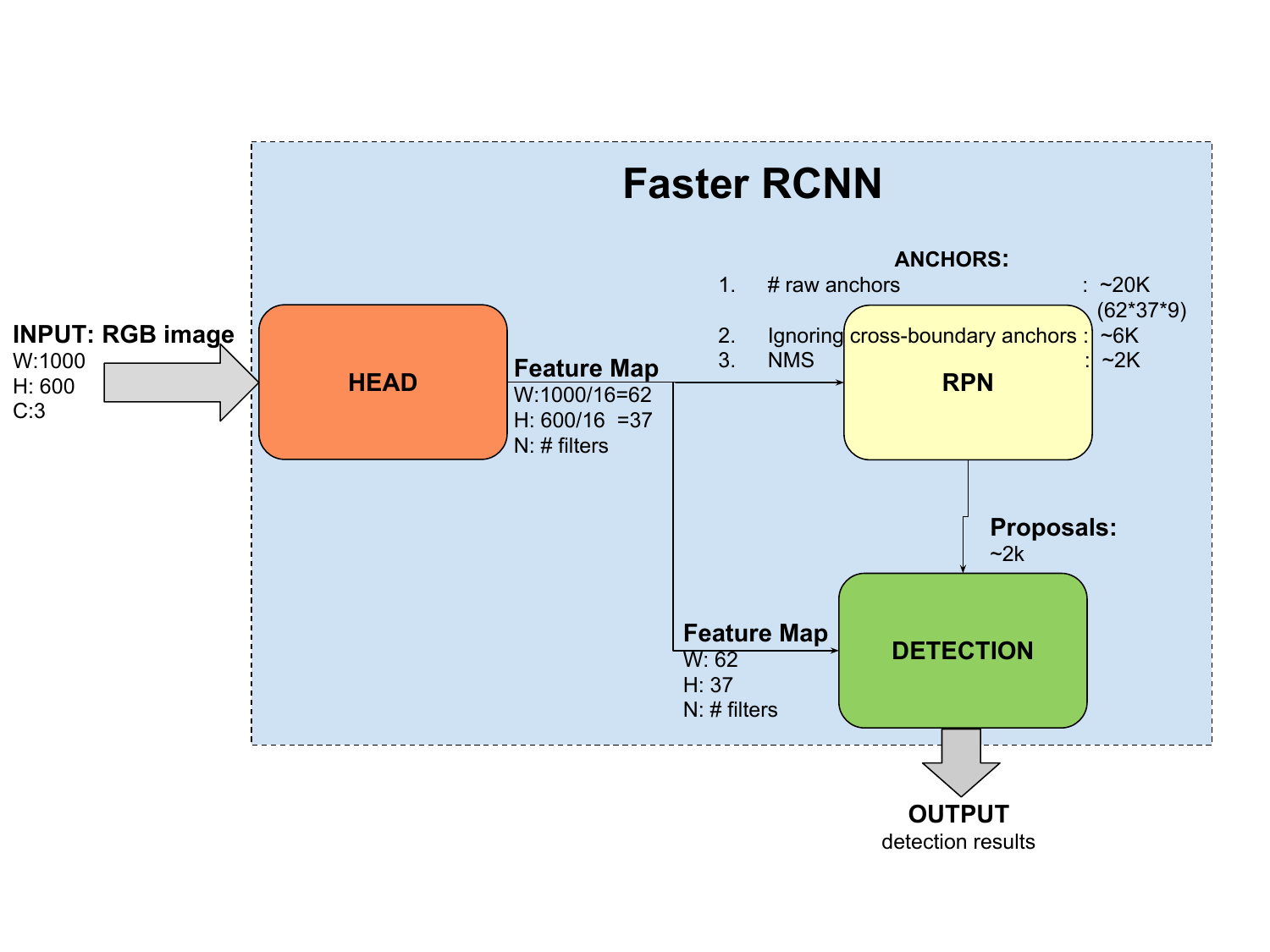}

\end{center}
   \caption{The three main steps for the Faster RCNN\cite{DBLP:conf/nips/RenHGS15}, system: head(backbone network), RPN and detection network.}
\label{fig:fast_rcnn}
\end{figure}

The main contribution of the Faster RCNN is introducing the region proposal network(RPN) under the deep learning framework. Before the RPN, the regional proposal is done by traditional method such as SS which is used in both RCNN and Fast RCNN. Since traditional methods such as SS and Edge Box are calculated by CPU, the speed is slow. At the same time, RPN can be calculated by GPU, also the convolutional layer of the  RPN and the detection network can be shared, the detection speed for the whole framework of Faster RCNN improved a lot. The framework of Faster RCNN is shown in Figure \ref{fig:fast_rcnn}. It mainly contains three steps: the head is used to extract the features by using CNN, then the RPN is used to get the region proposal. Finally, it is using the detection network to do the object detection.\\

\begin{figure}[H]
\begin{center}
%\fbox{\rule{0pt}{2in} \rule{.9\linewidth}{0pt}}
\includegraphics[width=1.0\linewidth]{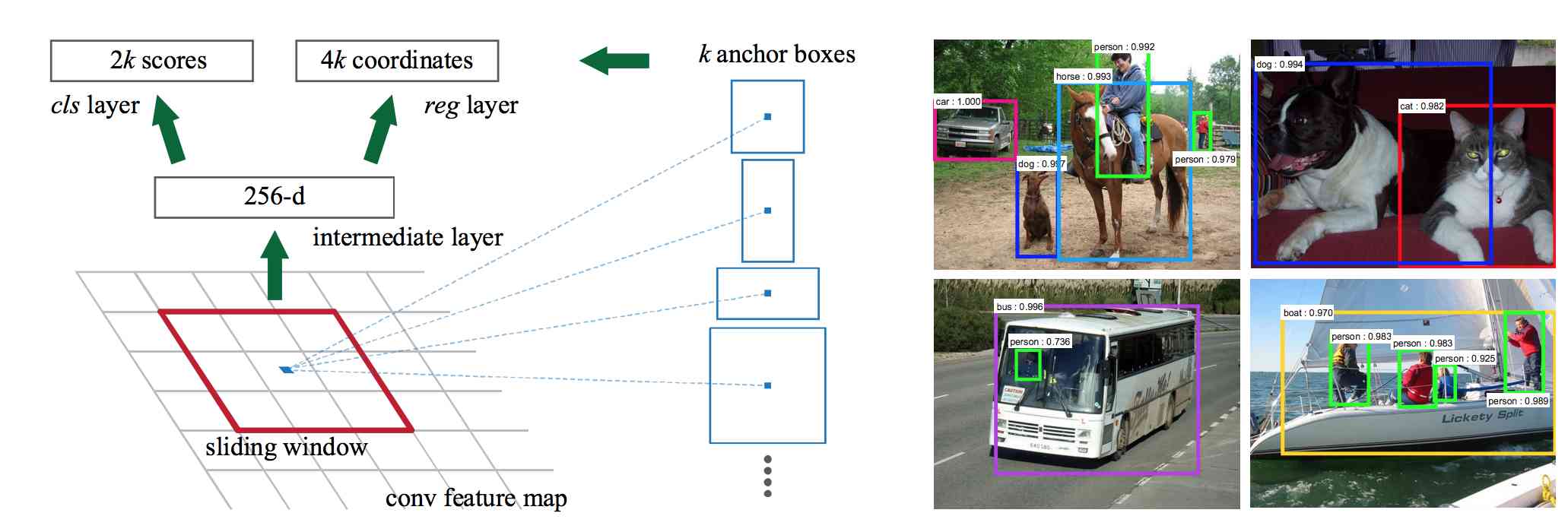}

\end{center}
   \caption{Left: Region Proposal Network (RPN)\cite{DBLP:conf/nips/RenHGS15}. Right: Example detections using RPN proposals on PASCAL
VOC 2007 test. Figure and Caption are from original paper. }
\label{fig:511}
\end{figure}

As shown in Figure \ref{fig:511}, 1) the RPN is done based on the feature map instead of the original image by using a sliding window method. The size of feature map is smaller than the original image by the pooling layer of the CNN. For example, when the original image size is $1000\times 600$, if the 4 pooling layers are used, then the size of the feature map will be  $62 \times 37$. The smaller size of feature map makes the regional proposals much faster. The proposal is done by using a $k \times k$ sliding window (the k in 3 in Faster RCNN), and different size and ratio anchors are used to get more accurate proposals. For each anchor, the RPN network will output 1) two scores: foreground score and background score 2) proposal bounding box: by using 2D center offset encoding, it will output 4 values. In Faster RCNN, 3 size and 3 ratio are selected. The comparison of the RPN and SS is shown in Table \ref{tab:rpn_result}. From the result, we can see, the RPN is faster and the performance is better than SS.\\

\begin{table}[H]
\begin{center}
\begin{tabular}{|c|c|c|c|}
\hline
method &$\#$ proposals& data& mAP \\
\hline\hline
SS& 2000 &07 &66.9\\
SS &2000 &07+12 &70.0\\
\hline
RPN+VGG,unshared &300 &07& 68.5\\
RPN+VGG, shared &300 &07& 69.9\\
RPN+VGG, shared &300 &07+12& 73.2\\
\hline
\end{tabular}
\end{center}
\caption{Detection results on PASCAL VOC 2007 test set. The detector is Fast R-CNN and VGG-16. Training
data: ``07": VOC 2007 trainval, ``07+12": union set of VOC 2007 trainval and VOC 2012 trainval. For RPN,
the train-time proposals for Fast R-CNN are 2000. Figure and Caption are from \cite{DBLP:conf/nips/RenHGS15},}
\label{tab:rpn_result}
\end{table}

 \subsubsection{One-stage systems: YOLO\cite{DBLP:journals/corr/RedmonDGF15} and YOLO v2\cite{DBLP:journals/corr/RedmonF16}}
 The YOLO and YOLO v2 systems are one-stage detection systems. They do not have a separately region proposal stage. Instead they divide the original image roughly into a $S\times S$ grid and then based on those grid cells will be used as a rough region to do the further processing.\\

 \begin{figure}[H]
\begin{center}
%\fbox{\rule{0pt}{2in} \rule{.9\linewidth}{0pt}}
\includegraphics[width=0.75\linewidth]{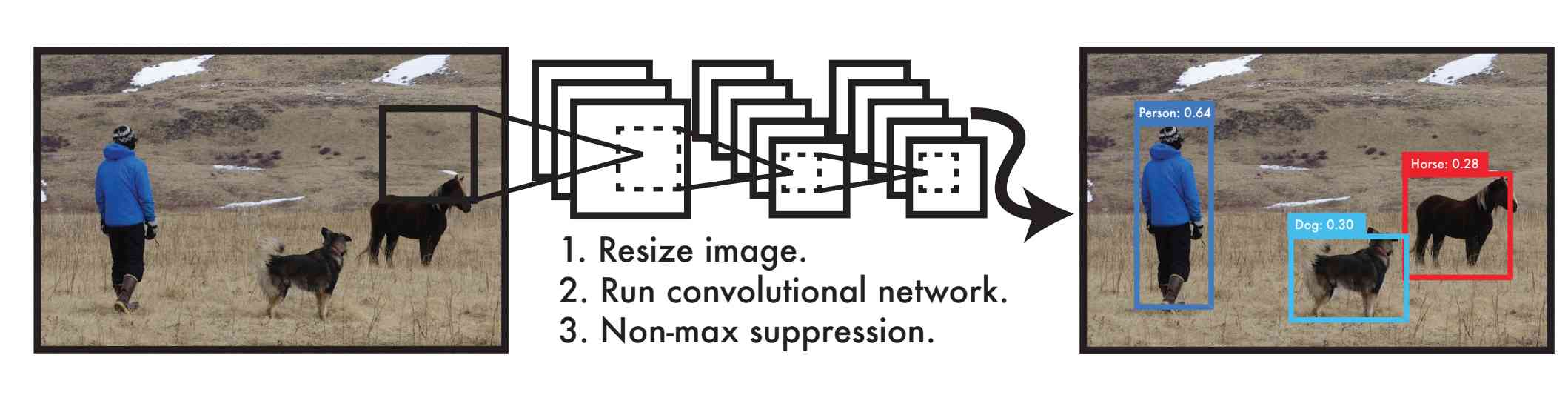}

\end{center}
   \caption{The YOLO\cite{DBLP:journals/corr/RedmonDGF15} Detection System. Processing images with YOLO\cite{DBLP:journals/corr/RedmonDGF15} is simple and straightforward. YOLO (1) resizes the input image to 448 x 448, (2) runs a single convolutional network on the image, and (3) thresholds the resulting detections by the model's confidence. Figure and Caption are from \cite{DBLP:journals/corr/RedmonDGF15}.}
\label{fig:611}
\end{figure}

 \begin{figure}[H]
\begin{center}
%\fbox{\rule{0pt}{2in} \rule{.9\linewidth}{0pt}}
\includegraphics[width=0.85\linewidth]{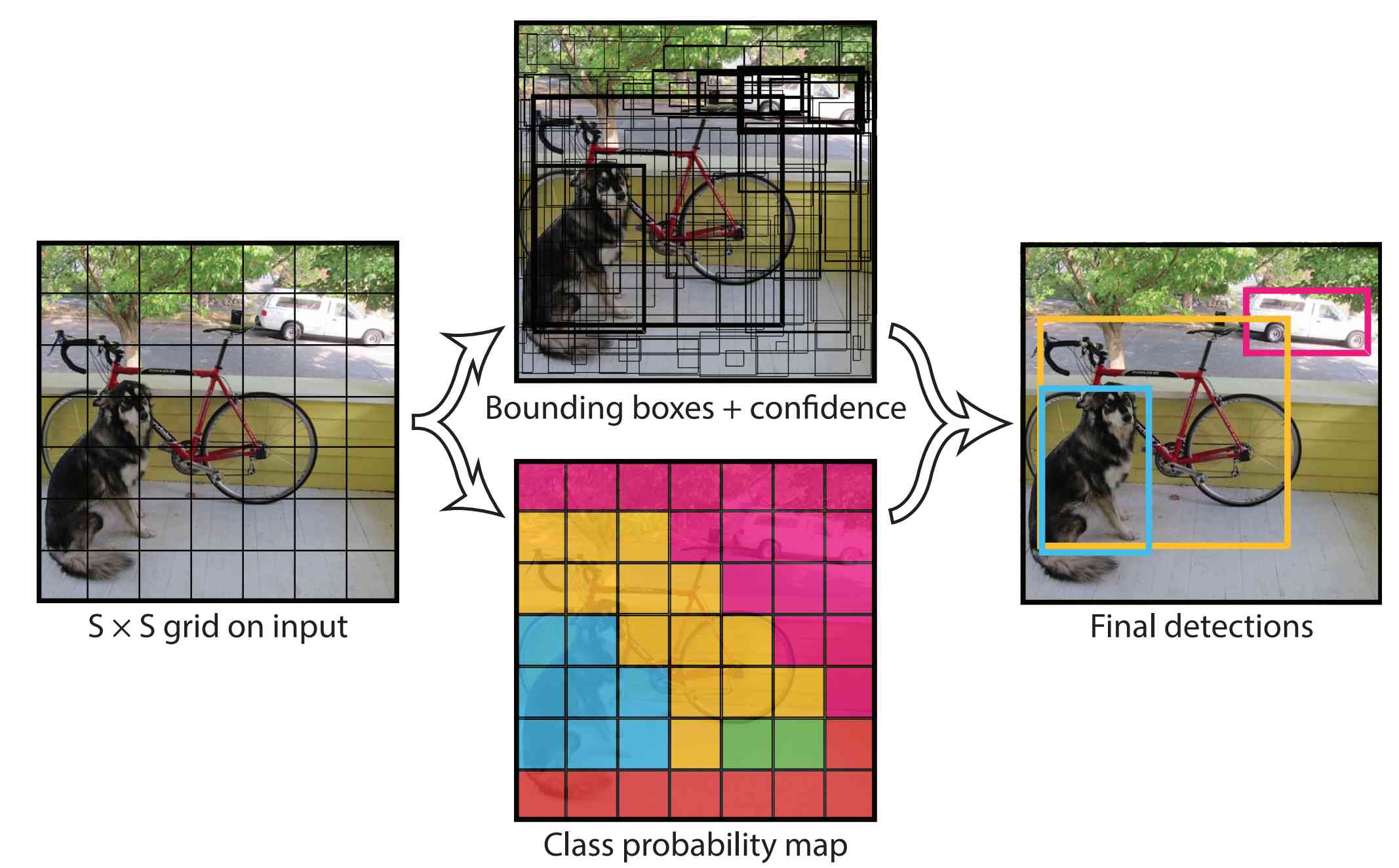}

\end{center}
   \caption{The Model. YOLO\cite{DBLP:journals/corr/RedmonDGF15} models detection as a regression
problem. It divides the image into an $S *S$ grid and for each
grid cell predicts $B$ bounding boxes, confidence for those boxes,
and $C$ class probabilities. These predictions are encoded as an
$S *S * (B * 5 + C)$ tensor. Figure and Caption are from \cite{DBLP:journals/corr/RedmonDGF15}.}
\label{fig:622}
\end{figure}

 \begin{figure}[H]
\begin{center}
%\fbox{\rule{0pt}{2in} \rule{.9\linewidth}{0pt}}
\includegraphics[width=1.0\linewidth]{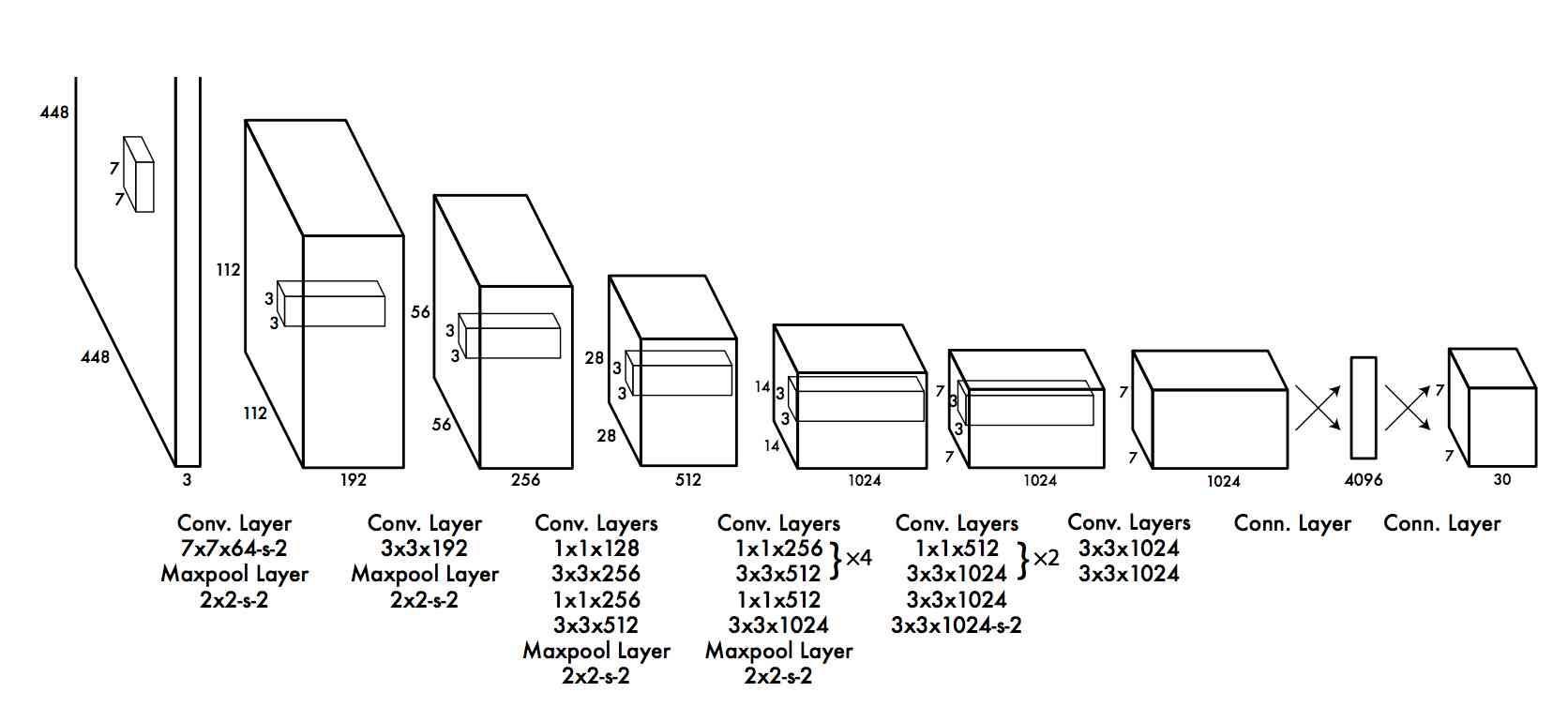}

\end{center}
   \caption{The Darknet Architecture. It has 24 convolutional layers followed by 2 fully connected layers. Alternating $1 \times 1$
convolutional layers reduce the features space from preceding layers. YOLO pretrains the convolutional layers on the ImageNet classification
task at half the resolution ($224 \times 224$ input image) and then double the resolution for detection. Figure and Caption are from \cite{DBLP:journals/corr/RedmonDGF15}.}
\label{fig:darknet}
\end{figure}

\begin{table}[H]
\begin{center}
\scalebox{0.9}{
\begin{tabular}{|c|c|c|c|c|c|}
\hline
method &network&box encoding& train data& mAP&FPS \\
\hline\hline
Faster RCNN\cite{DBLP:conf/nips/RenHGS15}&AlexNet&2D center offset encoding &07+12 &62.1&18\\
Faster RCNN\cite{DBLP:conf/nips/RenHGS15}&VGG16&2D center offset encoding &07+12 &73.2&7\\
\hline
YOLO\cite{DBLP:journals/corr/RedmonDGF15}&DarkNet&2D center offset encoding &07+12 &63.4&45\\
YOLO\cite{DBLP:journals/corr/RedmonDGF15}&VGG16&2D center offset encoding &07+12 &66.4&21\\
\hline
\end{tabular}
}
\end{center}
\caption{Comparison of the network, box encoding and performance for Faster RCNN and YOLO based on the detection results on PASCAL VOC 2007 test set. Training data: ``07+12": union set of VOC 2007 trainval and VOC 2012 trainval. The performance of both the Faster RCNN and YOLO can be improved by using more powerful network and other techniques such as batch normalization. In order to have a fair comparison, the improved results for both are not compared here. The improved results of the Faster RCNN can be found from the ResNet \cite{DBLP:journals/corr/HeZRS15} and YOLO can be found from YOLO v2\cite{DBLP:journals/corr/RedmonF16} }
\label{tab:yolo_faster_rcnn_result}
\end{table}

\begin{table}[H]
%\small
\begin{center}
\scalebox{0.8}{
\begin{tabular}{|c|c|c|c|c|c|c|}
\hline
method &input&resize& \makecell{feature map\\ or grid size}&\makecell{reception field \\for each\\ proposal}& \makecell{$\#$proposals \\for each\\ reception field}& \makecell{total \\proposals}\\
\hline\hline
\makecell{Faster\\ RCNN\cite{DBLP:conf/nips/RenHGS15}}&$224 \times 224$ &$446 \times 446$ & $27.9 \times 27.9$ & $48 \times 48$&9&7056\\
\hline
YOLO\cite{DBLP:journals/corr/RedmonDGF15}           &$224 \times 224$ &$446 \times 446$ & $7 \times 7$     & $63.7 \times 63.7$&2& 98\\
\hline
\end{tabular}
}
\end{center}
\caption{Comparison of the number of candidate proposals for the Faster RCNN and YOLO. In the Faster RCNN, the VOC image size is resized with the shorter length as 600 which can generate about 20K raw proposals. Here in order to make a fair comparison, it is resized by using the same dimension as YOLO.}
\label{tab:proposals_faster_yolo}
\end{table}

Another important paper for the object detection is YOLO\cite{DBLP:journals/corr/RedmonDGF15}\cite{DBLP:journals/corr/RedmonF16}. For the YOLO algorithm, the speed of detection is faster than the RCNN approach, however, the performance is slightly reduced. The Darknet structure which is used for YOLO is shown in Figure \ref{fig:darknet}. The Network, the bounding box encoding and the performance comparison is given in Table \ref{tab:yolo_faster_rcnn_result}. Meanwhile, the comparison of the proposal candidates numbers generated by the two method are compared in Table \ref{tab:proposals_faster_yolo}. From the comparison shown in Table \ref{tab:proposals_faster_yolo} we can explain that YOLO is faster because fewer proposals are considered by YOLO. However, this introduces worse performance and unsuccessful detection of small objects.\\

\subsubsection{Summary of 2D-based methods}

\begin{table}[H]
%\resizebox{\textwidth}{!}{%
%\small
\begin{center}
\begin{tabular}{|c|c|}
\hline
 & method\\
\hline\hline

\multirow{5}{*}{two-stage methods}
&R-CNN\cite{DBLP:journals/corr/GirshickDDM13}\\
\hhline{~-}
&fast/er R-CNN\cite{DBLP:conf/iccv/Girshick15}\cite{DBLP:conf/nips/RenHGS15}\\
\hhline{~-}
&mask R-CNN\cite{DBLP:journals/corr/HeGDG17}\\
\hhline{~-}
&Light-Head R-CNN\cite{DBLP:journals/corr/abs-1711-07264}\\
\hhline{~-}
&R-FCN\cite{DBLP:journals/corr/DaiLHS16}\\
\hline
\multirow{4}{*}{one-stage methods}
&YOLO\cite{DBLP:journals/corr/RedmonDGF15}, YOLO v2\cite{DBLP:journals/corr/RedmonF16}\\
\hhline{~-}
&SSD\cite{DBLP:journals/corr/LiuAESR15}\\
\hhline{~-}
&DSSD\cite{DBLP:journals/corr/FuLRTB17}\\
\hhline{~-}
&RetinaNet\cite{DBLP:journals/corr/abs-1708-02002}\\
\hline
\end{tabular}
\end{center}
\caption{Some common object detection frame work by stages: two-stage methods and one-stage methods }
\label{tab:two_stage_one_stage}
\end{table}

\begin{table}[H]
\begin{center}
\begin{tabular}{|c|c|}
\hline
 one-stage method & $\#$ candidate objects\\
\hline\hline
YOLO v1\cite{DBLP:journals/corr/RedmonDGF15} & 98\\
\hline
YOLO v2\cite{DBLP:journals/corr/RedmonF16}&$\sim$1k\\

\hline
OverFeat\cite{DBLP:journals/corr/SermanetEZMFL13}&$\sim$1-2k\\
\hline

SSD\cite{DBLP:journals/corr/LiuAESR15}&$\sim$8-26k(hard-example mining)\\
\hline
RetinaNet\cite{DBLP:journals/corr/abs-1708-02002}&$\sim$100-200k("soft'' hard-example mining)\\
\hline
\end{tabular}
\end{center}
\caption{ Number of candidate objects for different single-stage object detection methods.}
\label{tab:candidates}
\end{table}

\begin{figure}[H]
  \begin{center}
  \includegraphics[width=1.0\linewidth]{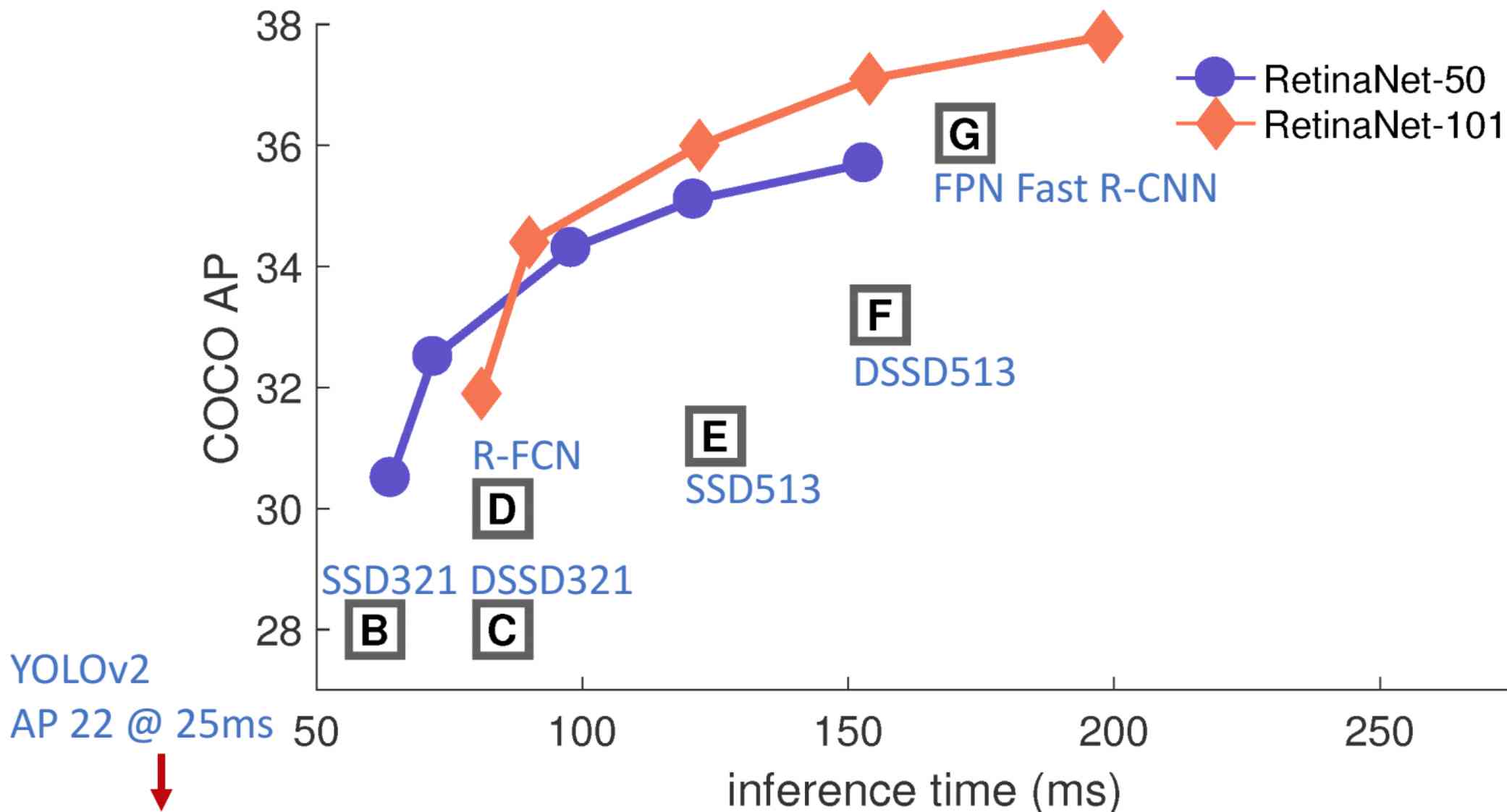}
\end{center}
\caption{Speed (ms) versus accuracy (AP) on COCO test-dev.
Enabled by the focal loss, RetinaNet detector outperforms all previous one-stage and two-stage detectors, including the best reported Faster R-CNN\cite{DBLP:conf/nips/RenHGS15} system from\cite{DBLP:journals/corr/LinDGHHB16}. Variants of RetinaNet with ResNet-50-FPN (blue circles) and ResNet-101-FPN (orange diamonds) are shown at five scales (400-800 pixels). Ignoring the low-accuracy regime ($AP < 25$), RetinaNet forms an upper envelope of all current detectors, and an improved variant (not shown) achieves 40.8 AP. Figure and Caption are adjusted from \cite{DBLP:journals/corr/abs-1708-02002}}
 \label{fig:RetinaNet}
 \end{figure}

\begin{figure}[H]
  \begin{center}
  \includegraphics[width=1.0\linewidth]{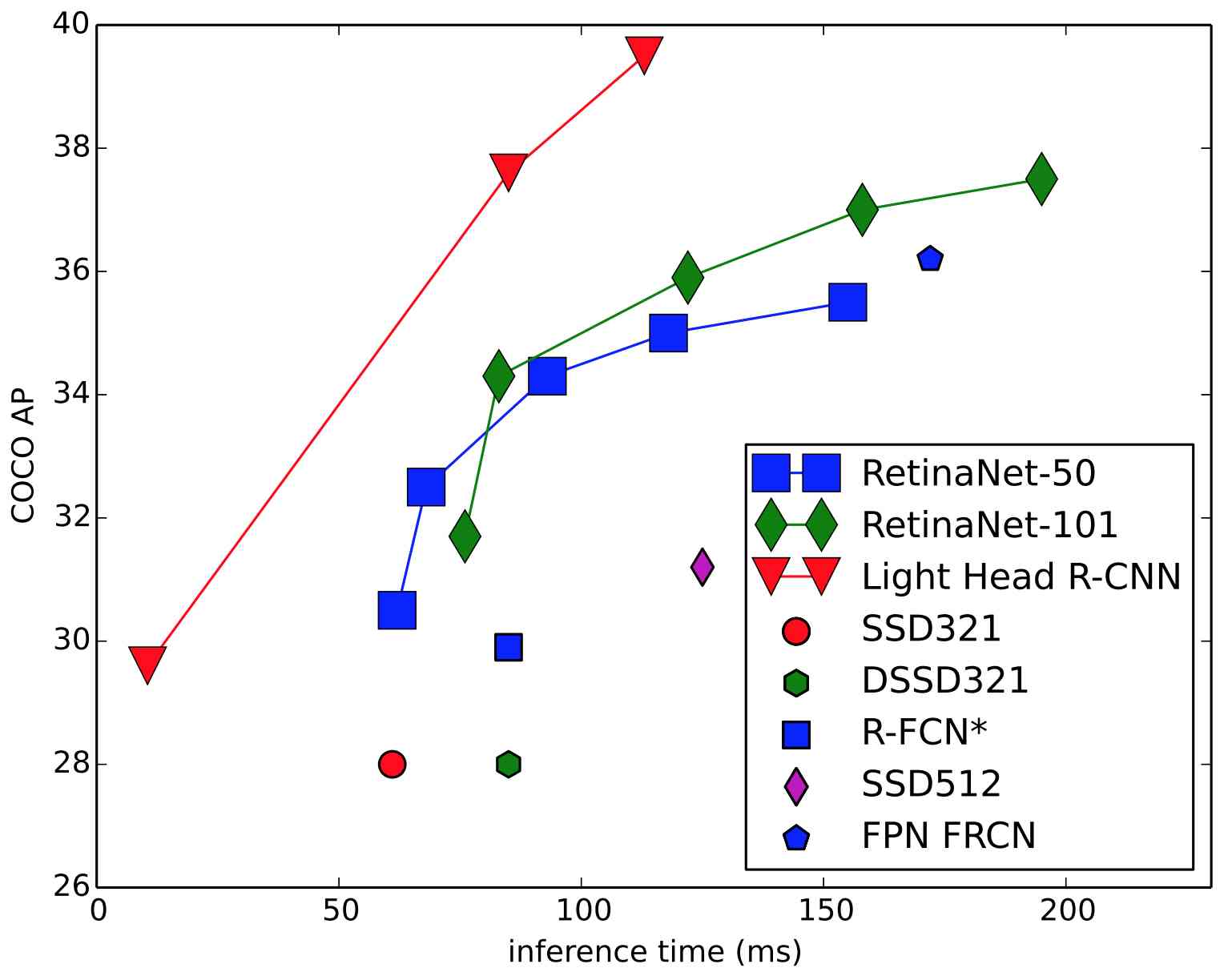}
\end{center}
\caption{Comparisons of some one-stage and two-stage methods. Figure is from \cite{DBLP:journals/corr/abs-1711-07264}.}
 \label{fig:light_head}
 \end{figure}

From the description above we can have a general understanding that for the object detection, there are two kinds of methods: 1) Two-stage methods Detector 2) One-stage methods. Part of those detection algorithms are organized by those two different stage methods and are shown in Table \ref{tab:two_stage_one_stage}. Actually,  there is a competition between the two-stage methods and one-stage methods. Generally, the one-stage method is improved by introducing more proposals as shown in Table \ref{tab:candidates} and by introducing new loss functions such as Focal loss\cite{DBLP:journals/corr/abs-1708-02002} to get rid of the unbalance between the positive proposals and negative proposals. Meanwhile, two-stage methods are also trying to improve the speed by introducing the lighter head, for example Light Head R-CNN\cite{DBLP:journals/corr/abs-1711-07264}.  The comparison of those different stage method's performance is shown in Figure \ref{fig:RetinaNet} and \ref{fig:light_head} based on the COCO dataset.\\

%In \cite{DBLP:journals/corr/abs-1708-02002} there is a comparison of the inference speed and the accuracy of the different methods.\\

%In the year of 2017, the FPN\cite{DBLP:journals/corr/LinDGHHB16} is used to do the object detection. The basic idea for this network structure is combing different sized network together to boost the performance. Similar structure is used in AVOD and get a performance boost for the small objects such as pedestrian and bicyclist. There is no improvement for the big object such as car. \\

%\subsection{Microsoft COCO: Common Objects in Context  }
%\subsection{Rich feature hierarchies for accurate object detection and semantic
%               segmentation\cite{DBLP:journals/corr/GirshickDDM13}}
%\subsection{Fast R-CNN\cite{DBLP:conf/iccv/Girshick15}}
%\subsection{Faster R-CNN: Towards Real-Time Object Detection with Region Proposal
   %            Networks\cite{DBLP:conf/nips/RenHGS15}}%
%\subsection{You Only Look Once: Unified, Real-Time Object Detection\cite{DBLP:journals/corr/RedmonDGF15}}
%\subsection{YOLO9000: Better, Faster, Stronger\cite{DBLP:journals/corr/RedmonF16}}

\subsection{Semantic segmentation}
For semantic segmentation, a pixel level detection of an object is provided. One important paper in this area is Fully Convolutional Network(FCN)\cite{DBLP:journals/corr/LongSD14}. It upsamples the feature map to make sure that a more accurate location information can be preserved. In addition, the data in the previous layers are combined with the deeper layer to preserve more information and thus improve the accuracy of the semantic segmentation. After this paper, FCN became mainstream in semantic segmentation.  DeepLab\cite{DBLP:journals/corr/ChenPK0Y16}, FCIS\cite{DBLP:journals/corr/LiQDJW16} and mask-RCNN\cite{DBLP:journals/corr/HeGDG17} are using FCN. For the DeepLab\cite{DBLP:journals/corr/ChenPK0Y16}, the CRF is used to further improve the result by benefiting of the redundant information of nearby pixels. The CRF approach is firstly introduced in \cite{DBLP:journals/corr/ZhengJRVSDHT15}. For FCIS\cite{DBLP:journals/corr/LiQDJW16}, location sensitive feature maps are generated to improve the pixel level prediction. The FCIS\cite{DBLP:journals/corr/LiQDJW16} is the improved version of \cite{DBLP:journals/corr/DaiLHS16}. So far, the FCN and the CRF approach became the standard method in the semantic segmentation area. Besides those papers, \cite{DBLP:journals/corr/HariharanAGM14a}\cite{DBLP:journals/corr/PinheiroCD15} are also using deep neural network approaches to improve the performance of the semantic segmentation.\\

\begin{figure}[H]
  \begin{center}
  %0.4
      \includegraphics[width=1.0\linewidth]{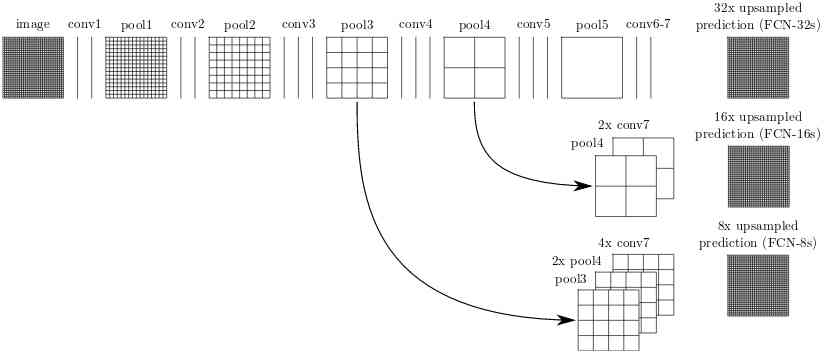}
\end{center}
\caption{The FCN\cite{DBLP:journals/corr/LongSD14} structure. Figure is from the original paper.}
 \label{fig:fcn}
 \end{figure}
 \subsubsection{Mask R-CNN\cite{DBLP:journals/corr/HeGDG17}}

\begin{figure}[H]
\begin{center}
%\fbox{\rule{0pt}{2in} \rule{.9\linewidth}{0pt}}
\includegraphics[width=0.8\linewidth]{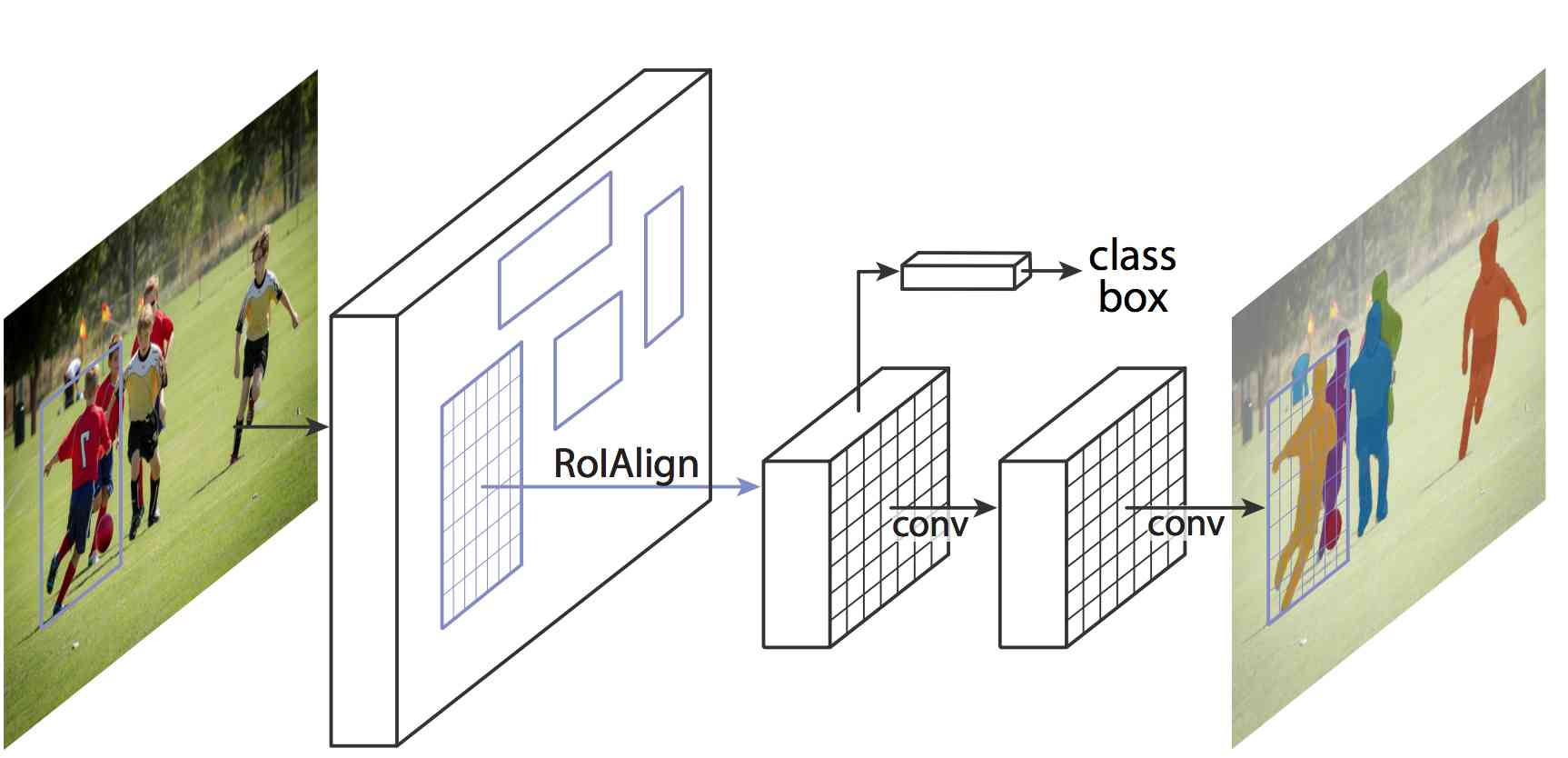}

\end{center}
   \caption{The Mask R-CNN\cite{DBLP:journals/corr/HeGDG17} framework for instance segmentation. Figure is from original paper.}
\label{fig:1011}
\end{figure}

\begin{figure}[H]
\begin{center}
%\fbox{\rule{0pt}{2in} \rule{.9\linewidth}{0pt}}
\includegraphics[width=0.5\linewidth]{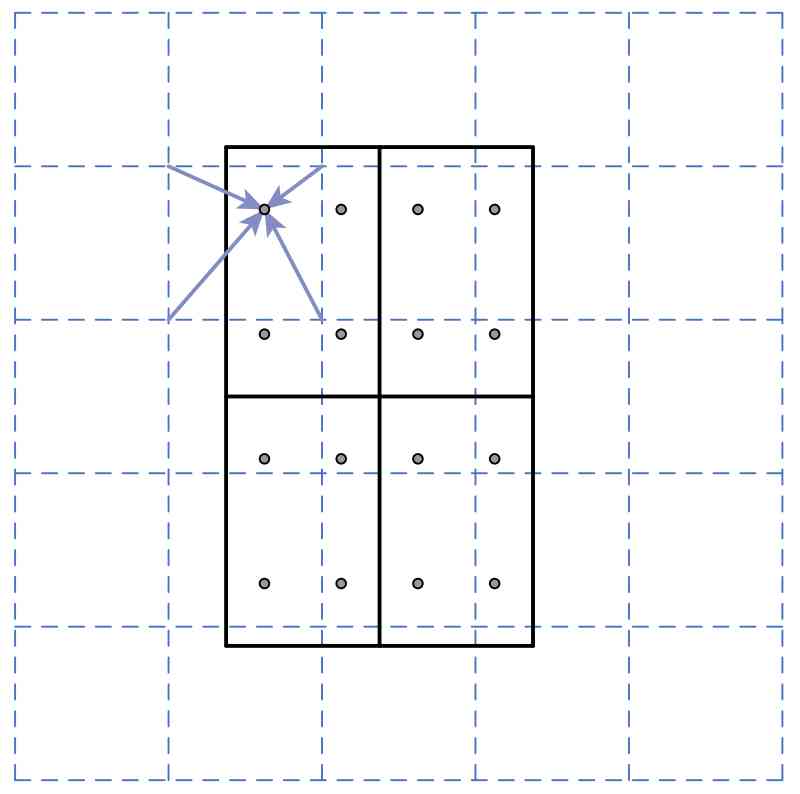}

\end{center}
   \caption{RoIAlign: The dashed grid represents
a feature map, the solid lines an RoI
(with $2\times2$ bins in this example), and the dots
the 4 sampling points in each bin. RoIAlign
computes the value of each sampling point
by bilinear interpolation from the nearby grid
points on the feature map. No quantization is
performed on any coordinates involved in the
RoI, its bins, or the sampling points. Figure and Caption are from \cite{DBLP:journals/corr/HeGDG17}}
\label{fig:1044}
\end{figure}

\begin{figure}[H]
\begin{center}
%\fbox{\rule{0pt}{2in} \rule{.9\linewidth}{0pt}}
\includegraphics[width=0.6\linewidth]{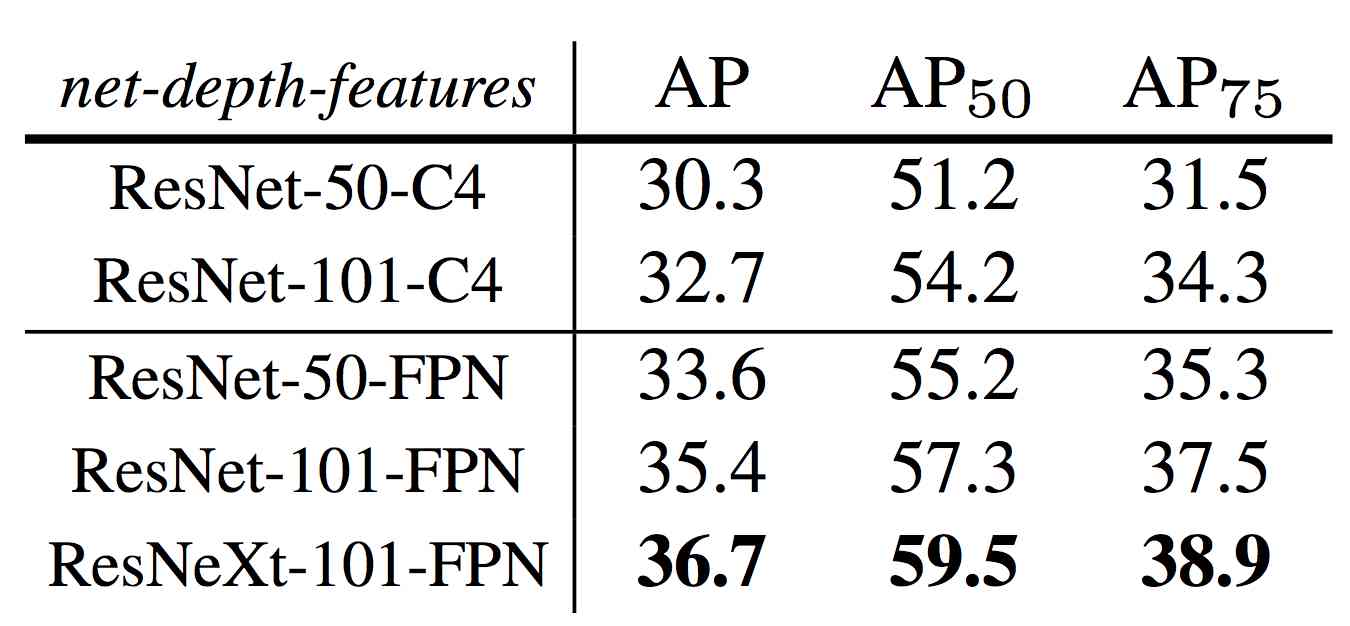}

\end{center}
   \caption{Backbone Architecture: Better backbones
bring expected gains: deeper networks
do better, FPN outperforms C4 features, and
ResNeXt improves on ResNet. Figure and Caption are from \cite{DBLP:journals/corr/HeGDG17}}
\label{fig:1022}
\end{figure}
 
 \begin{figure}[H]
\begin{center}
%\fbox{\rule{0pt}{2in} \rule{.9\linewidth}{0pt}}
\includegraphics[width=0.8\linewidth]{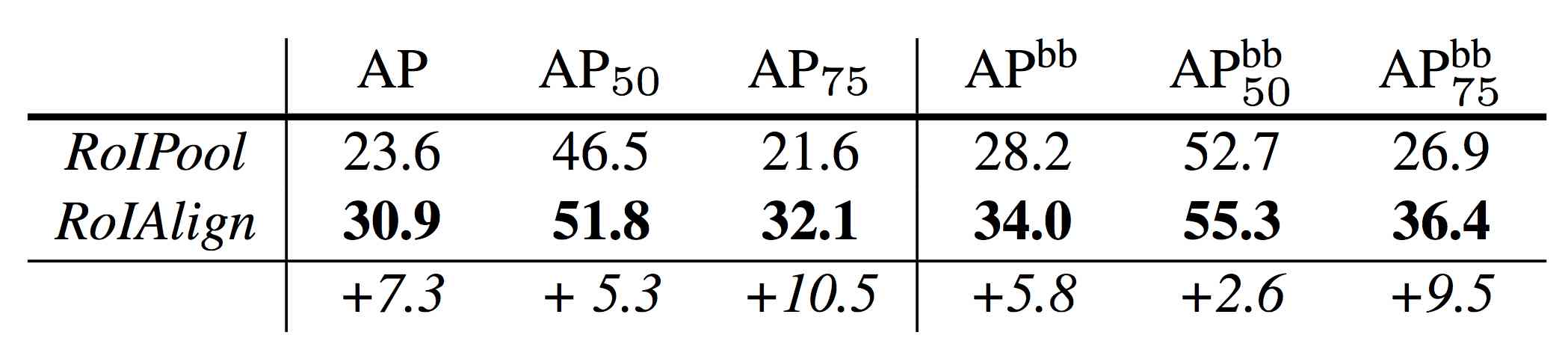}

\end{center}
   \caption{RoIAlign (ResNet-50-C5, stride 32): Mask-level and box-level
AP using large-stride features. Misalignments are more severe than
with stride-16 features , resulting in big accuracy gaps. Figure and Caption are from \cite{DBLP:journals/corr/HeGDG17}}
\label{fig:1033}
\end{figure}

The Mask R-CNN framework is shown in Figure \ref{fig:1011}. One important contribution of Mask R-CNN is RoIAlign as shown in Figure \ref{fig:1044}. The alignment of the ROI is not so sensitive in the object detection, however, in the semantic segmentation it is sensitive as it has to do the detection in pixel level and by employing the alignment, the performance is improved a lot. Another important fact for the improvement of the performance of the Mask R-CNN is using the more powerful backbone networks as observed from the results shown in Figure \ref{fig:1022}. The performance of RoIAlign is shown in Figure \ref{fig:1033}.\\
 
\section{3D-image based systems}
In this section, 3D-image based systems will be introduced. The main high-level tasks in 3D vision, 3D image data representation and methods used to do the 3D classification and object detection are briefly described. Several import papers in this area are surveyed.\\
\subsection{Main high-level tasks}

\begin{figure}[H]
\begin{center}
%\fbox{\rule{0pt}{2in} \rule{.9\linewidth}{0pt}}
\includegraphics[width=1.0\linewidth]{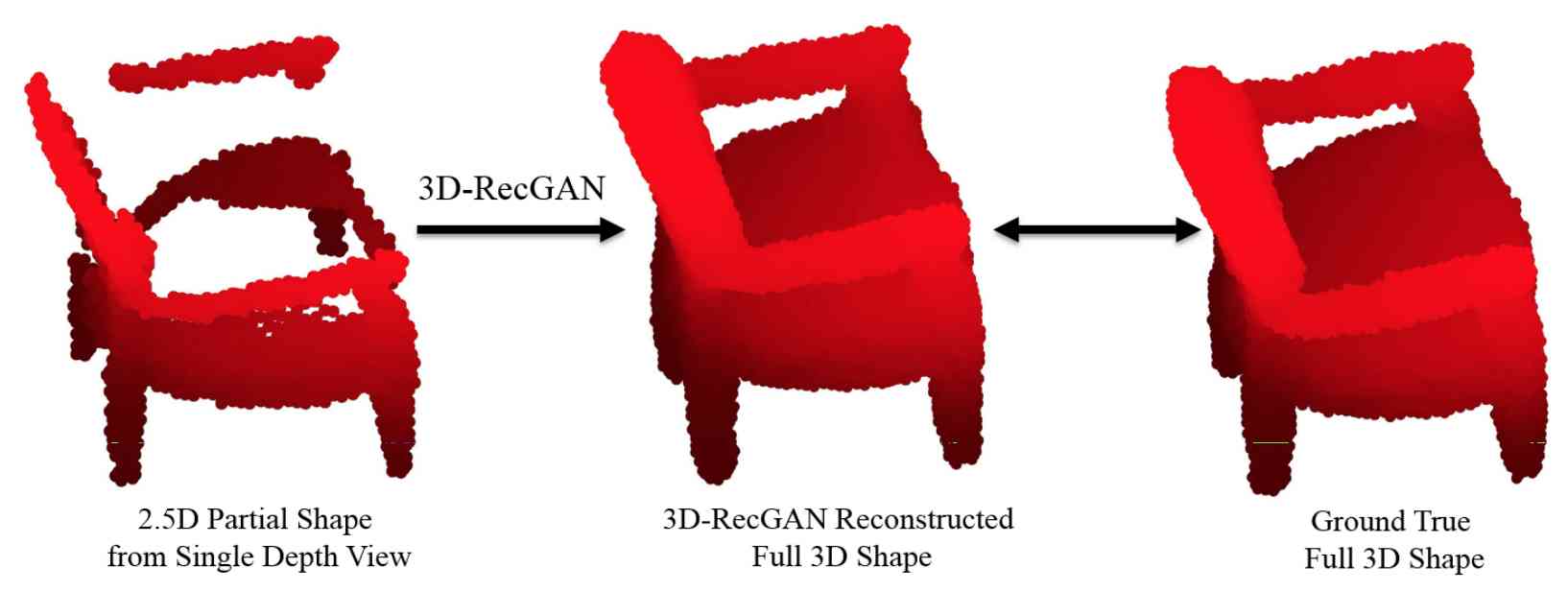}

\end{center}
   \caption{A 3D shape reconstruction example: 3D-RecGAN reconstructs a full 3D shape from a single 2.5D depth view\cite{DBLP:journals/corr/abs-1708-07969}}
\label{fig:3D-RecGAN}
\end{figure}

Similar to the 2D systems, the main tasks also include 3D object classification, 3D object detection, 3D semantic segmentation and 3D instance segmentation. Also, as 3D images commonly suffer from occlusion such as self occlusion and inter object occlusion. In order to address this issue, a 3D shape reconstruction task is also actively researched in the 3D vision understanding community. This part is important, however, it will not be surveyed. An example of 3D shape reconstruction is shown in Figure \ref{fig:3D-RecGAN} \\

\subsection{3D image data representation }
3D images can have multiple data representations, such as multi-view RGB-D images, volumetric images, polygonal meshes and point clouds.\\

\begin{figure}[H]
  \begin{center}
  %0.75
      \includegraphics[width=1.0\linewidth]{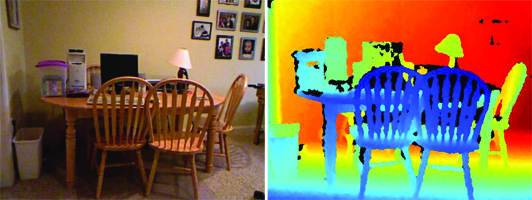}
\end{center}
\caption{Output from the RGB camera (left) and depth camera (right). Missing values in the depth image are a result of (a) shadows caused by the disparity between the infrared emitter and camera or (b) missing or spurious values caused by specular or low albedo surfaces.}
 \label{nyu_depth_v2_raw}
 \end{figure}

 \begin{figure}[H]
  \begin{center}
  %0.22
      \includegraphics[width=1.0\linewidth]{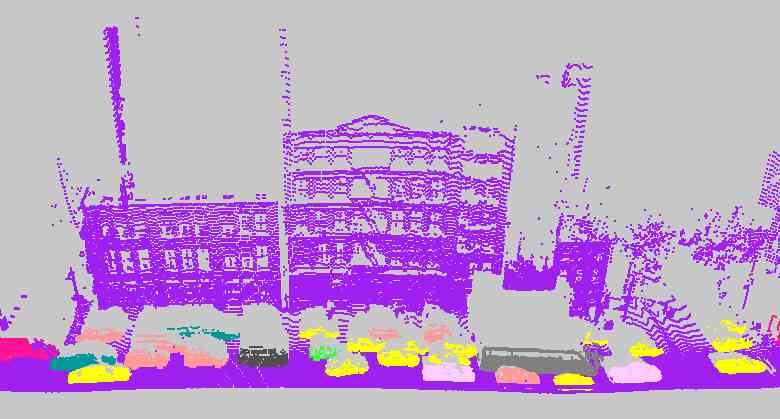}
\end{center}
\caption{An example of the 3D data from the outdoor urban LiDAR scans. The figure is provided by the Computer Vision Laboratory of Hunter College\cite{hunter_vision}. }
 \label{fig:LiDAR2}
 \end{figure}
 
   \begin{figure}[H]
  \begin{center}
  %0.4
      \includegraphics[width=1.0\linewidth]{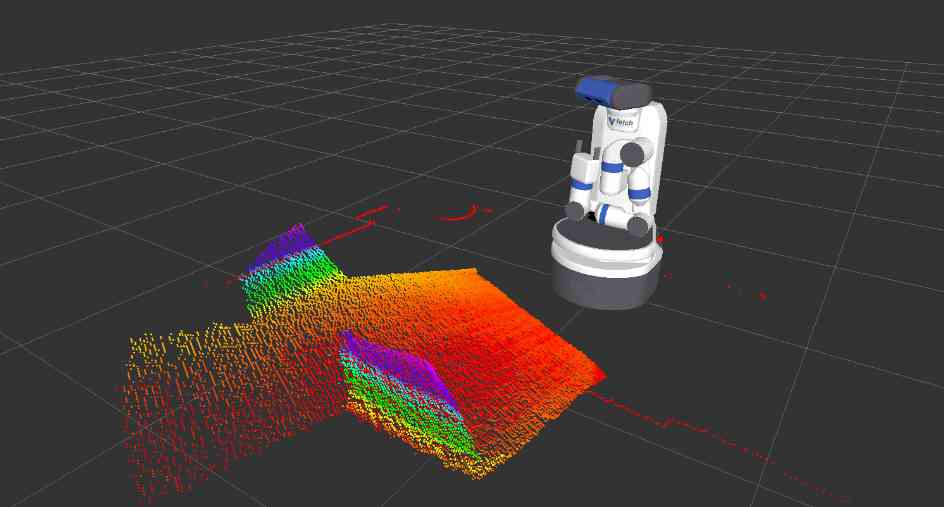}
\end{center}
\caption{The 3D view from the robot's perspective.}
 \label{fig:fetch}
 \end{figure}

   \begin{figure}[H]
  \begin{center}
  %0.4
      \includegraphics[width=1.0\linewidth]{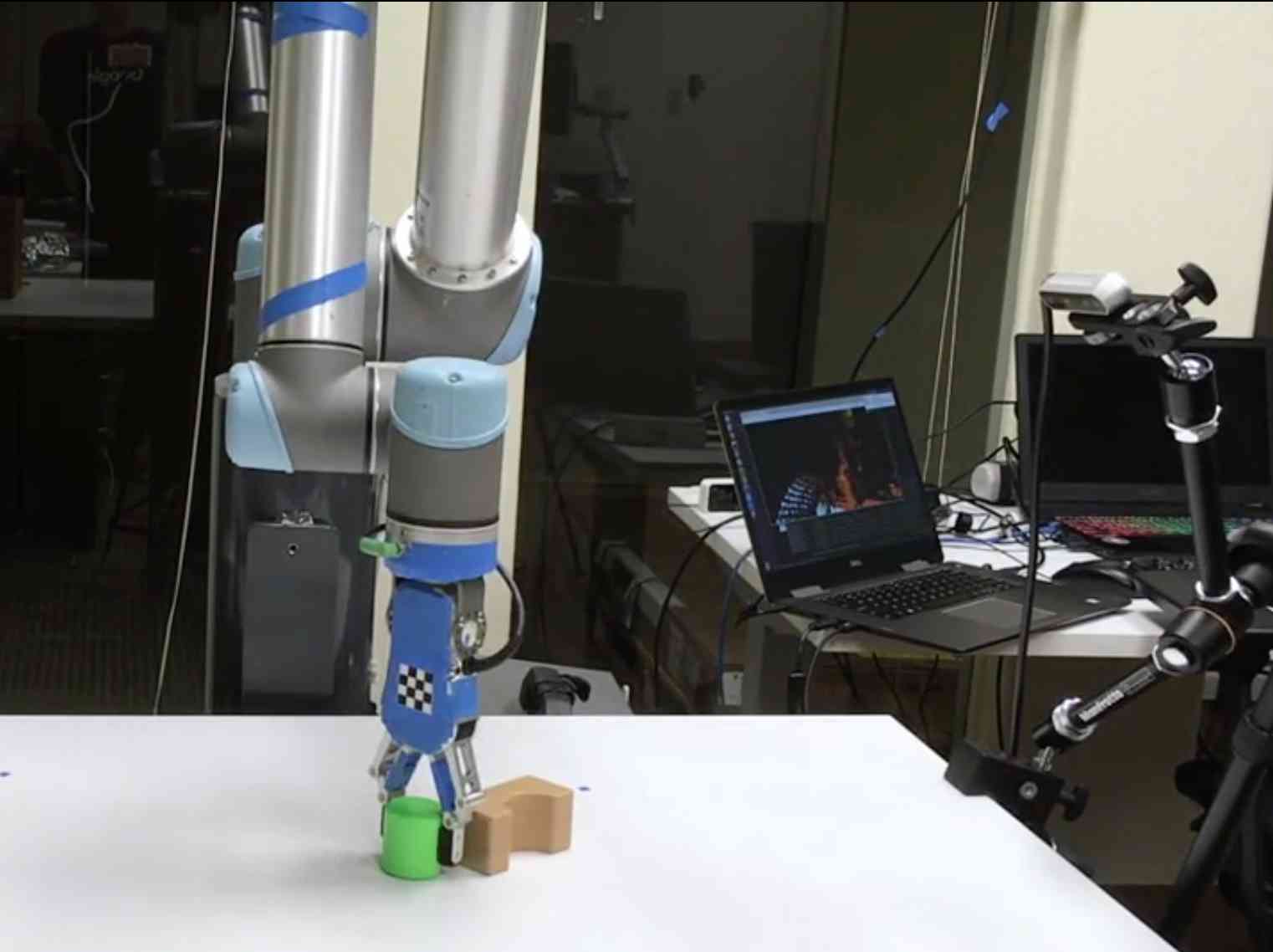}
\end{center}
\caption{A robot is picking up an object with the help of the RGB-D camera based object detection system.}
 \label{fig:fetch2}
 \end{figure}

3D images are becoming more and more important and are widely used in reconstructing architectural models of buildings, navigation of self-driving cars, detection face (such as face ID for iPhone X), preservation of at-risk historical sites, and recreation of virtual environments for film and video game industries.\\
Mainly, there are two basic kinds of hardware available for the 3D data generation in outdoor and indoor environments. For outdoors, one typical hardware is LiDAR( Light Detection and Ranging). The coverage of this equipment can achieve to hundreds and even thousands meters. Most self-driving cars use a LiDAR scanner. For indoors, in recent years the availability of low-cost sensors such as the Microsoft Kinect
have enabled the acquisition of short-range indoor 3D data at the consumer level. Meanwhile, smart phone such as iPhone X a equipped will a depth camera. In Figure \ref{fig:LiDAR2}, one example of the 3D data collected from the outdoor urban LiDAR scanner is shown. In Figure \ref{nyu_depth_v2_raw}, depth map generated by a Kinect camera is provided. Capturing a 3D environment by a robot is shown in Figure \ref{fig:fetch}. A RGB-D camera based object detection system is used to help a robot to grasp objects as shown in Figure \ref{fig:fetch2}.\\

\begin{figure}[H]
  \begin{center}
  %0.22
      \includegraphics[width=0.5\linewidth]{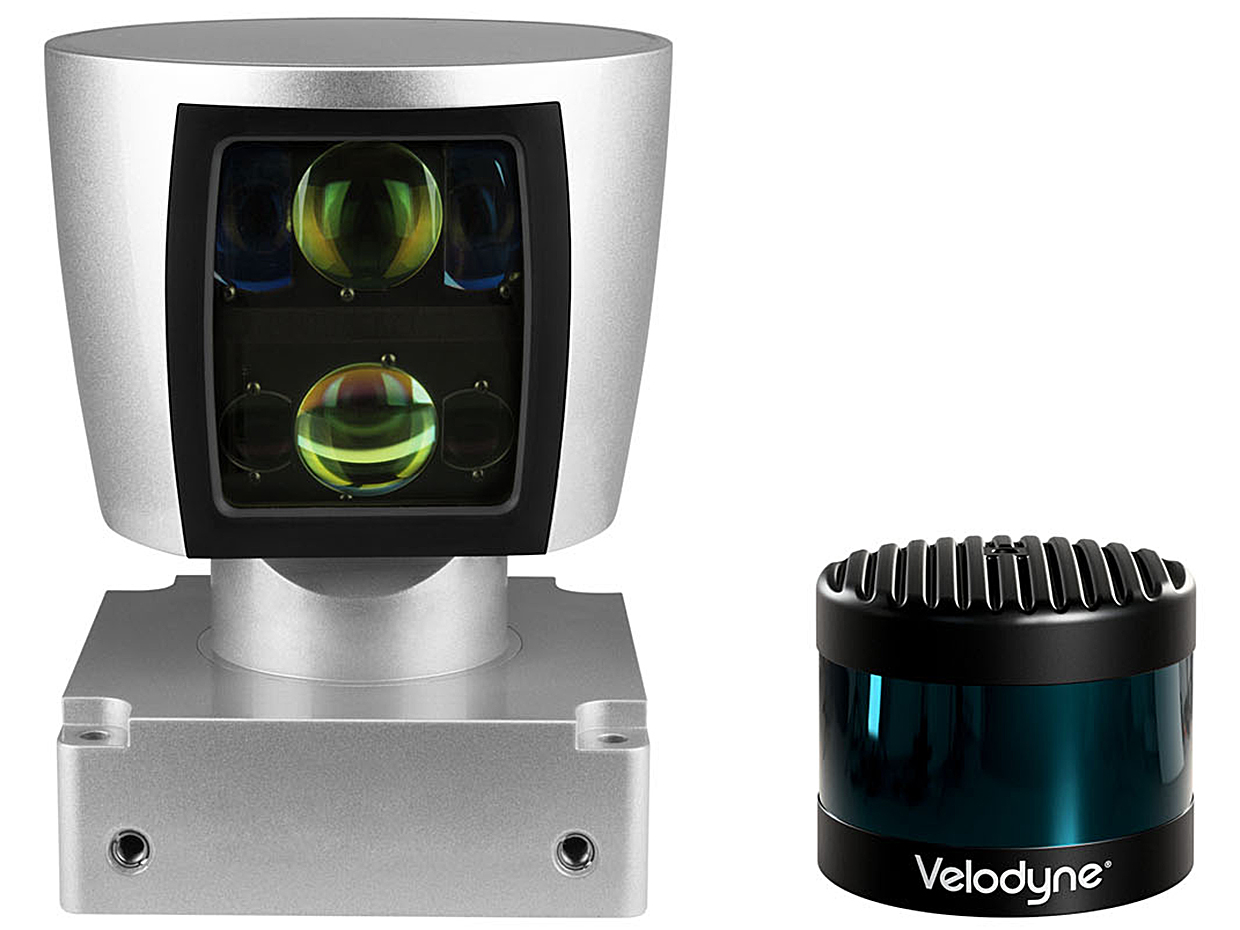}
\end{center}
\caption{Shown here are Velodyne's HDL-64E LiDAR sensor (left) and the company's VLS-128 sensor (right).\cite{velodyne}}
 \label{velodyne}
 \end{figure}

   \begin{figure}[H]
  \begin{center}
  %0.35
      \includegraphics[width=1.0\linewidth]{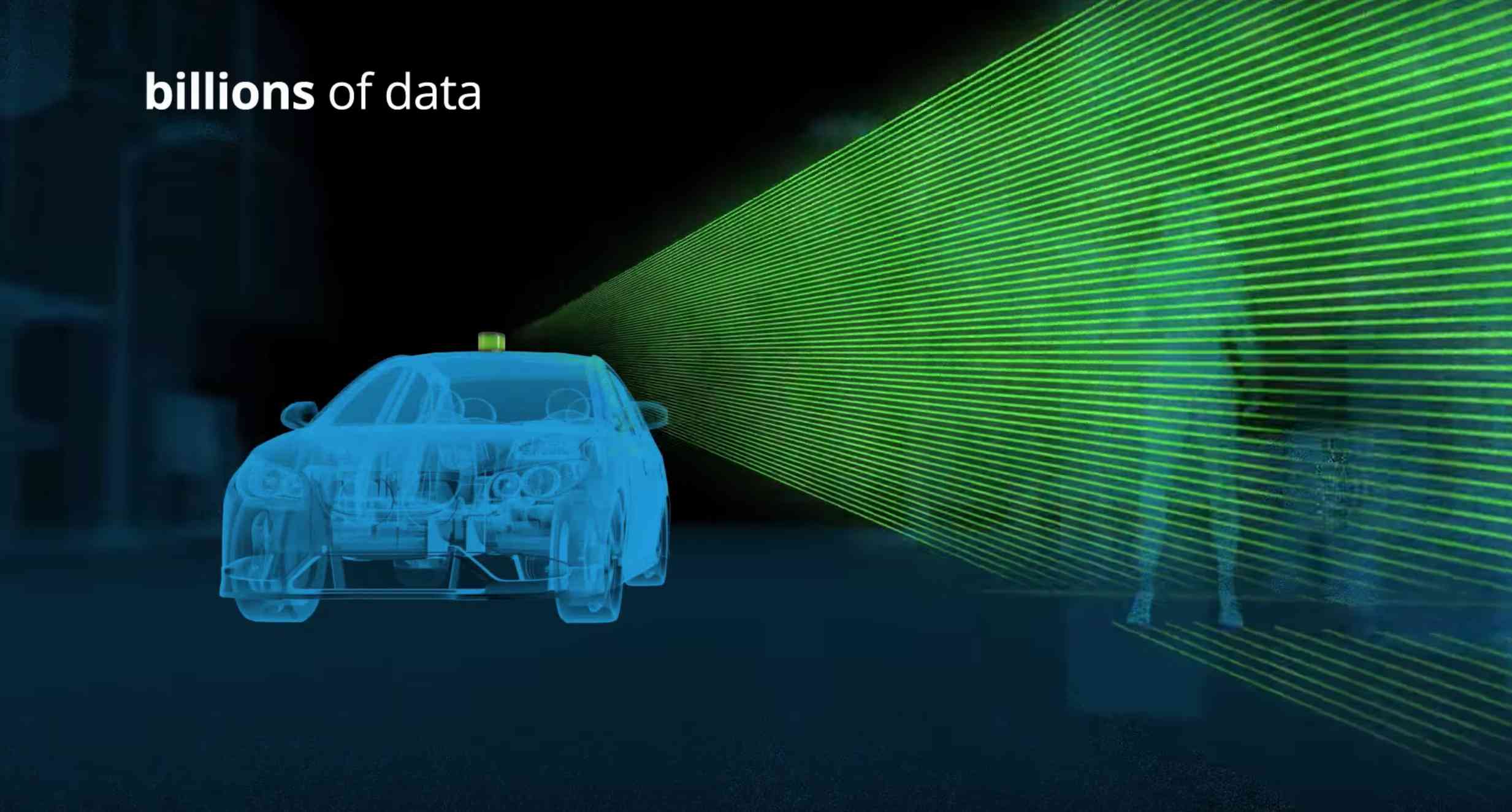}
\end{center}
\caption{The vertical scan range of LiDAR sensor for Self-driving cars. The photo is collected from  \cite{velodyne101}}
 \label{velodyne_v}
 \end{figure}
 
  \begin{figure}[H]
  \begin{center}
  %0.65
      \includegraphics[width=0.75\linewidth]{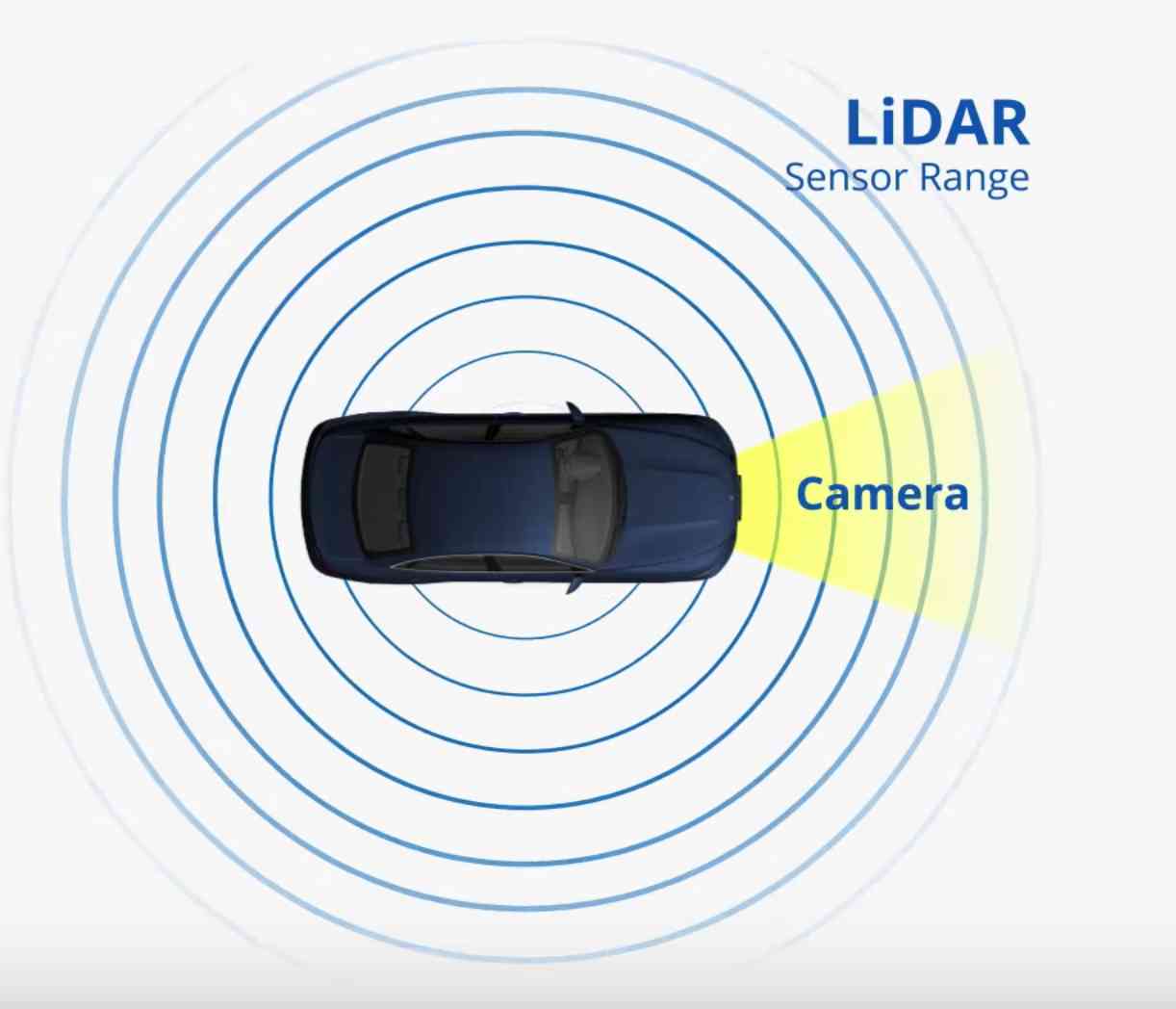}
\end{center}
\caption{The horizontal scan range of LiDAR sensor for Self-driving cars. The photo is collected from  \cite{velodyne101} }
 \label{velodyne_h}
 \end{figure}

    \begin{figure}[H]
  \begin{center}
  %0.6
      \includegraphics[width=1.0\linewidth]{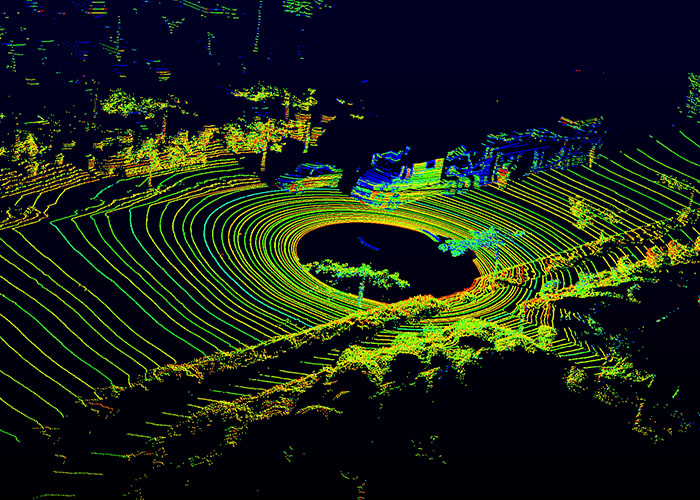}
\end{center}
\caption{Raw data from Velodyne LiDAR HDL-64E. The photo is collected from Velodyne's official website \cite{velodyne64-data}.}
 \label{fig:raw_data}
 \end{figure}

 Velodyne's HDL-64E LiDAR sensor is commonly used for Self-driving cars. There is a new LiDAR sensor with 128 laser beams released from Velodyne in 2017. The photos of both the 64 beams sensor and the 128 beams sensors is given in Figure \ref{velodyne}. The sensor's vertical scan range is shown in Figure \ref{velodyne_v} while the horizontal scan range is shown in Figure \ref{velodyne_h}. Raw data collected by Velodyne's HDL-64E LiDAR sensor is shown in Figure \ref{fig:raw_data}. \\

%For the 3D computer vision, the data representation is different with the 2D computer vision. \\

\subsection{Classification}

The classification in 3D is mainly based on the CAD(Computer-aided design) model. One import dataset is ModelNet\cite{DBLP:journals/corr/WuSKTX14}. ModelNet has 127915 3D CAD models from 662 categories. ModelNet10 and ModelNet40 are two subsets of the ModelNet. ModelNet10 has 4899 models from 10 categories and ModelNet40 has12311 models from 40 categories. For the classification task, the papers will be discussed based on the leaderboard of the ModelNet40. The leaderboard is given next.\\

\subsubsection{ModelNet40 leaderboard}
\begin{table}[H]
\scriptsize
\begin{center}
\begin{tabular}{|c|c|}
\hline
Algorithm  &ModelNet40 Classification (Accuracy)\\
\hline
RotationNet\cite{DBLP:journals/corr/QiSMG16}&	97.37$\%$\\ \hline
PANORAMA-ENN \cite{SFIKAS2018208}  &	95.56$\%$\\ \hline
VRN Ensemble\cite{DBLP:journals/corr/BrockLRW16} &	95.54$\%$\\ \hline
Wang et al. \cite{Wang2017DominantSC}   &	93.80$\%$\\ \hline
SO-Net\cite{2018arXiv180304249L} &	93.40$\%$\\ \hline
Kd-Net\cite{DBLP:journals/corr/KlokovL17} &	91.80$\%$\\ \hline
3DmFV-Net\cite{DBLP:journals/corr/abs-1711-08241}  &	91.60$\%$\\ \hline
MVCNN-MultiRes \cite{DBLP:journals/corr/QiSNDYG16} &	91.40$\%$\\ \hline
FusionNet \cite{DBLP:journals/corr/HegdeZ16} &90.80$\%$\\ \hline
PANORAMA-NN \cite{PANORAMA}&90.70$\%$\\ \hline
Pairwise\cite{DBLP:journals/corr/JohnsLD16} &	90.70$\%$\\ \hline
3D-A-Nets\cite{DBLP:journals/corr/abs-1711-10108}  &	90.50$\%$\\ \hline
MVCNN\cite{su15mvcnn}  &	90.10$\%$\\ \hline
Set-convolution\cite{2016arXiv161104500R}  &	90$\%$\\ \hline
Minto et al.\cite{visapp18} &	89.30$\%$\\ \hline
PointNet\cite{DBLP:journals/corr/QiSMG16} &	89.20$\%$\\ \hline
3DShapeNets\cite{DBLP:journals/corr/WuSKTX14} &	77$\%$\\ \hline
\end{tabular}
\end{center}
\caption{ModelNet40 leaderboard. Top 16 as of April 4, 2018 and the baseline 3DShapeNets are shown here. For more results please visit \cite{modelnet}}
\label{model_leaderboard}
\end{table}

\subsubsection{Inputs for CAD classification}

\begin{table}[H]
\scriptsize
\begin{center}
\scalebox{1.0}{
\begin{tabular}{|c|c|c|c|c|}
\hline
Algorithm  &Accuracy&input&$\#$views&others\\
\hline
RotationNet\cite{DBLP:journals/corr/QiSMG16}&	97.37&multiple-view &20 & \makecell{w/o upright orientation.\\ Max result based on 11 trials. \\The average accuracy\\ is 94.68}\\ \hline
PANORAMA-ENN \cite{SFIKAS2018208} &	95.56&multiple-view&3& \\ \hline
VRN Ensemble \cite{DBLP:journals/corr/BrockLRW16} &	95.54&volumetric &1&\\ \hline
Wang et al. \cite{Wang2017DominantSC} &	93.80&volumetric&12&\\ \hline
SO-Net\cite{2018arXiv180304249L} &	93.40& points&1&\\ \hline
Kd-Net\cite{DBLP:journals/corr/KlokovL17} &	91.80&points &1&\\ \hline
3DmFV-Net\cite{DBLP:journals/corr/abs-1711-08241}  &	91.60&\makecell{3D Modified\\ Fisher Vectors} &1&\\ \hline
MVCNN-MultiRes \cite{DBLP:journals/corr/QiSNDYG16} &	91.40&  \makecell{both volumetric\\ and multiple-view} &&\\ \hline
FusionNet\cite{DBLP:journals/corr/HegdeZ16}  &90.80&\makecell{both volumetric\\ and multiple-view} &&\\ \hline
PANORAMA-NN \cite{PANORAMA}&90.70&PANORAMA &1&\\ \hline
Pairwise\cite{DBLP:journals/corr/JohnsLD16} &	90.70& multiple-view&12&\makecell{decomposing
an image sequence\\ into a set of image pairs}\\ \hline
3D-A-Nets\cite{DBLP:journals/corr/abs-1711-10108}  &	90.50& volumetric&1&\\ \hline
MVCNN\cite{su15mvcnn}  &	90.10&\makecell{multiple-view \\RGB}&80&with upright orientation\\ \hline
Set-convolution\cite{2016arXiv161104500R}  &	90& points&1&\\ \hline
Minto et al.\cite{visapp18} &	89.30& \makecell{both volumetric\\ and multiple-view} &&\\ \hline
PointNet\cite{DBLP:journals/corr/QiSMG16} &	89.20&points &1&\\ \hline
3DShapeNets\cite{DBLP:journals/corr/WuSKTX14} &	77&volumetric &1&\\ \hline

\end{tabular}
}
\end{center}
\caption{ModelNet40 leaderboard. Top 16 as of April 4, 2018 and the baseline 3DShapeNets are shown here. For more results please visit \cite{modelnet}}
\label{model_leader_boar_new}
\end{table}

Inputs for each system are list in Table \ref{model_leader_boar_new}. Mainly three kind of inputs are used to do the classification: multiple-view images, 3D voxel grids and point clouds. The multiple-view images and 3D voxel grids are regular input data formats which can be easily fed to a 2D or 3D CNN network. The point clouds are not in a regular format. However, recently, several works are based on this irregular format and achieve impressive results. We will introduce several of those works next.\\

\subsubsection{MVCNN\cite{su15mvcnn}}

\begin{figure}[H]
\begin{center}
%\fbox{\rule{0pt}{2in} \rule{.9\linewidth}{0pt}}
\includegraphics[width=1.0\linewidth]{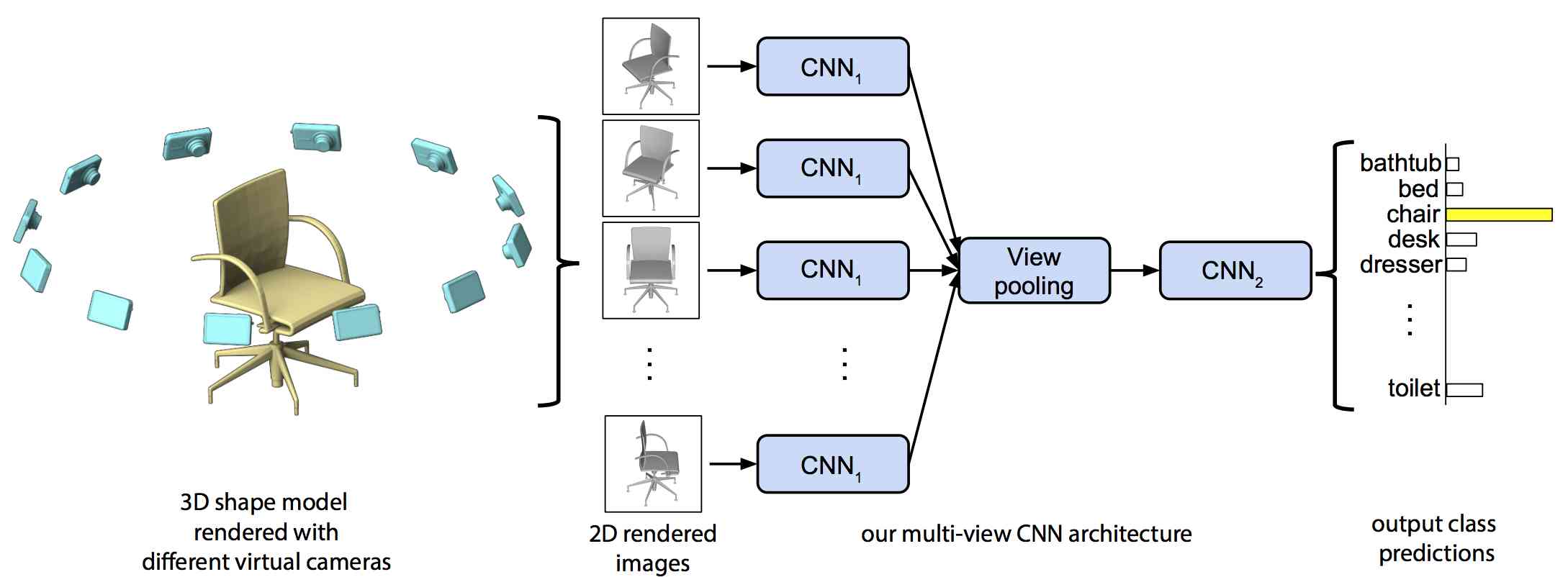}

\end{center}
   \caption{Multi-view CNN\cite{su15mvcnn} for 3D shape classification. Figure is from original paper.}
\label{fig:mvcnn}
\end{figure}

MVCNN is a powerful model regarding the performance on 3D shape classification. It is using the multiple rendered 2D images from the 3D CAD model. Features are extracted by those rendered 2D images by using a 2D CNN and the view pooling layer. Finally, based on these extracted features, it can achieve good classification results. From the results of ModelNet40 leaderboard, we can see the power of this multiple view approach.\\

\subsubsection{RotationNet\cite{DBLP:journals/corr/Kanezaki16}}

\begin{figure}[H]
\begin{center}
%\fbox{\rule{0pt}{2in} \rule{.9\linewidth}{0pt}}
\includegraphics[width=1.0\linewidth]{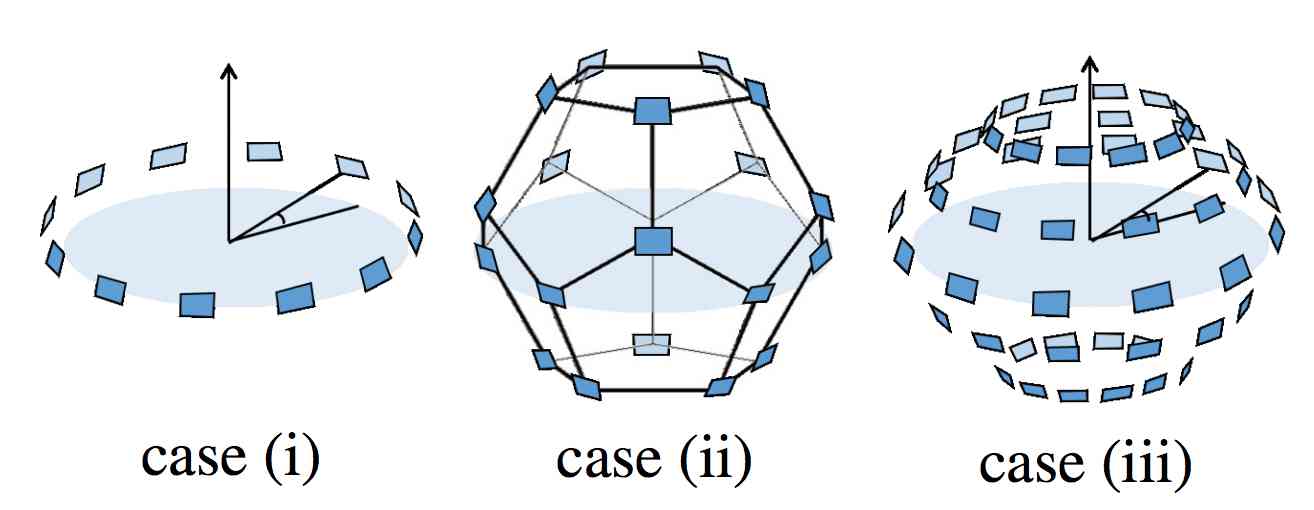}

\end{center}
   \caption{Illustration of three viewpoint setups considered in RotationNet\cite{DBLP:journals/corr/Kanezaki16}. A target object is placed on the center of each circle. Figure is from original paper.}
\label{fig:1211}
\end{figure}

\textit{Novel view point models}\\

RotationNet is an extension of MVCNN\cite{su15mvcnn}. In this paper, multiple views from different angles are explored. Three models of camera views are proposed as shown in Figure \ref{fig:1211}. The performance of case(i) (the same view points model as MVCNN\cite{su15mvcnn}) and case(ii) are compared. Case (ii) achieves a better performance based on the ModelNet40 task. For the ModelNet40, the case(iii) model is not used.\\

\textit{Unsupervised view point estimation}\\

RotationNet is using an unsupervised approach when using the view point information: the view point is not provided directly during the training process (the view point are also not available in most dataset such as ModelNet40) but inferred during the training process. The authors of RotationNet use the following optimization method to find the possible viewpoints.\\
RotationNet is defined as a differentiable multi-layer neural network $R(·)$. The final layer of RotationNet is the concatenation of $M$ softmax layers, each of which outputs the category likelihood $P(\hat{y_i} ,\mathbf{x_i},v_i =j )$ where $j \in\{1, ..., M\}$ for each image $\mathbf{x_i}$. Here, $\hat{y_i}$ denotes an estimate of the object category label for $\mathbf{x_i}$. For the training of RotationNet,  the set of images $\mathbf{x_i}_{i=1}^M $ are used simultaneously and the following optimization problem is solved to get the estimated viewpoints\cite{DBLP:journals/corr/Kanezaki16}:

\begin{equation}
\max \limits_{R,\{V_i\}_{i=1}^M } \prod_{i=1}^{M}  P(\hat{y_i} = y | \mathbf{x_i},v_i)
\end{equation}
The parameters of $R$ and latent variables $\{V_i\}_{i=1}^M$ are optimized to output the highest probability of $y$ for the input of multi-view images $\mathbf{x_i}_{i=1}^M$.\\

\textit{RotationNet framework and the training process}\\

\begin{figure}[H]
\begin{center}
%\fbox{\rule{0pt}{2in} \rule{.9\linewidth}{0pt}}
\includegraphics[width=1.0\linewidth]{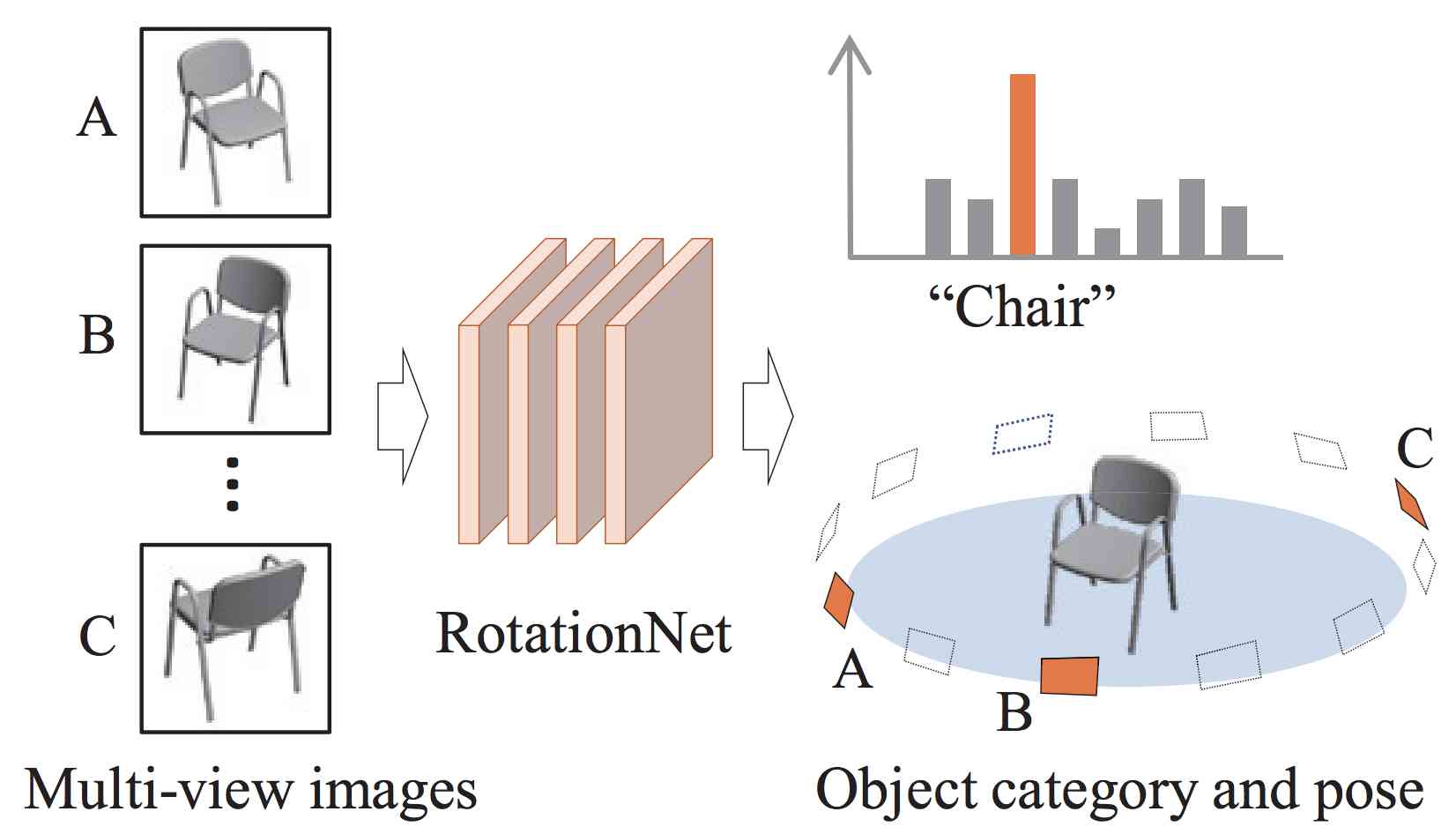}

\end{center}
   \caption{ Illustration of the proposed method RotationNet. RotationNet
takes a partial set ($\geq 1$ images) of the full multi-view
images of an object as input and predicts its object category by
rotation, where the best pose is selected to maximize the object
category likelihood. Here, viewpoints from which the images are
observed are jointly estimated to predict the pose of the object. Figure and Caption are from original paper.}
\label{fig:1222}
\end{figure}

\begin{figure}[H]
\begin{center}
%\fbox{\rule{0pt}{2in} \rule{.9\linewidth}{0pt}}
\includegraphics[width=1.0\linewidth]{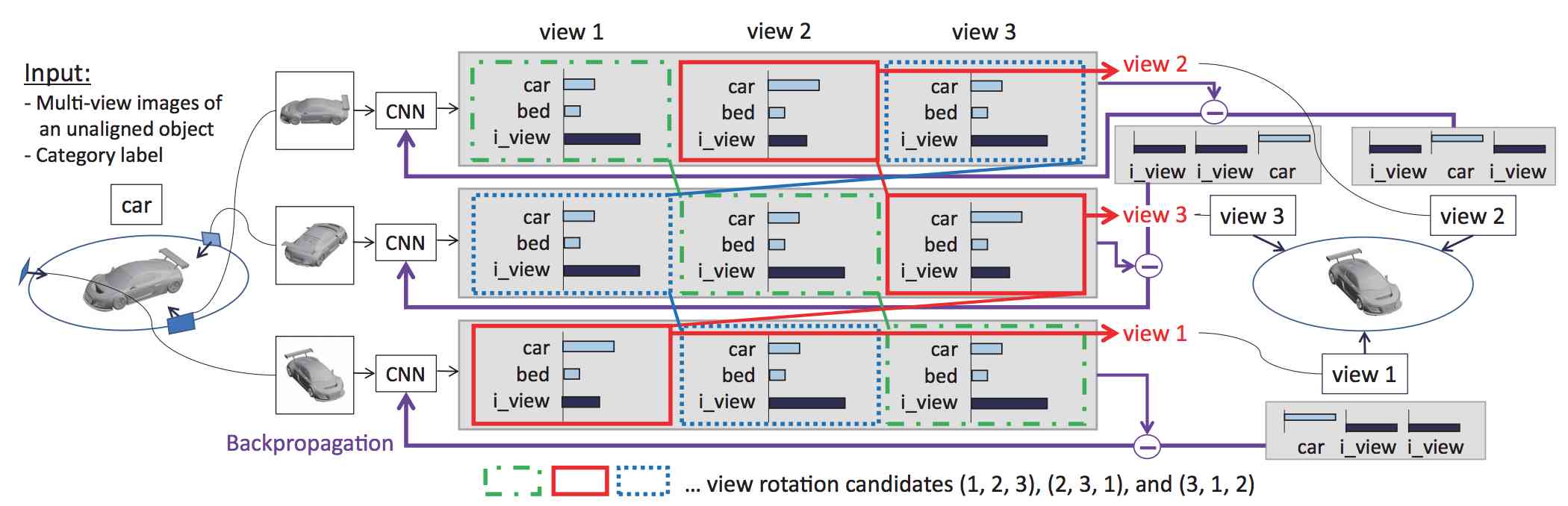}

\end{center}
   \caption{Illustration of the training process of RotationNet, where the number of views M is 3 and the number of categories N is 2. A
training sample consists of M images of an unaligned object and its category label y. For each input image, the CNN (RotationNet) outputs
M histograms with $N + 1$ bins whose norm is 1. The last bin of each histogram represents the ``incorrect view" class, which serves as a
weight of how likely the histogram does not correspond to each viewpoint variable. According to the histogram values, they decide which
image corresponds to views 1, 2, and 3. There are three candidates for view rotation: $(1, 2, 3), (2, 3, 1)$, and $(3, 1, 2)$. For each candidate,
they calculate the score for the ground-truth category (``car" in this case) by multiplying the histograms and selecting the best choice: $(2,
3, 1)$ in this case. Finally, they update the CNN parameters in a standard back-propagation manner with the estimated viewpoint variables.
Note that it is the same CNN that is being used. Figure and Caption are from original paper.}
\label{fig:1233}
\end{figure}

The general RotationNet framework is shown in Figure \ref{fig:1222}. The training process of RotationNet is illustrated in Figure \ref{fig:1233} where three views are provided to give the illustration. In order to obtain a stable estimation of the viewpoint by using deep neural networks, the ``incorrect view" class is introduced in the training process of the RotationNet. The ``incorrect view" will play the similar role as the ``background" class in object detection task. The RotationNet calculates $P(\hat{y_i} = y | \mathbf{x_i},v_i )$ by applying softmax functions to the $(N +1)$-
dimensional outputs where $N$ is the number of the categories.\\

\subsubsection{PointNet\cite{DBLP:journals/corr/QiSMG16}}

\textit{Applications of PointNet}\\

\begin{figure}[H]
\begin{center}
%\fbox{\rule{0pt}{2in} \rule{.9\linewidth}{0pt}}
\includegraphics[width=1.0\linewidth]{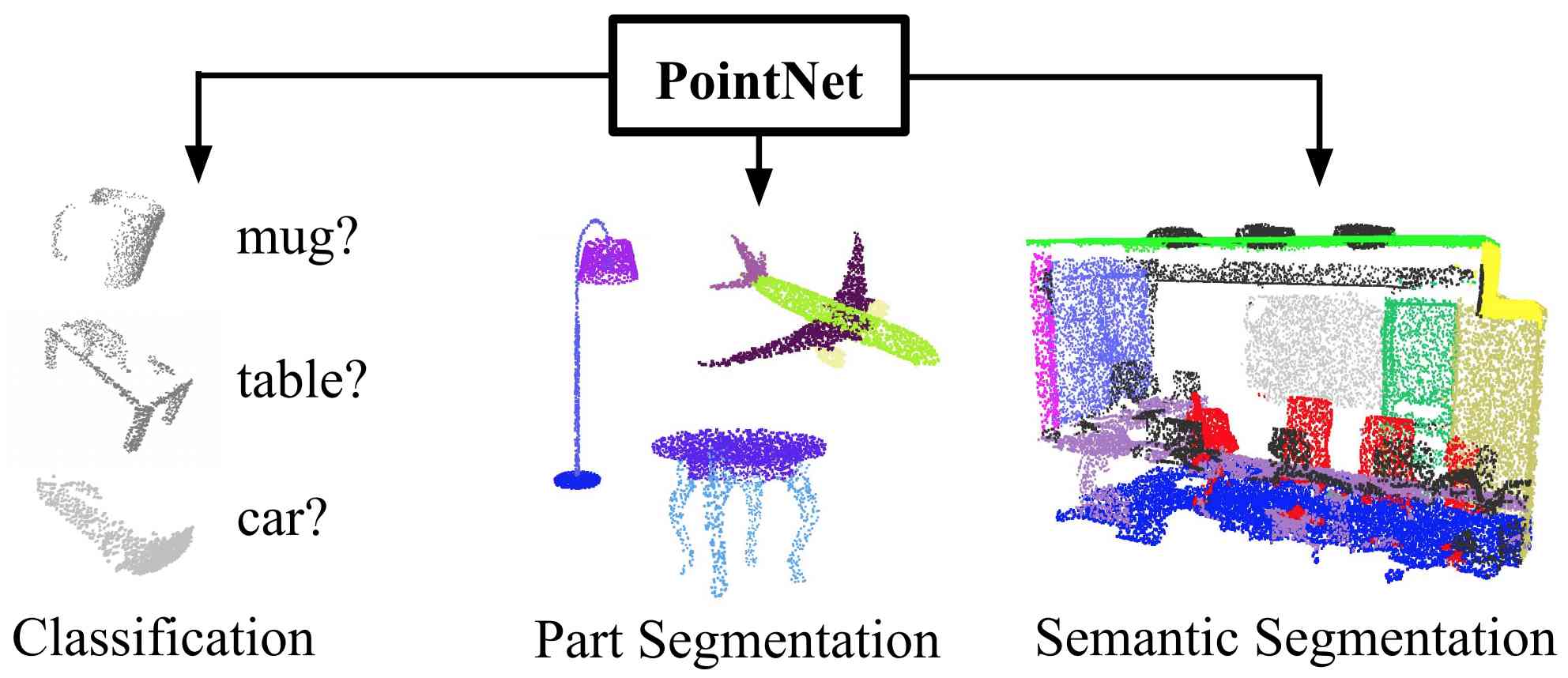}

\end{center}
   \caption{Applications of PointNet. Figure is from \cite{DBLP:journals/corr/QiSMG16}.}
\label{fig:911}
\end{figure}

PointNet is a new work which is performing 3D vision understanding directly on the raw cloud point data. Applications of PointNet are shown in Figure \ref{fig:911}.  PointNet consumes raw point clouds (set of points) without voxelization or rendering. It is a unified architecture that learns both global and local point features, providing a simple, efficient and effective approach for a number of 3D classification tasks. \\

\textit{PointNet Architecture}\\

\begin{figure}[H]
\begin{center}
%\fbox{\rule{0pt}{2in} \rule{.9\linewidth}{0pt}}
\includegraphics[width=1.0\linewidth]{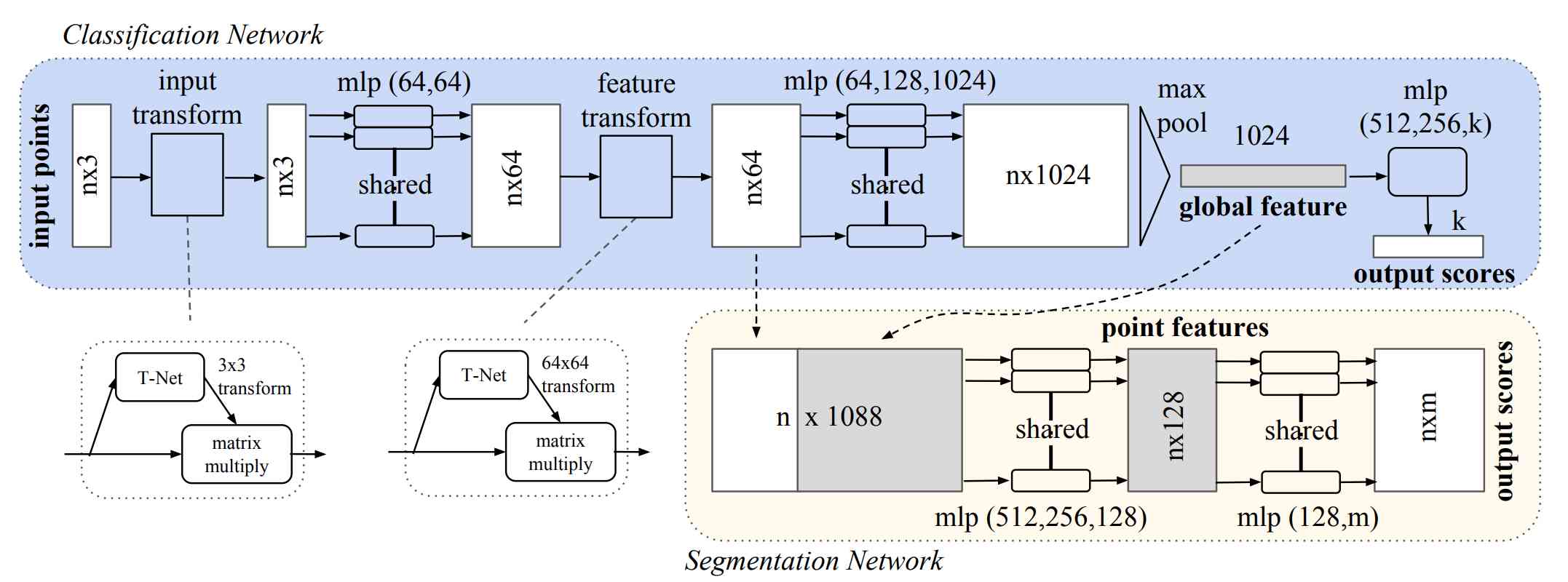}

\end{center}
   \caption{PointNet Architecture. The figure is from\cite{DBLP:journals/corr/QiSMG16}. Here the ``mlp" means multi-layer perceptron.}
\label{fig:922}
\end{figure}

PointNet Architecture is shown in Figure \ref{fig:922}. The classification network takes $n$ points as input, applies input and feature transformations, and then aggregates point features by max pooling. The output is classification scores for $k$ classes. The segmentation network is an extension to the
classification net. It concatenates global and local features and outputs per point segmentation result. \\

The structure of classification network is summarized as follows.\\

\begin{enumerate}[label=\arabic*)]	\item takes $n$ points as input. Each point has $x, y, z$ three coordinate values. An input transform will be done based on the $n \times 3$ input data:
      \begin{enumerate}[label=\alph*)]
    	\item T-Net provides a $3\times 3$ transformation matrix
        \item matrix multiplication between the $n \times 3$ input data and the $3\times 3$ transformation matrix is done to finish the \textbf{input transform}
    \end{enumerate}
    \item Transformed $n \times 3$ data is passed to a $64\times 64$ multi-layer perceptron(MLP) network to output a $n \times 64$ tensor
    \item The $n \times 64$ tensor is fed to a \textbf{feature transform} net:
    \begin{enumerate}
    	\item T-Net provides a $64\times 64$ transformation matrix
        \item matrix multiplication between the $n \times 64$ input data and the $64\times 64$ transformation matrix is done to finish the feature transform
    \end{enumerate}
     \item Transformed $n \times 64$ data is passed to a $ 64\times 128\times 1024$ MLP  to output a $n \times 1024$ tensor
    \item Global feature is generated by an element wise max pooling layer and output a  $1 \times 1024$ tensor
    \item Feeding the  $1 \times 1024$ global feature to another $512  \times 256 \times k$ MLP to generate a classification vector with $ k$ values(corresponding to $k$ class labels).
\end{enumerate}

The segment network structure is as follows.\\

\begin{enumerate}[label=\arabic*)]
	\item the $1 \times 1024$ global features are concatenated to every $1 \times 64$ point features. As the system totally has $n$ points, the concatenation generates a $n \times 1088$ tensor.
	  \item the $n \times1088 $tensor is fed to a $512\times 256 \times 128$ MLP to further generate a $n \times 128$ tensor.
    \item the $n \times 128$ tensor is then fed to a $128\times m $ to generate a $n \times m$ tensor (corresponding to $m$ segmentation labels for each point)
   \end{enumerate}

Batch Normalization\cite{DBLP:journals/corr/IoffeS15} is used for all layers with ReLU\cite{Nair:2010:RLU:3104322.3104425}. Dropout layers are used for the last MLP in the classification network.\\

\textit{T-Net in PointNet}\\

For the semantic labelling task for a point cloud, the output is supposed to be invariant if the point cloud undergoes certain geometric transformations, such as rigid transformation. In order to achieve a transformation invariant learning representation for the raw cloud point, a T-Net is introduced in PointNet as shown in Figure \ref{fig:922}. The main functionality of this T-Net is predicting an affine transformation matrix.\\

\textit{Performance of PointNet}\\

The results of the PointNet on ModelNet40 are shown in Table \ref{model_leader_boar_new}.\\

\textit{More results of PointNet}\\

\begin{figure}[H]
\begin{center}
%\fbox{\rule{0pt}{2in} \rule{.9\linewidth}{0pt}}
\includegraphics[width=1.0\linewidth]{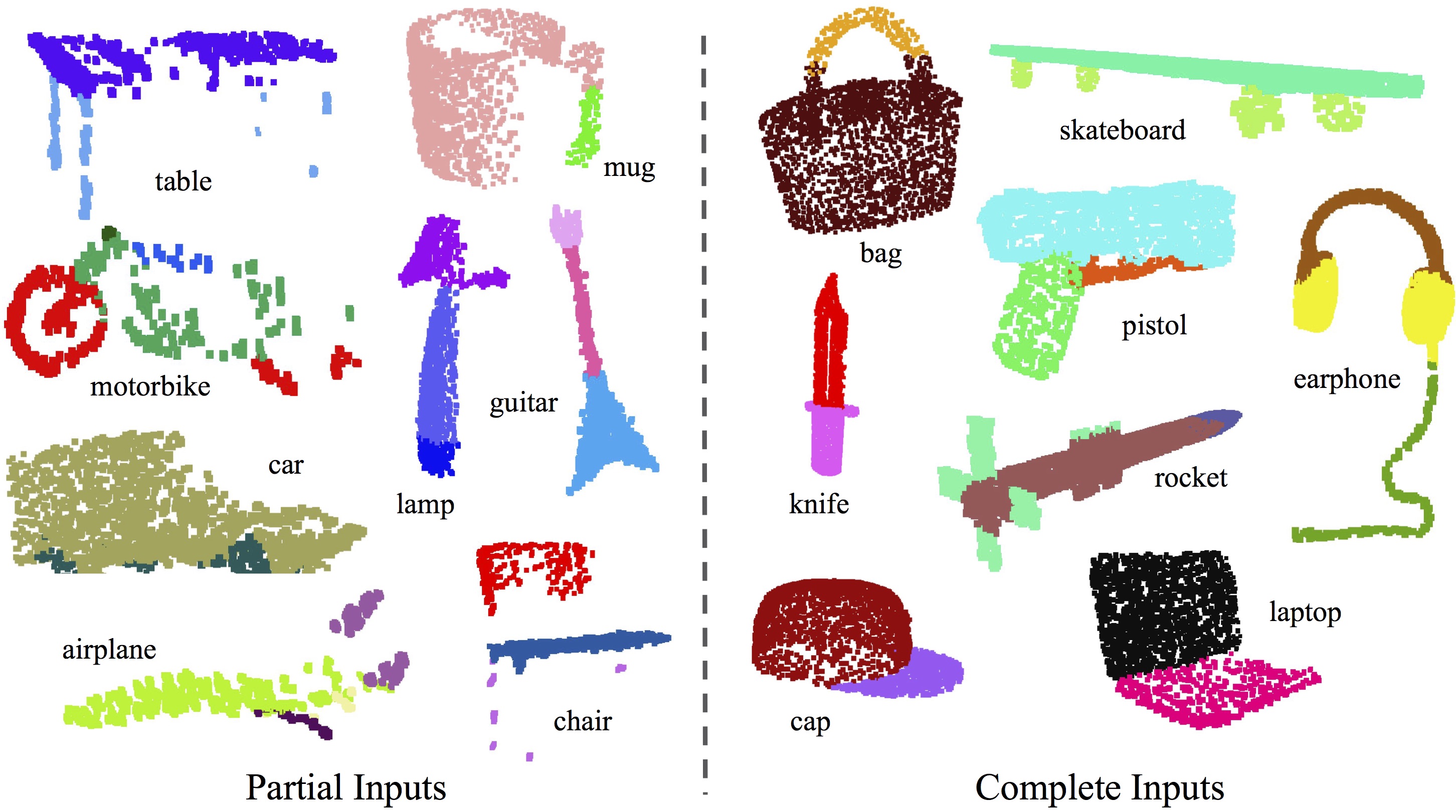}

\end{center}
   \caption{Part Segmentation Results\cite{DBLP:journals/corr/QiSMG16}. The CAD part segmentation results are visualized across all 16 object categories. Both results for partial simulated Kinect scans (left block) and complete ShapeNet CAD models (right block) are shown. Figure and Caption are adjusted from original paper.}
\label{segres}
\end{figure}

\begin{figure}[H]
\begin{center}
%\fbox{\rule{0pt}{2in} \rule{.9\linewidth}{0pt}}
\includegraphics[width=1.0\linewidth]{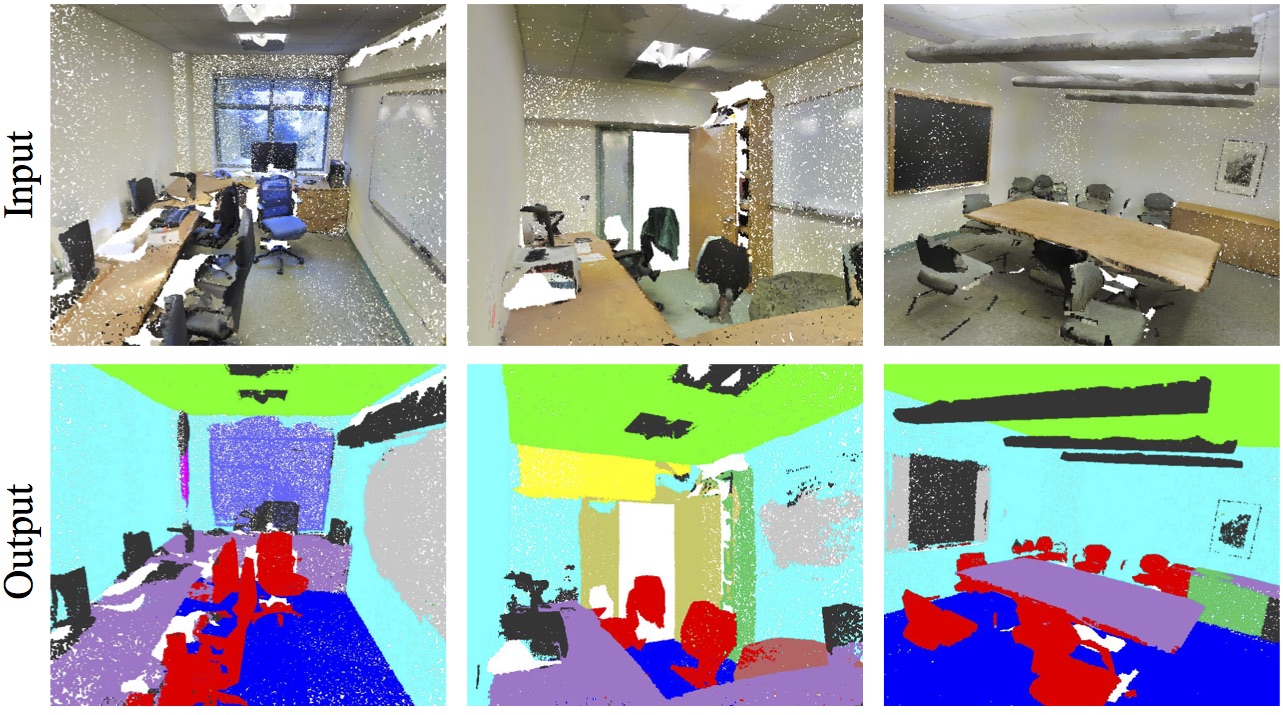}

\end{center}
   \caption{Semantic Segmentation Results. Top row is input point cloud with color. Bottom row is output semantic segmentation result (on points) displayed in the same camera viewpoint as input. Figure and Caption are from \cite{DBLP:journals/corr/QiSMG16}.}
\label{semantic}
\end{figure}

Part segmentation and semantic segmentation results are shown in Figure \ref{segres} and \ref{semantic}. \\

\begin{figure}[H]
\begin{center}
%\fbox{\rule{0pt}{2in} \rule{.9\linewidth}{0pt}}
\includegraphics[width=0.85\linewidth]{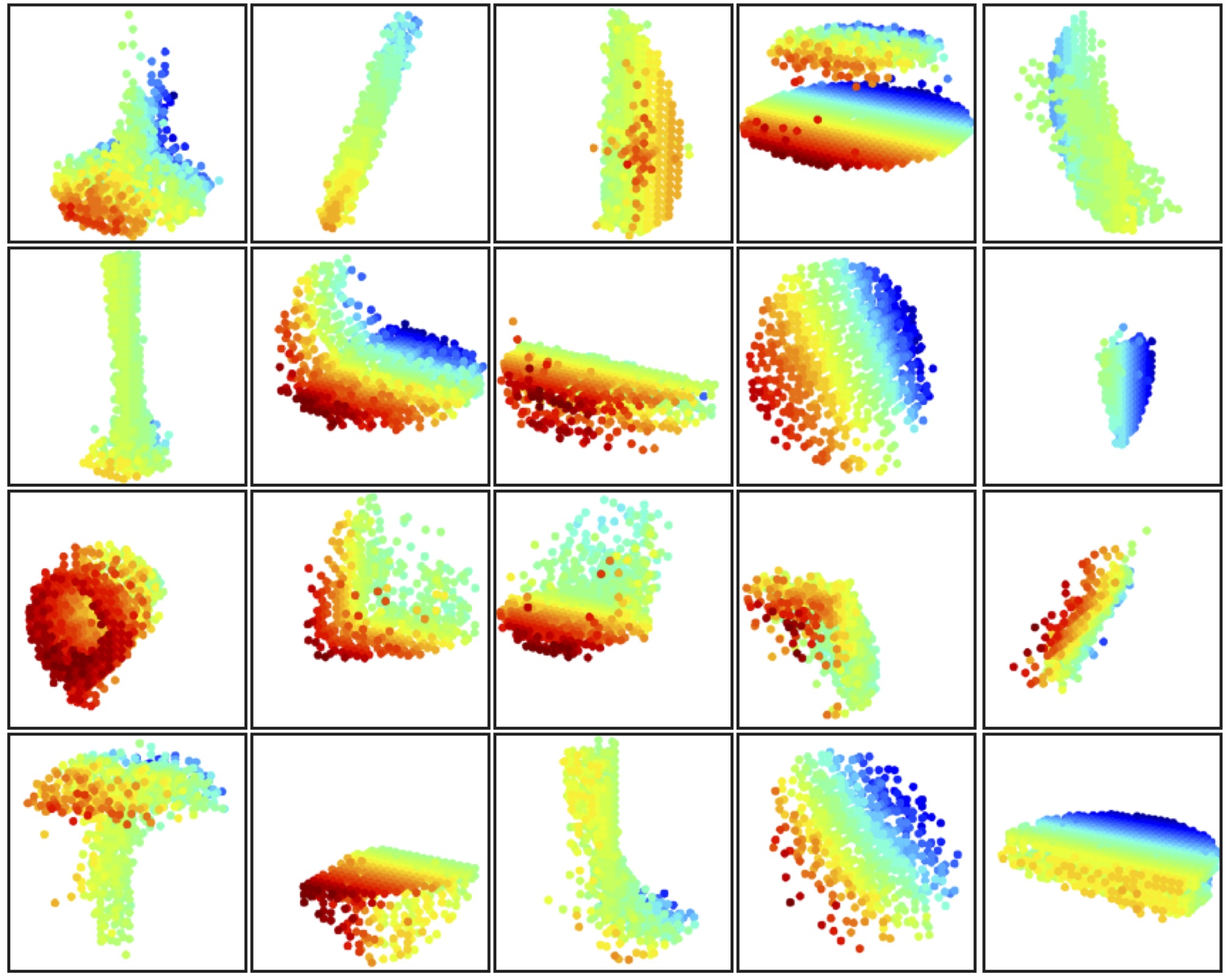}

\end{center}
   \caption{3D point cloud patterns learned from the first layer kernels. The model is trained for ModelNet40 shape classification(20 out of the 128 kernels are randomly selected). Color indicates point depth (red is near, blue is far). Figure and Caption are from \cite{DBLP:journals/corr/QiYSG17}.}
\label{kernels}
\end{figure}

\begin{figure}[H]
\begin{center}
%\fbox{\rule{0pt}{2in} \rule{.9\linewidth}{0pt}}
\includegraphics[width=1.0\linewidth]{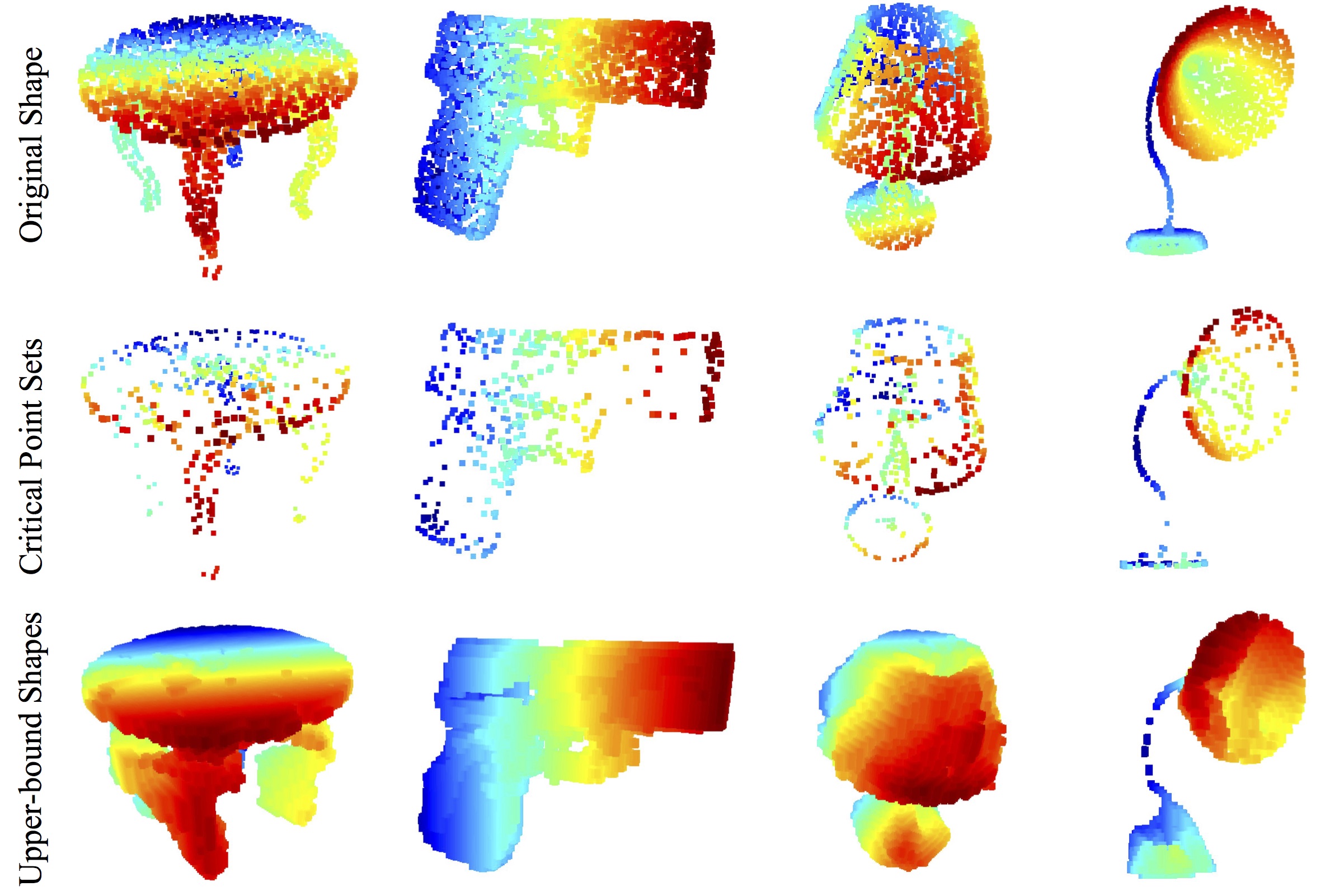}

\end{center}
   \caption{Visualizing Critical Points and Shape Upper-bound. The first row shows the input point clouds. The second row show the critical points picked by PointNet. The third row shows the upper-bound shape for the input -- any input point sets that falls between the critical point set and the upper-bound set will result in the same classification result. Figure and Caption are from \cite{DBLP:journals/corr/QiSMG16}.}
\label{fig:kp_ss_visu1}
\end{figure}
3D point cloud patterns learned from the PointNet are shown in Figure \ref{kernels}. This result is from \cite{DBLP:journals/corr/QiYSG17}, an improved  version of PointNet. \\
The critical points learned by the PointNet and Shape Upper-bound of some objects are shown in Figure \ref{fig:kp_ss_visu1}. \\

%\begin{figure}[H]
%\begin{center}
%\fbox{\rule{0pt}{2in} \rule{.9\linewidth}{0pt}}
%\includegraphics[width=1.0\linewidth]{933}

%\end{center}
   %\caption{Qualitative results for part segmentation. \cite{DBLP:journals/corr/QiSMG16} visualize the CAD part segmentation results across all 16 object categories. \cite{DBLP:journals/corr/QiSMG16} show both results for partial simulated Kinect scans (left block) and complete ShapeNet CAD models (right block).\cite{DBLP:journals/corr/QiSMG16}}
%\label{fig:933}
%\end{figure}

\subsection{Detection}
Similar to 2D-image based object systems, most 3D systems are also using the two-stage methods to do the 3d object detection: first, generate proposals and then do detection. At the same time, the unique properties of the 3D systems, such as different data representation and the availability of both 2D and 3D images, make the 3D detection framework more complicated and more interesting. We will discuss the datasets used for detection and main works in indoor and outdoor scenarios next.\\
\subsubsection{Datasets used for 3D object detection}
\begin{figure}[H]
\begin{center}
%\fbox{\rule{0pt}{2in} \rule{.9\linewidth}{0pt}}
\includegraphics[width=1.0\linewidth]{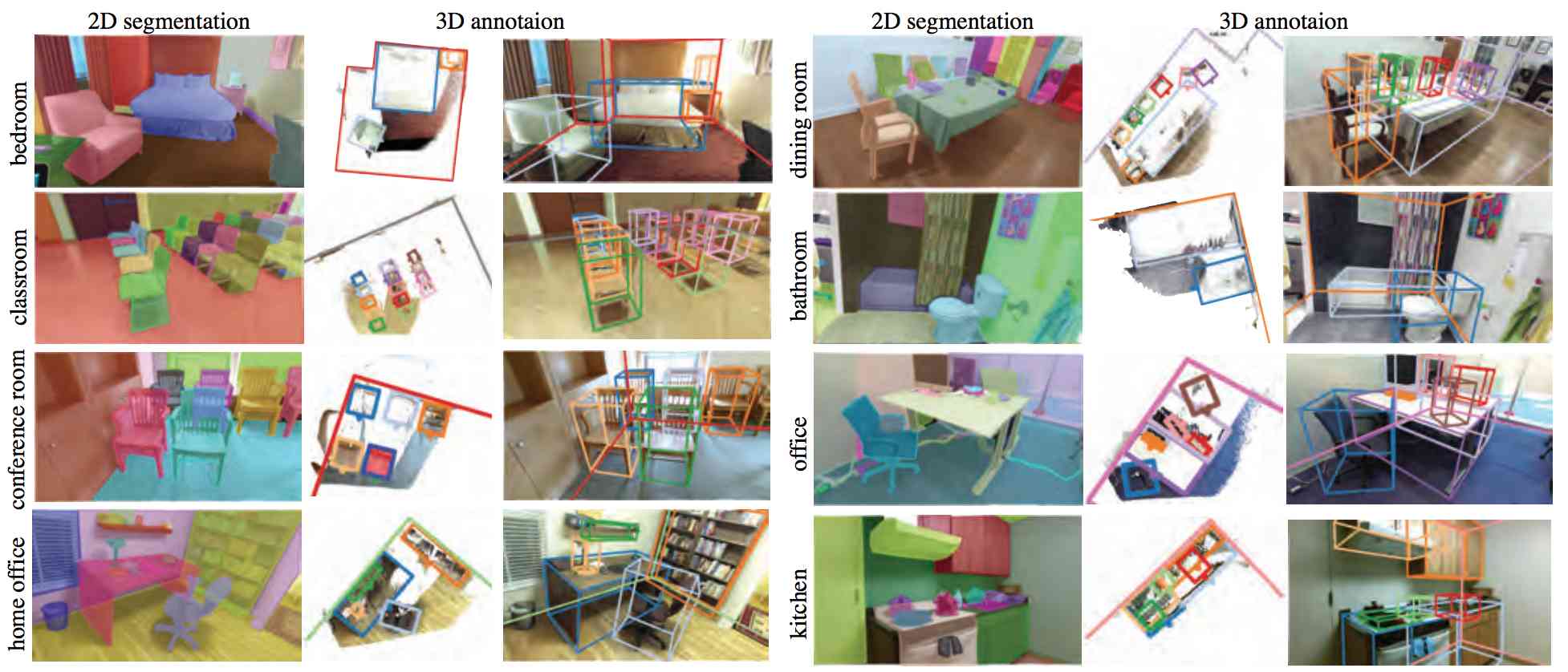}

\end{center}
   \caption{Example images with annotation from the SUN-RGBD dataset\cite{Song_2015_CVPR}.}
\label{sunrgb}
\end{figure}

Several datasets are used as the test bed of different detection algorithms. \textbf{NYU Depth Dataset V2 \cite{Silberman:ECCV12}} has 1449 densely labeled pairs of aligned RGB and depth images from Kinect video sequences for a variety of indoor scenes. \\
\textbf{SUN-RGBD dataset\cite{Song_2015_CVPR}} has 19 object categories for predicting a 3D bounding box in real world dimension. It has 10,355 RGB-D images for training set and 2860 for testing set. The whole dataset is densely annotated and includes
146,617 2D polygons and 64,595 3D bounding boxes with accurate object orientations. Several examples of the SUN-RGBD dataset is shown in Figure \ref{sunrgb}.\\
 \textbf{KITTI\cite{Geiger2012CVPR}} 3D object detection dataset consists of 7481 training images and 7518 test images as well as the corresponding point clouds, comprising a total of 80,256 labeled objects. The result of 3D object detection performance is evaluated  by using the PASCAL criteria. Far objects are thus filtered based on their bounding box height in the \text{image plane}. Since only objects appearing on the image plane are labeled, objects in non car areas do not count as false positives. For cars a 3D bounding box overlap of 70$\%$ is required, while for pedestrians and cyclists the requirement is 50$\%$. Difficulties are defined as follows:\\

Easy: minimum bounding box height: 40 Pixel, Maximum occlusion level: Fully visible, Maximum truncation: 15 $\%$\\
Moderate: minimum bounding box height: 25 Pixel, Maximum occlusion level: Partly occluded, Maximum truncation: 30 $\%$\\
Hard: minimum bounding box height: 25 Pixel, Maximum occlusion level: Difficult to see, Maximum truncation: 50 $\%$\\

\begin{figure}[H]
\begin{center}
%\fbox{\rule{0pt}{2in} \rule{.9\linewidth}{0pt}}
\includegraphics[width=0.65\linewidth]{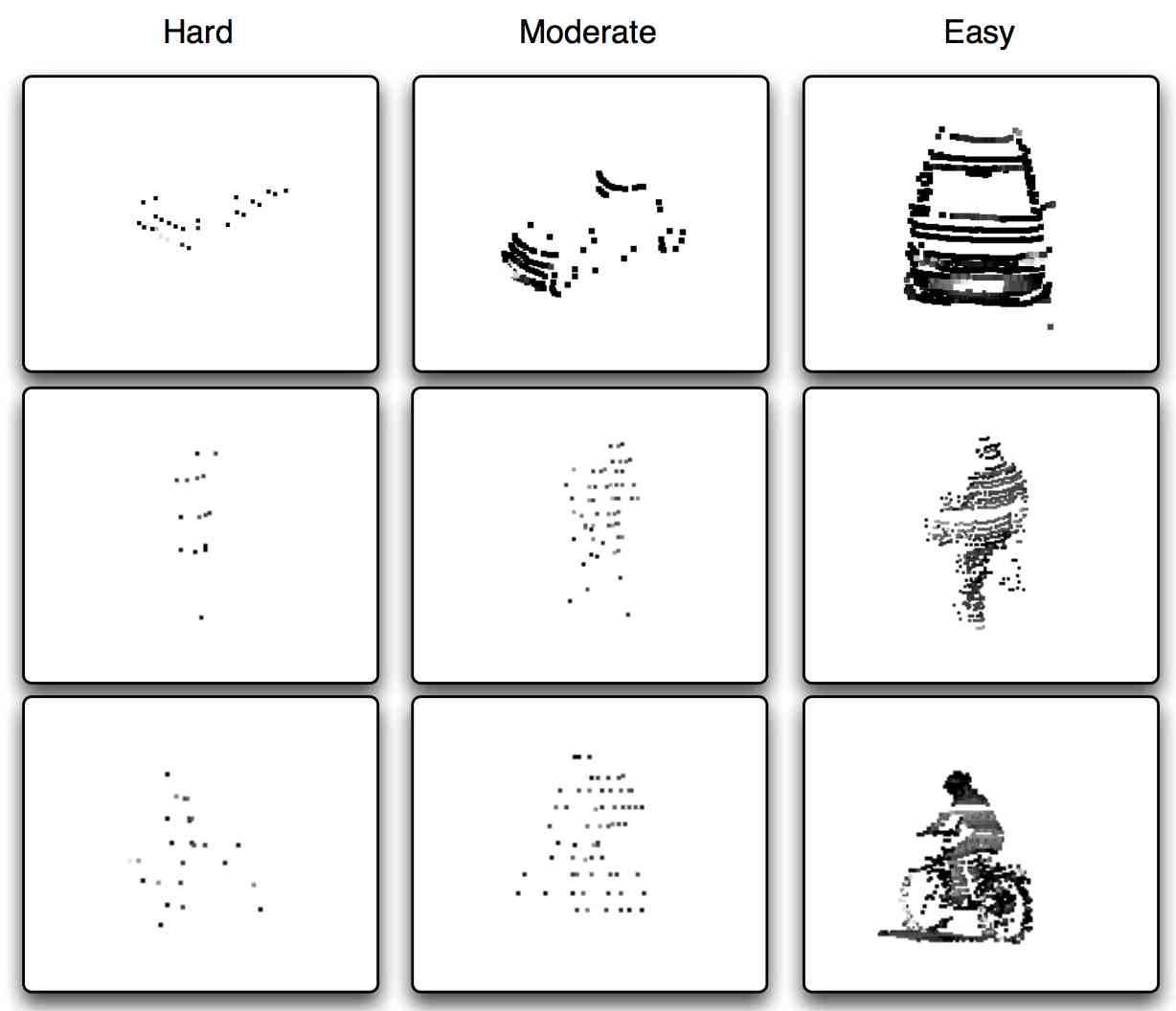}

\end{center}
   \caption{Examples of labelled object instances from the training set of
different difficulties. Left column: hard. Middle column: moderate. Right
column: easy. The figure is from \cite{DBLP:conf/rss/WangP15}.}
\label{fig:example}
\end{figure}

Examples of the labelled object instances from the training set of different difficulties are shown in Figure \ref{fig:example}

\subsubsection{Detection outputs}

\begin{table}[H]
\scriptsize
\begin{center}
\begin{tabular}{|c|c|}
\hline
\textbf{Type}& \textbf{output}\\
 \hline
\multirow{2}{*}{
Class} & label\\
         & confidence value\\
 \hline
\multirow{4}{*}{ Bounding Box}&Image plane BBOX\\
                       &BEV BBOX\\
                       &3D BBOX\\
                        \hhline{~-}
                       & orientation\\
\hline
\end{tabular}
\end{center}
\caption{Different detection outputs based on 3D image data}
\label{detection_outputs}
\end{table}

\begin{figure}[H]
\begin{center}
%\fbox{\rule{0pt}{2in} \rule{.9\linewidth}{0pt}}
\includegraphics[width=1.0\linewidth]{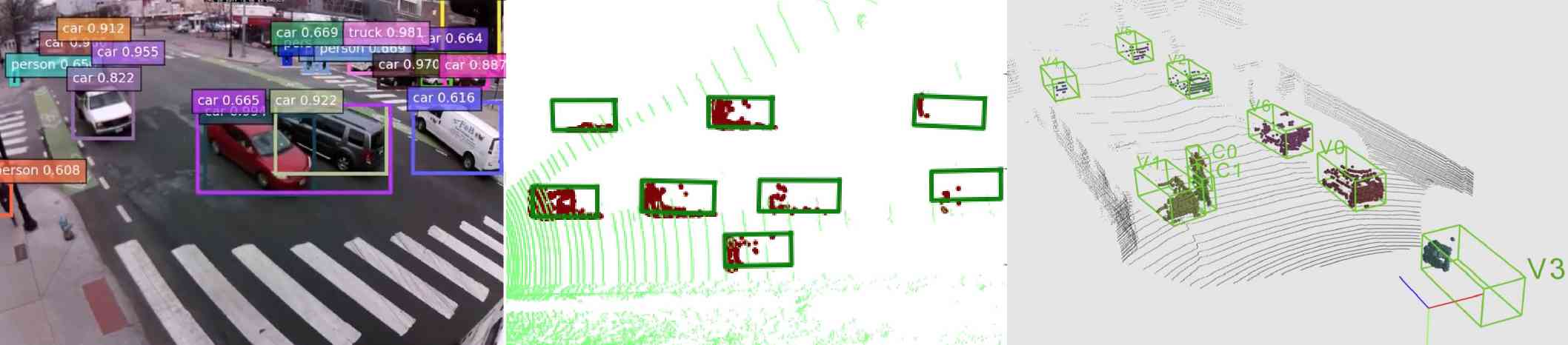}
\end{center}
   \caption{Examples for the bounding box for each detection task: left is the detection bounding box for the image plane, middle is the bbox for the BEV and right is 3D detection bounding box.}
\label{fig:image_plane_bev_3d}
\end{figure}

For 3D understanding, the detection output is more complicated than the 2D case. It contains the output of the class and the bounding box. The class output is similar to the 2D detection task. For the bounding box detection, it can output the image plane bounding box, bird's eye view(BEV) bounding box and 3D bounding box.  Meanwhile, for the 2D object detection, the bounding boxes are axis aligned, however, the BEV and 3D bounding boxes are not axis aligned. Finally, the orientation of BEV and 3D bounding box will also be detected in some tasks. The detection outputs are summarized in Table \ref{detection_outputs}. Examples of each output are shown in Figure \ref{fig:image_plane_bev_3d}. In this survey, we mainly focus on the 3D bounding box detection.\\

\subsubsection{Comparison by number of stages used}
\begin{table}[H]
\scriptsize
\begin{center}
\begin{tabular}{|c|c|}
\hline
$\#$ of stage & System\\
\hline
\multirow{4}{*}{Two-stage} &Deep Sliding Shape\cite{DBLP:journals/corr/SongX15}\\
\hhline{~-}
& MV3D\cite{DBLP:journals/corr/ChenMWLX16}	\\
\hhline{~-}
 & AVOD\cite{2017arXiv171202294K}	\\
\hhline{~-}
 &VoxelNet\cite{DBLP:journals/corr/abs-1711-06396}\\
\hline
 Three-stage& F-PointNet\cite{DBLP:journals/corr/abs-1711-08488}\\
\hline
\end{tabular}
\end{center}
\caption{Number of stages used for the different detection frameworks. }
\label{by_scenario}
\end{table}

Most frameworks are based on two-stage methods where the proposals are firstly generated and then the detection will be done based on the proposals. F-PointNet\cite{DBLP:journals/corr/abs-1711-08488} is a special case. It has three stages: first, get a frustum proposal from the detected 2D bounding based on the RGB image. The difference of the frustum proposal with the proposal from two-stage methods is it has a class label. second, 3D Instance segmentation is done based on the proposal. Finally, a 3D box is estimated based on the segmentation results.\\

\subsubsection{The 3D bounding box encoding methods}

\begin{figure}[H]
\begin{center}
%\fbox{\rule{0pt}{2in} \rule{.9\linewidth}{0pt}}
\includegraphics[width=1.0\linewidth]{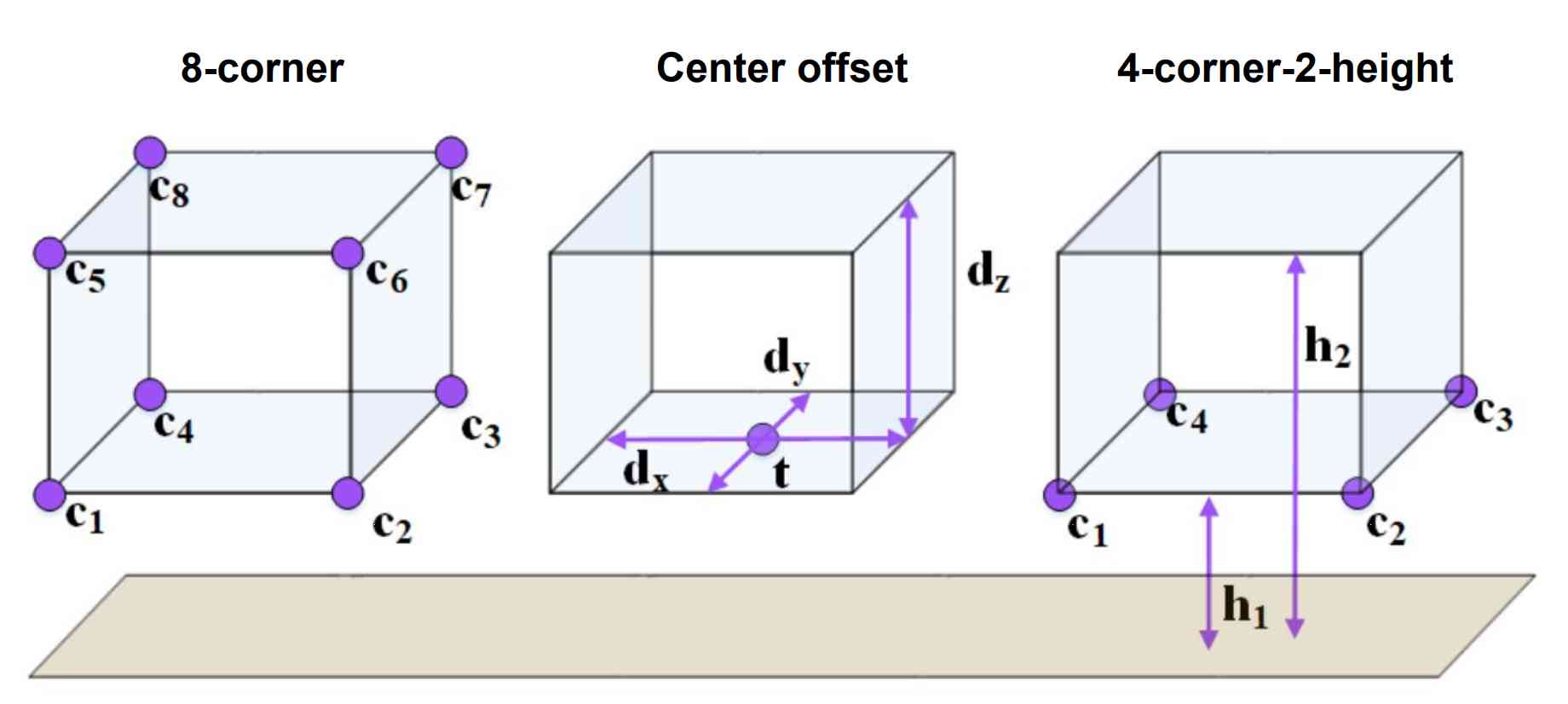}

\end{center}
   \caption{Different 3D bounding box encoding methods. The figure is adjusted from \cite{2017arXiv171202294K} }
\label{fig:bbox_encoding}
\end{figure}

%The 3D bounding box can be encoded by different methods as shown in Figure 
In order to represent a 3D bounding box, different methods are proposed, such as 8-corner method, 3D center offset method and 4-corner-2-height method as shown in Figure \ref{fig:bbox_encoding}.\\

\subsubsection{Comparison by input data, feature representation, BBOX encoding method and CNN kernel used}
\begin{table}[H]
\scriptsize
\begin{center}
\begin{tabular}{|c|c|}
\hline
\textbf{Input data}& \textbf{System}\\
\hline
Monocular&Mono3D\cite{cvpr16chen}\\
\hline
Stereo&3DOP\cite{NIPS2015_5644}\\
\hline
\multirow{3}{*}{Depth/LiDAR only}
&VoxelNet\cite{DBLP:journals/corr/abs-1711-06396}\\
%\hhline{~-}
& MV3D\cite{DBLP:journals/corr/ChenMWLX16}	\\
&F-VoxNet v1/v2\cite{Shen_2020_WACV}\cite{Shen_2020_v2}\\
\hline
\multirow{5}{*}{Monocular$+$ Depth/LiDAR}
&Deep Sliding Shape\cite{DBLP:journals/corr/SongX15}\\
%\hhline{~-}
& MV3D\cite{DBLP:journals/corr/ChenMWLX16}\\
%\hhline{~-}
& AVOD\cite{2017arXiv171202294K}	\\
%\hhline{~-}
& F-PointNet\cite{DBLP:journals/corr/abs-1711-08488}\\
&F-VoxNet v1/v2\cite{Shen_2020_WACV}\cite{Shen_2020_v2}\\
\hline
Stereo$+$ Depth/LiDAR&None\\
\hline
\end{tabular}
\end{center}
\caption{Input data for the whole system of different 3D bounding box detection systems. }
\label{by_input_data}
\end{table}

Different input data can be used to detect the 3D bounding box, such as the Monocular image, Stereo image and depth/LiDAR image. The detection system organized by the input data type is given in Table \ref{by_input_data}. Generally speaking, the 2D image only system including both the monocular and stereo image will perform worse than the 3D only or 2D$+$3D system. The comparison of the 2D image only and the 2D$+$3D system is provided in Figure \ref{fig:mv3d_vs_2d}. In the following section we will focus on the 3D only or 2D$+$3D detection systems.\\

%8-corner method, 3D center offset method and 4-corner-2-height 

\begin{table}[H]
\scriptsize
\begin{center}
\scalebox{0.73}{
\begin{tabular}{|c|c|c|c|c|c|c|c|c|}
\hline
\multicolumn{4}{|c|}{\textbf{3D BBOX Detection}}& \textbf{System}&\multicolumn{4}{c|}{\textbf{Proposals generation}}\\
\hline
Input data & features&\makecell{BBOX \\encoding\\ method}&\makecell{2D \\or 3D\\ cnn}&&Input data & features&\makecell{BBOX \\encoding \\method}&\makecell{2D \\or 3D\\ cnn}\\
\hline
\multirow{2}{*}{\makecell{Depth/LiDAR\\ only}}&\makecell{7-feature\\ per point}&\makecell{3D center \\offset $+\theta$}
&\makecell{3D\\+2D}&VoxelNet\cite{DBLP:journals/corr/abs-1711-06396}&LiDAR&\makecell{7-feature}&\makecell{3D center \\offset $+\theta$}&\makecell{3D\\+2D}\\
\hhline{~--------}
&\makecell{FV$+$ \\BEV}&8-corner&2D& MV3D\cite{DBLP:journals/corr/ChenMWLX16}&LiDAR&\makecell{FV$+$ \\BEV}&8-corner&2D\\
\hhline{~--------}
&depth&\makecell{3D center \\offset $+\theta$}&2D + 3D& F-VoxNet v1/v2\cite{Shen_2020_WACV}\cite{Shen_2020_v2}&depth&DHS&2D center offset&2D\\
\hline
\multirow{5}{*}{\makecell{Monocular$+$ \\Depth/LiDAR}}
&\makecell{voxel, \\+ RGB}&3D center offset &\makecell{Depth: 3D\\image:2D}&DeepSliding\cite{DBLP:journals/corr/SongX15}&\makecell{Monocular\\$+$ Depth}&\makecell{voxel, \\+RGB projected \\to each \\cloud point}&\makecell{3D center \\offset}&3D\\
\hhline{~--------}
&\makecell{FV$+$\\ BEV \\ +RGB}&8-corner&2D& MV3D\cite{DBLP:journals/corr/ChenMWLX16}&LiDAR&BEV&8-corner&2D\\
\hhline{~--------}
&\makecell{BEV \\+RGB}&\makecell{4-corner-\\2-height}&2D &AVOD\cite{2017arXiv171202294K}&LiDAR&\makecell{BEV\\$+$RGB}&\makecell{4-corner-\\2-height}&2D	\\
\hhline{~--------}
&point+image&\makecell{3D center \\offset $+\theta$}&T-NET& F-PointNet\cite{DBLP:journals/corr/abs-1711-08488}&Monocular&RGB&2D center offset&2D\\
\hhline{~--------}
&depth+image&\makecell{3D center \\offset $+\theta$}&2D + 3D& F-VoxNet v1/v2\cite{Shen_2020_WACV}\cite{Shen_2020_v2}&Monocular&RGB&2D center offset&2D\\
\hline
\end{tabular}
}
\end{center}
\caption{Input data for the whole system of different 3D bounding box detection systems. }
\label{by_input_feature_for_proposal_and_detection}
\end{table}

A comprehensive comparison of the input data, the feature representation of input data and the bounding box encoding methods for both the proposals and the final 3D bounding box detection is provided in Table \ref{by_input_feature_for_proposal_and_detection}. DeepSliding\cite{DBLP:journals/corr/SongX15} and VoxelNet\cite{DBLP:journals/corr/abs-1711-06396} are using the 3D convolutional neural network to do the proposals generation and the bounding box detection. MV3D\cite{DBLP:journals/corr/ChenMWLX16}, AVOD\cite{2017arXiv171202294K} are projecting the depth or LiDAR data to a 2D similar images and are using a 2D CNN to do the proposal generation and bounding box detection. The relationship between MV3D and AVOD is explained later. F-PointNet\cite{DBLP:journals/corr/abs-1711-08488} is using 2D RGB images to help generate the proposals and it is a special framework. At the same time, the different proposal generation method will have an influence on the application scenario of those frameworks, to be discussed later.\\

\subsubsection{Comparison by performance}
The performance comparison of the different systems is provided in Table \ref{car} , \ref{pedestrian} and \ref{cyclist} for the outdoor scenario based on the KITTI\cite{Geiger2012CVPR} dataset and in Table \ref{performance_comp_sun_rgbd} for the indoor scenario based on the SUN-RGBD dataset\cite{Song_2015_CVPR}.\\

\begin{table}[H]
\scriptsize
\begin{center}
\scalebox{0.65}{
\begin{tabular}{|c|cccccccccccc|c|c|}
\hline
&bathtub& bed &bookshelf& chair& desk &dresser& nightstand &sofa& table& toilet& Sofa Chair& Garbage Bin&Runtime &mAP\\
\hline
\makecell{DeepSliding\\ \cite{DBLP:journals/corr/SongX15}} & 44.2& 78.8 &11.9 &61.2 &20.5 &6.4 &15.4 &53.5 &50.3& 78.9 &N/A&N/A&19.55s& 42.1\\
\hline
\makecell{Frustum VoxNet v1\\ \cite{Shen_2020_WACV}} &42.4 &78.5 &18.0 &47.2 &12.4 &18.2 &34.5& 40.3& 30.4 &84.5 &47.1&47.6&0.16s&41.8 \\
\hline
\makecell{Frustum VoxNet v2\\ \cite{Shen_2020_v2}} &56.7 &79.9 &19.8 &48.1 &15.1 &22.3 &38.8& 43.2& 34.1 &91.6 &N/A&N/A&0.21s& 45.0\\
\hline
\makecell{Frustum PointNet\\ \cite{DBLP:journals/corr/abs-1711-08488}} &43.3 &81.1 &33.3 &64.2 &24.7 &32.0 &58.1& 61.1& 51.1 &90.9 &N/A&N/A&0.12s& 54.0\\
\hline
\end{tabular}
}
\end{center}
\caption{3D object detection AP on SUN-RGBD val set. Evaluation metric is average precision with 3D IoU threshold 0.25 as proposed
by \cite{Song_2015_CVPR}. }
\label{performance_comp_sun_rgbd}
\end{table}

\subsubsection{Comparison by application scenario}

\begin{table}[H]
\scriptsize
\begin{center}
\begin{tabular}{|c|c|c|c|c|}
\hline
Scenario &dataset&X(meter)&Y(meter)&Z(meter)\\
\hline
indoor&\makecell{SUN RGB-D}&5.2&6&2.5\\
\hline
outdoor&KITTI&70.4&80&4\\
\hline
\end{tabular}
\end{center}
\caption{Indoor(data is collected from Deep Sliding Shape \cite{DBLP:journals/corr/SongX15}) vs. outdoor(data is collected from VoxelNet\cite{DBLP:journals/corr/abs-1711-06396}) }
\label{different_range_of_indoor_outdoor}
\end{table}
3D object detection systems can be categorized by the supported application scenarios: indoor only, outdoor only or both. There are two main differences between indoor and outdoor scenarios: first, the range of indoor is small and of outdoor is large. A comparison of the indoor range and outdoor range based on two typical datasets is shown in Table \ref{different_range_of_indoor_outdoor}. Second, as the distribution of the outdoor objects is more sparse and the categories of interesting objects of the outdoor scenarios is less compared with indoor scenarios, the outdoor scenarios can use BEV to generate the proposals and then do the detection. However, the indoor scenarios, generation only based on BEV will get a bad performance since there might be multiple objects in the vertical direction.\\

BEV only  proposal generation algorithms such as MV3D\cite{DBLP:journals/corr/ChenMWLX16} do not performance well indoors.  At the same time, as the space is too large for the outdoor scenario, some 3D CNN based algorithms, such as Deep Sliding Shape \cite{DBLP:journals/corr/SongX15}, can work well for indoors but may have a high possibility to fail for the outdoor scenario without adjustment.\\

\begin{table}[H]
\scriptsize
\begin{center}
\begin{tabular}{|c|c|}
\hline
Scenario & System\\
\hline
\multirow{2}{*}{Indoor}&Deep Sliding Shape\cite{DBLP:journals/corr/SongX15}\\
\hhline{~-}
& Frustum Voxnet v1 and v2\cite{Shen_2020_WACV}\cite{Shen_2020_v2}	\\
\hline
\multirow{3}{*}{Outdoor} & MV3D\cite{DBLP:journals/corr/ChenMWLX16}	\\
\hhline{~-}
 & AVOD\cite{2017arXiv171202294K}	\\
\hhline{~-}
 &VoxelNet\cite{DBLP:journals/corr/abs-1711-06396}\\
\hline
Both & F-PointNet\cite{DBLP:journals/corr/abs-1711-08488}\\
\hline
\end{tabular}
\end{center}
\caption{Object detection system based on 3D image data for different application scenarios. }
\label{by_scenario}
\end{table}

In the rest of this section we will introduce several of the latest papers organized by different application scenarios as shown in Table \ref{by_scenario}. \\

\subsubsection{Indoor}
Some 3D object detection systems can work well for the indoor scenarios are introduced in this part.\\

%\textit{SUN-RGBD dataset\cite{Song_2015_CVPR}}\\

%The SUN-RGBD dataset focus on indoor environments. For this dataset as many as 700 object categories are labeled. The dataset is collected via different types of RGB-D cameras with varying resolutions. It has 5285 images for training and 5050 images for testing.\\
%With the arrival of reliable and affordable RGB-D sensor such as Kinect, lots of the RGB-D data are available. Meanwhile, we also have some label RGB-D data set such as NYUv2 and SUN RGB-D.\\

 %\subsubsection{classification}

%In the detection task based on the indoor RGB-D data, Sliding Shapes was proposed to slide a 3D detection window in 3D space and detect the 3D object based on each 3D sliding window. The Depth RCNN takes a 2D approach by encoding the depth information as extra channels and padding them to the original RGB channels.\\

\textbf{Deep Sliding Shapes \cite{DBLP:journals/corr/SongX15}}\\
%The Deep Sliding Shapes  is using the 3D CNN directly on the 3D voxel space by using the  Truncated Signed Distance Function(TSDF) coding. The deep sliding shapes framework will be detailed introduced in this survey.\\

\textit{The TSDF 3D Representation}\\

\begin{figure}[H]
\begin{center}
%\fbox{\rule{0pt}{2in} \rule{.9\linewidth}{0pt}}
\includegraphics[width=1.0\linewidth]{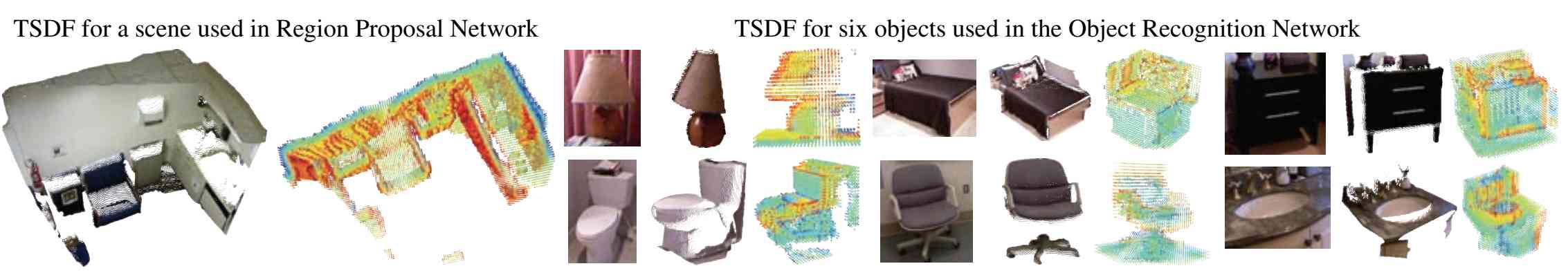}

\end{center}
   \caption{Visualization of TSDF Encoding. Only the TSDF values when close to the surface visualized . Red indicates the voxel is in
front of surfaces and blue indicates the voxel is behind the surface. The resolution is $208\times208\times100$ for the Region Proposal Network,
and $30\times30\times30$ for the Object Classification Network. Figure and Caption are from \cite{DBLP:journals/corr/SongX15}.}
\label{fig:tsdf}
\end{figure}

A directional Truncated Signed Distance Function(TSDF) is used to encode 3D shapes.  The 3D space is divided into 3D voxel grid and the value in each voxel is defined to be the shortest distance between the voxel center and the surface from the input depth map. To encode the direction of the surface point, a directional TSDF stores a three-dimensional vector $[dx,dy,dz]$ in each voxel
in order to record the distance in three directions to the closest surface point instead of a single distance value\cite{DBLP:journals/corr/SongX15}. The example of the directional TSDF for the RPN and detection network is shown in Figure \ref{fig:tsdf}.\\

\textit{The 3D RPN}\\

\begin{figure}[H]
\begin{center}
%\fbox{\rule{0pt}{2in} \rule{.9\linewidth}{0pt}}
\includegraphics[width=1.0\linewidth]{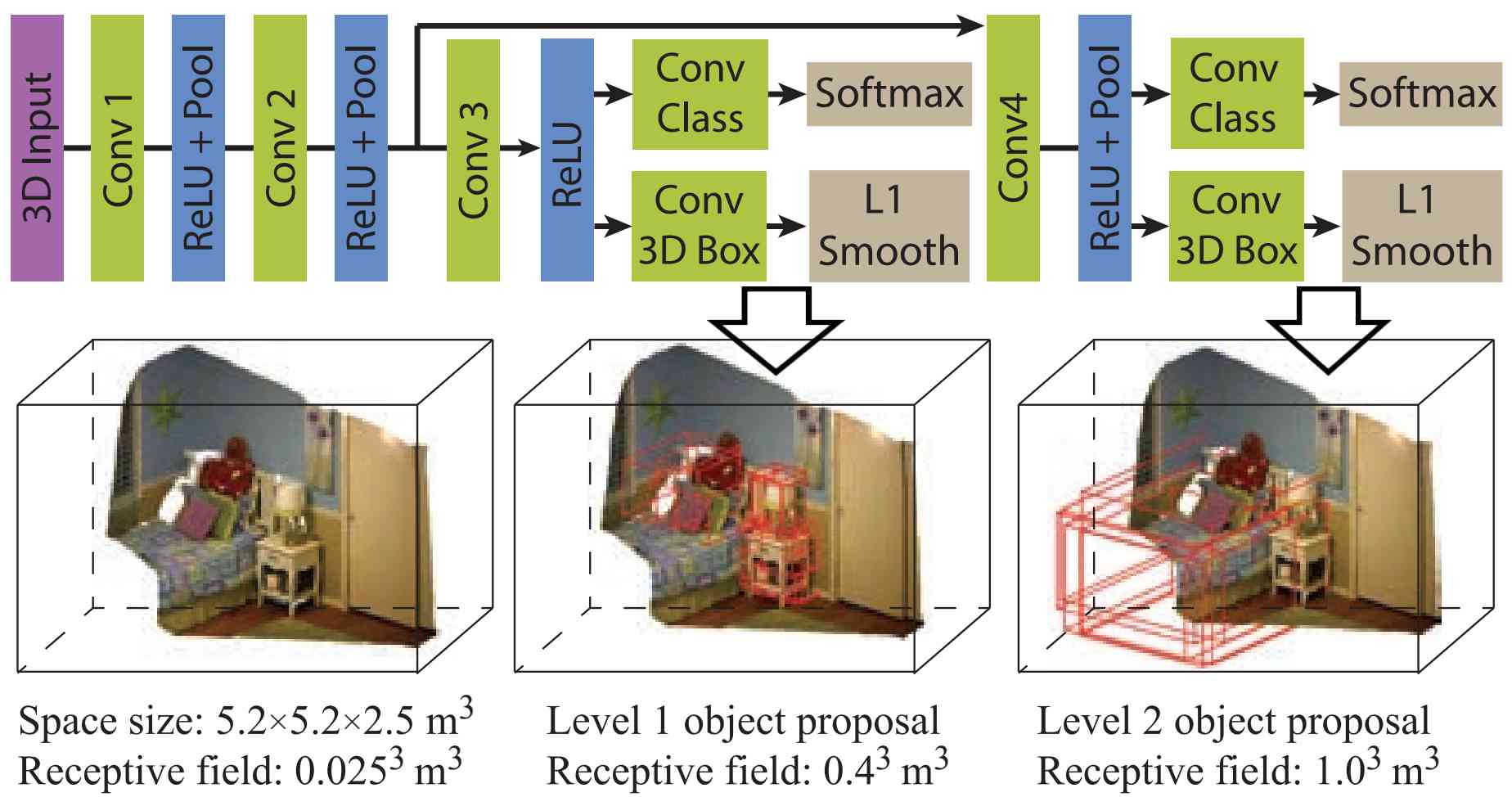}

\end{center}
   \caption{3D Region Proposal Network: Taking a 3D
volume from depth as input,  a fully convolutional 3D network extracts 3D proposals at two scales with different receptive fields. Figure and Caption are from \cite{DBLP:journals/corr/SongX15}.}
\label{fig:11}
\end{figure}
Deep Sliding Shapes \cite{DBLP:journals/corr/SongX15} is inspired by the Faster RCNN\cite{DBLP:conf/nips/RenHGS15} framework. The proposals are generated by a CNN based RPN. The RPN is shown in Figure \ref{fig:11}. The network structure shown in Figure  \ref{fig:11}:\\
\begin{itemize}
\item is using 3D CNN to do the feature extraction.\\
\item has two level region proposals since the variation of the physical size of the 3D object is large.\\
\end{itemize}

\begin{figure}[H]
\begin{center}
%\fbox{\rule{0pt}{2in} \rule{.9\linewidth}{0pt}}
\includegraphics[width=1.0\linewidth]{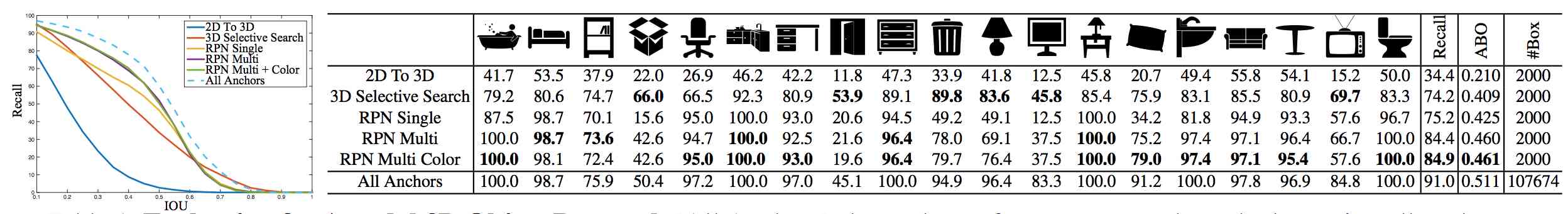}

\end{center}
   \caption{Evaluation for 3D Object Proposal. ``All Anchors" shows the performance upper bound when using all anchors. ``RPN Multi" means two-level proposal RNP network. ``RPN Multi Color" means the RGB color information is projected to each voxel. Figure and Caption are from \cite{DBLP:journals/corr/SongX15}.}
\label{fig:55}
\end{figure}

\begin{figure}[H]
\begin{center}
%\fbox{\rule{0pt}{2in} \rule{.9\linewidth}{0pt}}
\includegraphics[width=1.0\linewidth]{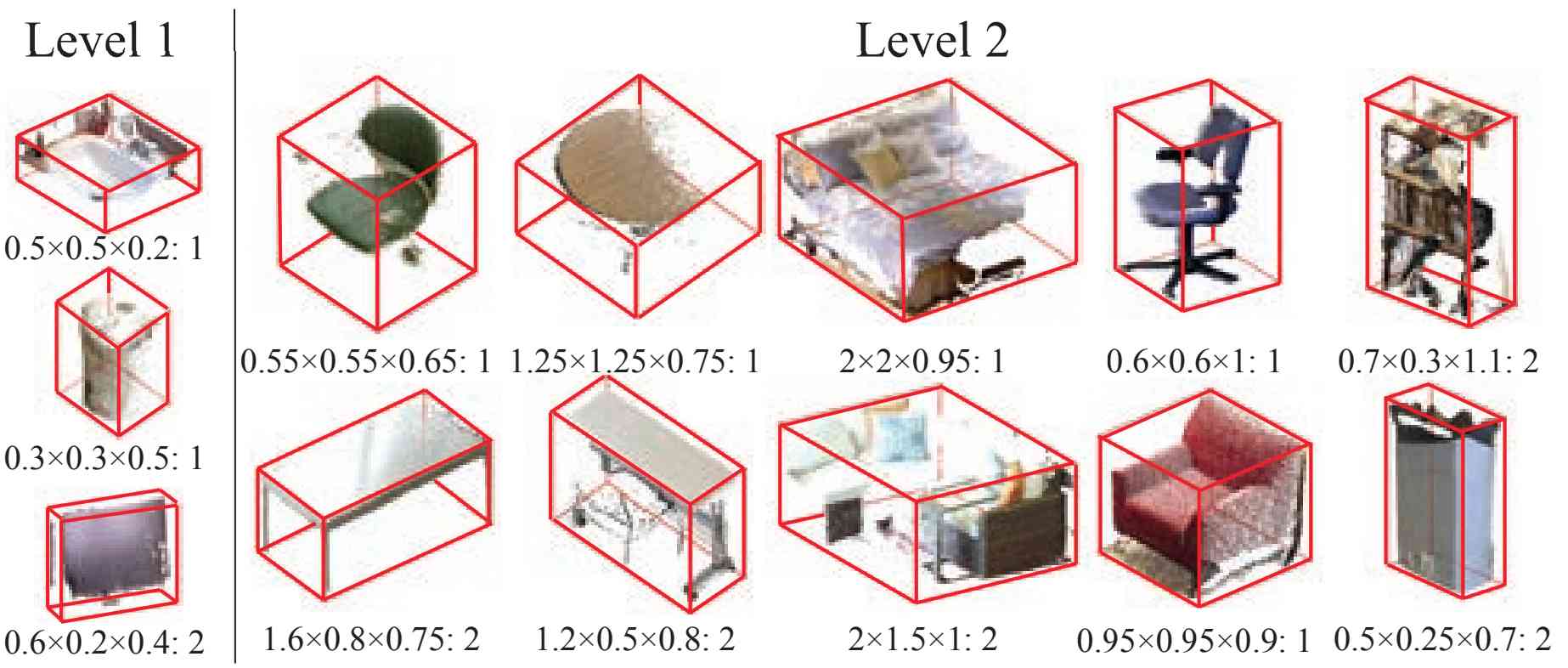}

\end{center}
   \caption{List of All Anchors Types. The subscripts show the
$width *depth * height$ in meters, followed by the number of
orientations for this anchor after the colon. Figure and Caption are from \cite{DBLP:journals/corr/SongX15}.}
\label{fig:33}
\end{figure}

3D Selective Search(SS) is also proposed to compare the performance with the CNN based on RPN. The result is shown in Figure \ref{fig:55}. From the result we can see similarities to the conclusion from Faster RCNN\cite{DBLP:conf/nips/RenHGS15}: the CNN based RPN has a better performance compared with the traditional SS method for the 3D proposals generation. Anchors used for the RPN are shown in Figure \ref{fig:33}.\\ 

Features fed to the RPN is from both the depth channel and the RGB image. A directional Truncated Signed Distance Function(TSDF) encoding is used to change the depth channel information to a 3D voxel grid data representation. Also the RGB color is projected to each voxel to improve the proposal results as shown in Figure \ref{fig:55}.\\

\textit{The detection network}\\

\begin{figure}[H]
\begin{center}
%\fbox{\rule{0pt}{2in} \rule{.9\linewidth}{0pt}}
\includegraphics[width=1.0\linewidth]{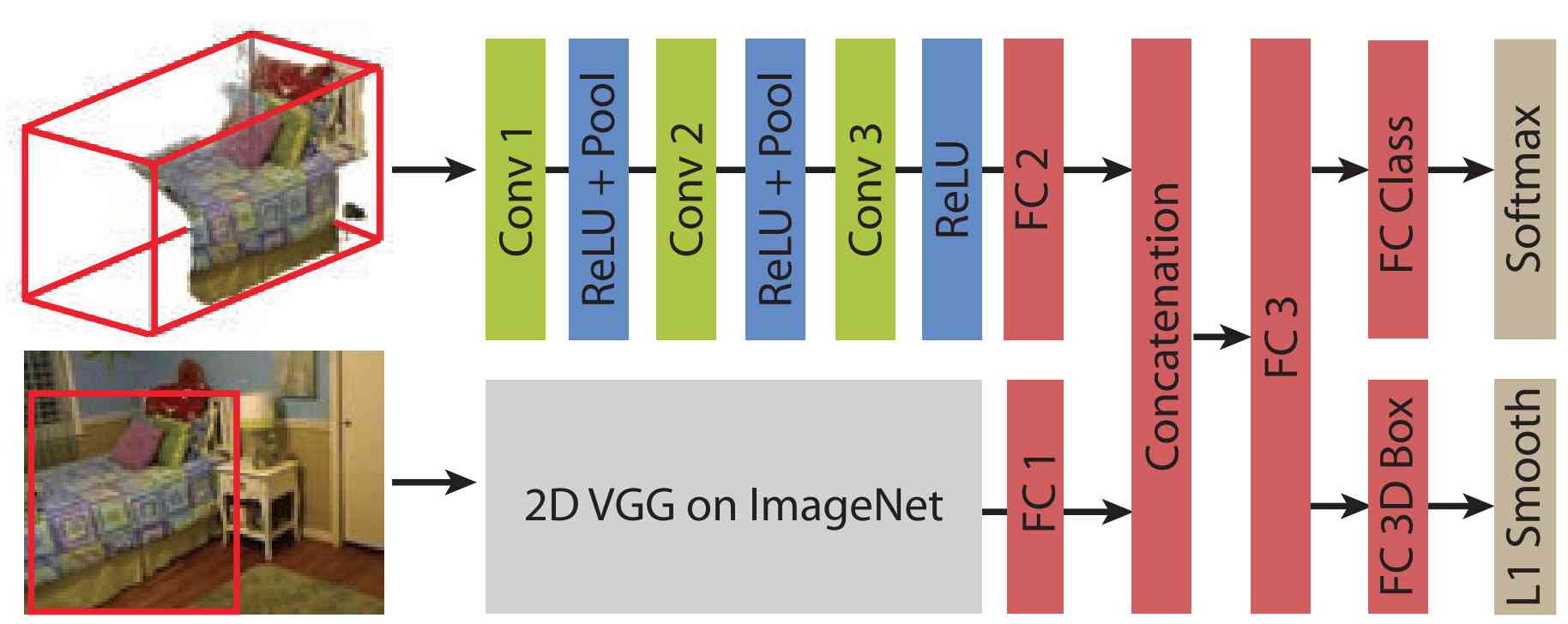}

\end{center}
   \caption{Joint Object Classification Network: For each 3D proposal, 3D volume from depth is fed to a 3D ConvNet, and feed the 2D color patch (2D projection of the 3D proposal) to a 2D ConvNet, to jointly learn object category and 3D box regression. Figure and Caption are from \cite{DBLP:journals/corr/SongX15}.}
\label{fig:22}
\end{figure}

The detection network is shown in Figure \ref{fig:22}. The detection network is using early fusing model to fuse the depth and RGB images together. The depth image is using the TSDF which is the same as RPN. The feature of 2D image is extracted based on the ImageNet pre-trained VGG network. The difference between the Faster RCNN\cite{DBLP:conf/nips/RenHGS15} and Deep Sliding Shape \cite{DBLP:journals/corr/SongX15} is: the RPN and detection network are trained separately in Deep Sliding Shape while in Faster RCNN the two networks share the convolutional layers. \\

Another important part for the Deep Sliding Shape is that it is using the different resolutions for the RPN and detection networks. The comparison of this difference is shown in Table \ref{rpn_detection}.\\

\begin{table}[H]
\scriptsize
\begin{center}
\begin{tabular}{|c|c|c|c|}
\hline
network & resolution& 3D CNN input data shape& physical size\\
\hline
RPN&$0.025\times0.025\times0.025$&$208\times208\times 100$&$5.2\times6.0\times 2.5$\\
\hline
Detection for bed&$0.067\times0.067\times0.032$&$30\times30\times 30$&$2.0\times2.0\times 0.95$\\
\hline
Detection for trash can&$0.010\times0.010\times0.012$&$30\times30\times 30$&$0.3\times0.3\times 0.5$\\
\hline
\end{tabular}
\end{center}
\caption {Resolution and shape comparison between the RPN and detection network. The detection network has a fixed input shape which is $30\times30\times 30$ and the resolution is decided by the proposed region size and the input shape. In this table, an anchor of the bed and an anchor of the trash can are used as examples of proposal's physical size to make the comparison. }
\label{rpn_detection}
\end{table}

\textit{The results}\\

\begin{figure}[H]
\begin{center}
%\fbox{\rule{0pt}{2in} \rule{.9\linewidth}{0pt}}
\includegraphics[width=1.0\linewidth]{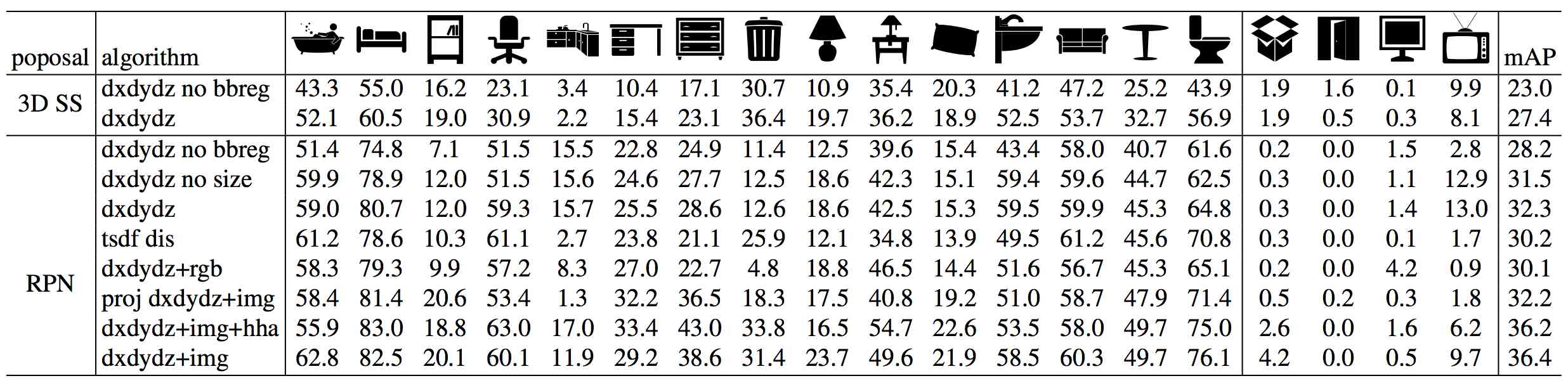}

\end{center}
   \caption{Control Experiments on NYUv2 Test Set. Not working: box (too much variance), door (planar), monitor and tv (no depth). Figure and Caption are from \cite{DBLP:journals/corr/SongX15}.}
\label{fig:66}
\end{figure}
 The performance of 3D region proposal and object detection
algorithm of Deep Sliding Shapes is evaluated on the standard NYUv2 dataset \cite{Silberman:2012:ISS:2403138.2403195} and SUN
RGB-D dataset\cite{Song_2015_CVPR}. The result of the  NYUv2 Test Set is shown in Figure \ref{fig:66}. From Row [RPN: dxdydz] to Row
[RPN: dxdydz+img] in Figure \ref{fig:66}, the performance of different feature encodings is compared and the main conclusions are given below: (1)
TSDF with directions encoded is better than single TSDF distance ([dxdydz] vs. [tsdf dis]). (2)Directly encoding color on 3D voxels
is not as good as using 2D image VGGnet ([dxdydz+rgb] vs. [dxdydz+img]), probably because the latter one can
preserve high frequency signal from images. (3) Adding HHA\footnote{HHA is proposed in \cite{DBLP:journals/corr/GuptaGAM14}. In \cite{DBLP:journals/corr/GuptaGAM14} the depth image is encoded with three channels at each pixel:  \textbf{H}orizontal disparity, \textbf{H}eight above ground, and the  \textbf{A}ngle the pixel's local surface normal makes with the inferred gravity direction.} does not help, which indicates the depth information from HHA is already exploited by the 3D representation([dxdydz+img+hha] vs. [dxdydz+img]).\\

%Depth RCNN detects objects in the 2D image plane by treating depth as extra channels of a color image, then fit a 3D model to the points inside the 2D detected window by using ICP alignment.

%\begin{figure}[H]
%\begin{center}
%\fbox{\rule{0pt}{2in} \rule{.9\linewidth}{0pt}}
%\includegraphics[width=1.0\linewidth]{77}

%\end{center}
  % \caption{Comparison on 3D Object Detection}
%\label{fig:77}
%\end{figure}

\begin{figure}[H]
\begin{center}
%\fbox{\rule{0pt}{2in} \rule{.9\linewidth}{0pt}}
\includegraphics[width=1.0\linewidth]{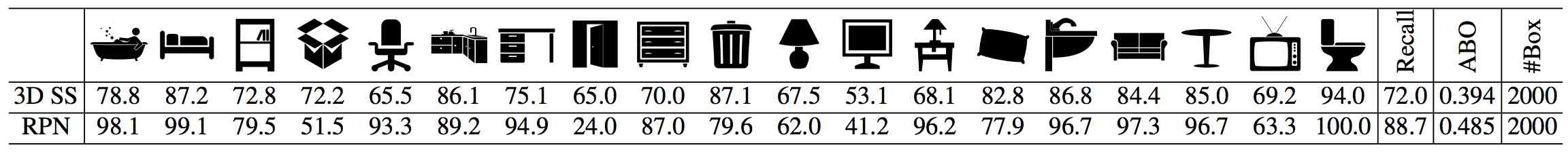}

\end{center}
   \caption{Evaluation for region proposal generation on SUN RGB-D test set\cite{DBLP:journals/corr/SongX15}.}
\label{fig:88}
\end{figure}

\begin{figure}[H]
\begin{center}
%\fbox{\rule{0pt}{2in} \rule{.9\linewidth}{0pt}}
\includegraphics[width=0.80\linewidth]{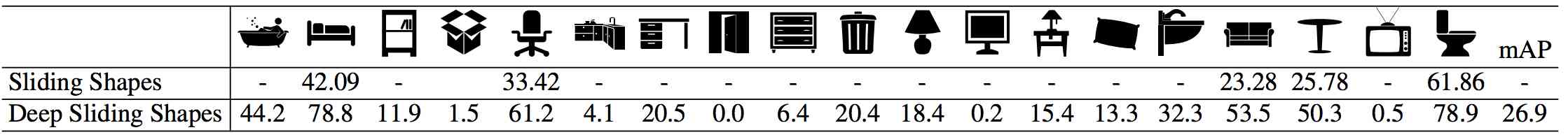}

\end{center}
   \caption{Evaluation for 3D object detection on SUN RGB-D test set\cite{DBLP:journals/corr/SongX15}.}
\label{fig:99}
\end{figure}

\begin{figure}[H]
\begin{center}
%\fbox{\rule{0pt}{2in} \rule{.9\linewidth}{0pt}}
\includegraphics[width=1.0\linewidth]{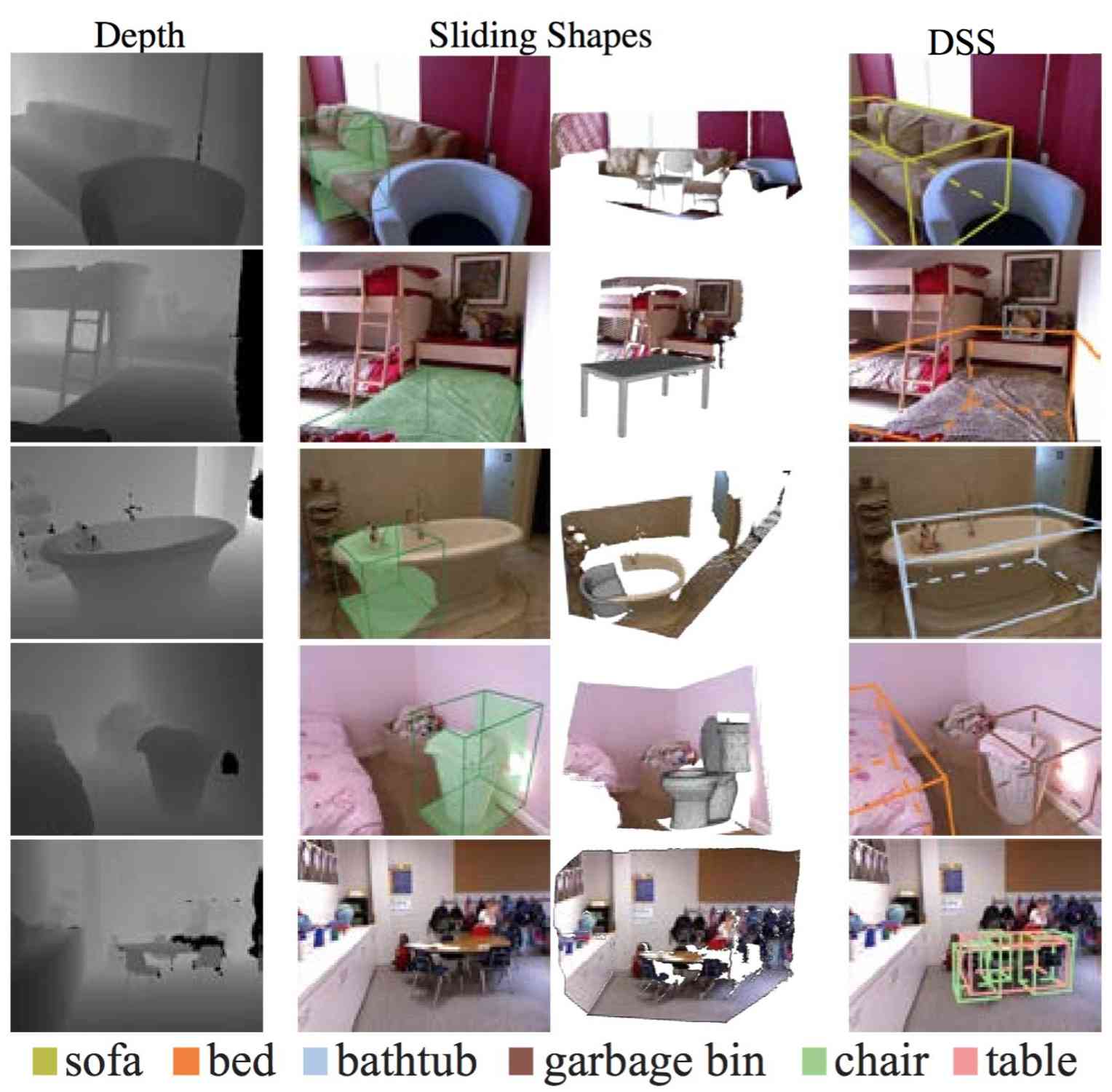}

\end{center}
   \caption{Comparison with Sliding Shapes. DSS algorithm is able to better use shape, color and contextual information to handle more object categories, solve the ambiguous cases, and detect objects with a typical size. Figure and Caption are adjusted from \cite{DBLP:journals/corr/SongX15}.}
\label{fig:1010}
\end{figure}

\begin{figure}[H]
\begin{center}
%\fbox{\rule{0pt}{2in} \rule{.9\linewidth}{0pt}}
%\includegraphics[width=1.0\linewidth]{whole_system_v3.eps}
\includegraphics[width=1.0\linewidth]{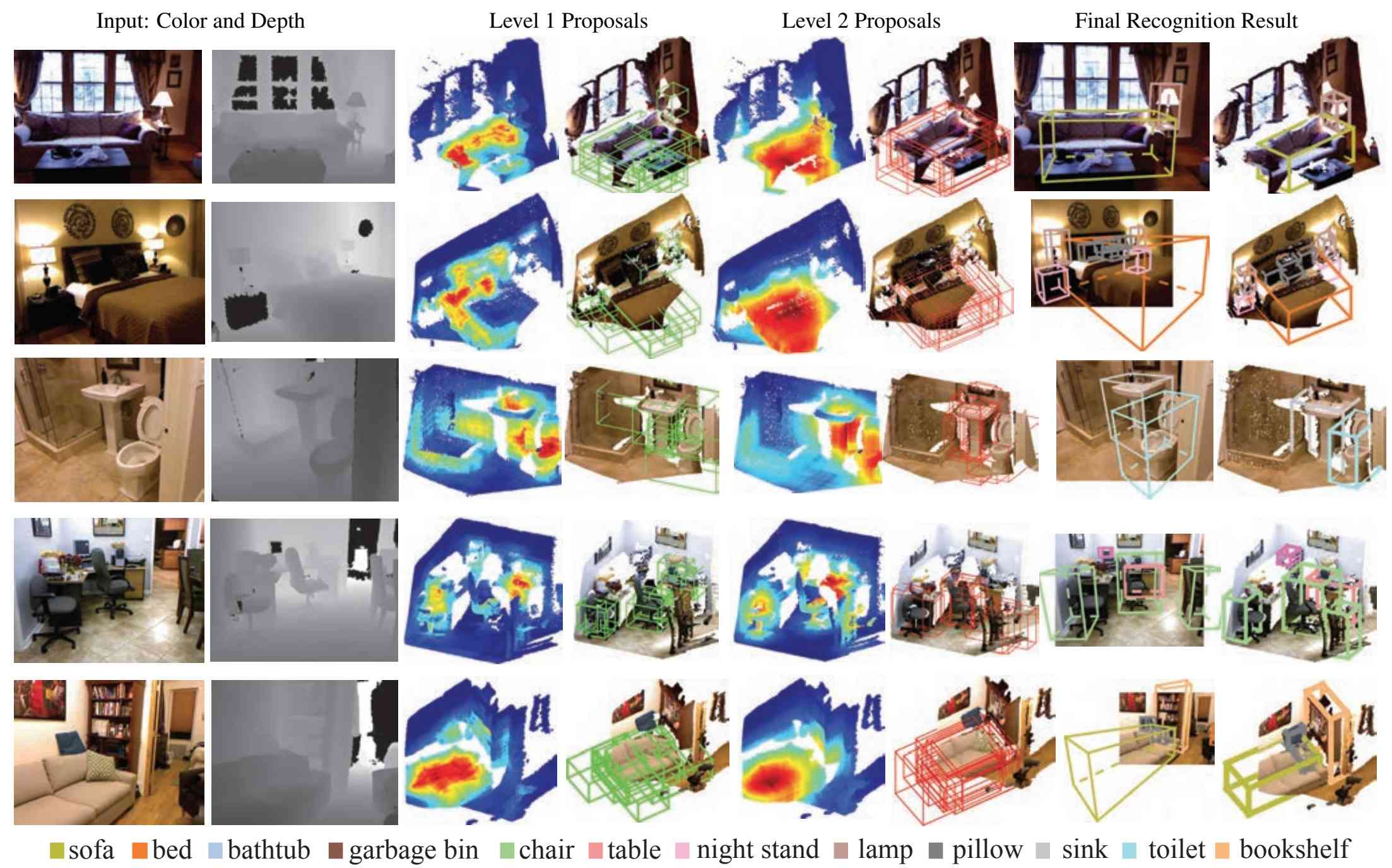}
\end{center}
   \caption{Examples for Detection Results. For the proposal results, The heat map for the distribution of the top proposals (red is
the area with more concentration) and a few top boxes after NMS are shown. For the classification results, the 3D detection can estimate the
full extent of 3D both vertically (e.g. bottom of a bed) and horizontally (e.g. full size sofa in the last row). Figure and Caption are from \cite{DBLP:journals/corr/SongX15}.}
\label{fig:44}
\end{figure}

Results on the SUN RGB-D dataset are shown in Figure \ref{fig:99} and \ref{fig:1010}. The comparison of the detection results of DSS with the F-PointNet\cite{DBLP:journals/corr/abs-1711-08488} is shown in Table\ref{performance_comp_sun_rgbd}. From the results, we can see that RPN has a better performance than Selective Search. Deep sliding shape has a better performance than their previous work Sliding Shape\cite{DBLP:journals/corr/SongX14}.  Sliding Shapes\cite{DBLP:journals/corr/SongX14} is a 3D object
detector that runs sliding windows in 3D to directly classify each 3D window by using SVM. Figure \ref{fig:1010} shows side-by-side comparisons to Sliding
Shapes. Figure \ref{fig:44} shows some general example for the Deep Sliding Shape system.\\

\textbf{Frustum VoxNet v1/v2\cite{Shen_2020_WACV}\cite{Shen_2020_v2}}\\

\begin{figure}[H]
\begin{center}
%\fbox{\rule{0pt}{2in} \rule{.9\linewidth}{0pt}}
%\includegraphics[width=1.0\linewidth]{whole_system_v3.eps}
\includegraphics[width=1.0\linewidth]{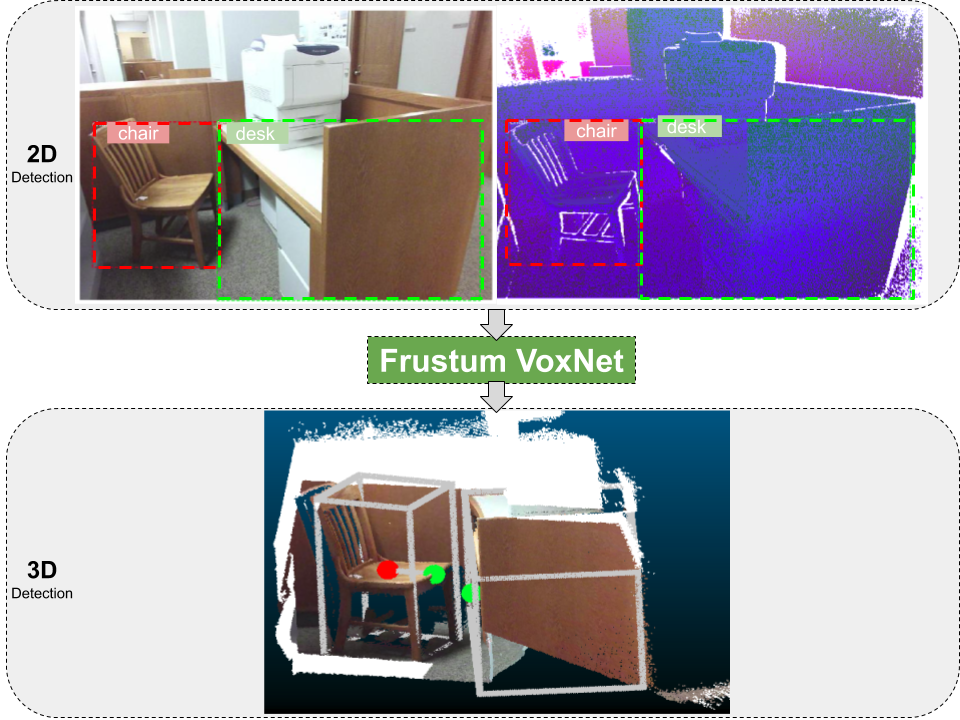}
\end{center}
   \caption{Overview of the whole system. Upper left: RGB image and detected 2D bounding boxes. Upper right: DHS (\textbf{D}epth \textbf{H}eight and \textbf{S}igned angle) image, and detected 2D bounding boxes. A DHS image is a pseudo-RGB image generated by a depth image (see text). 
	Bottom: The final 3D detected objects from the associated 3D range image. The 3D detection not only provides an amodal bounding box but also an orientation. The red point is the center of the bounding box and the green one is the front center. 
	%These two points indicate the orientation.
	The detected 2D bounding boxes from either and RGB or DHS image, generate 3D frustums (which are prisms having as apex the sensor location and extend through the 2D bounding boxes to the 3D space). They are then fed to our Frustum VoxNet network, which produces the 3D detections. Figure and Caption are from \cite{Shen_2020_WACV}.}
\label{fig:9999}
\end{figure}

Frustum VoxNet v1/v2 are 2D driven 3D object detection system. Interesting 2D objects are detected from the RGB or depth only images. For the depth only image, the depth is further converted to pseudo-RGB image with DHS (\textbf{D}epth \textbf{H}eight and \textbf{S}igned angle) 3 channels. Frustums are generated through the bounding box of the 2D detection. After that, parts of the frustums (since frustums can be re- ally large), is voxelized instead of using the whole frustums. These 3D proposals are generated and voxelized and then fed to an efficient ResNet-based 3D Fully Convolutional Network (FCN).\\

\begin{figure}[H]
\begin{center}
   \includegraphics[width=1.0\linewidth]{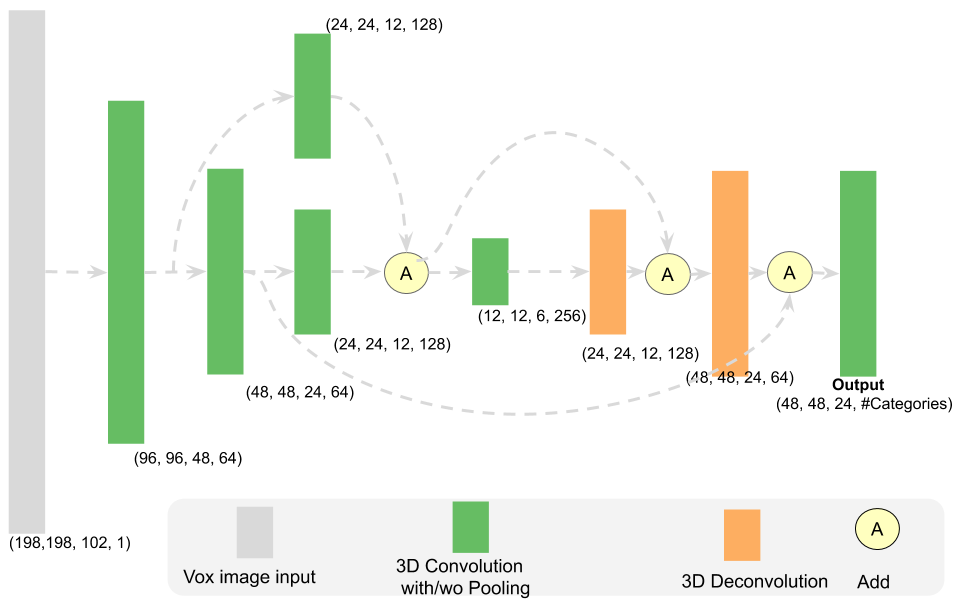}
\end{center}
    \caption{Segmentation architecture (used for large short scale). Every 3D CNN layer will be followed by a dropout layer. The tensor shape shown here is the output shape of each block. It provides the (width, depth, height, channel) information of the network. Figure and Caption are from \cite{Shen_2020_v2}.}
        \label{3DCNNnetwork_seg}
\end{figure}

\begin{figure}[H]
\begin{center}
   \includegraphics[width=1.0\linewidth]{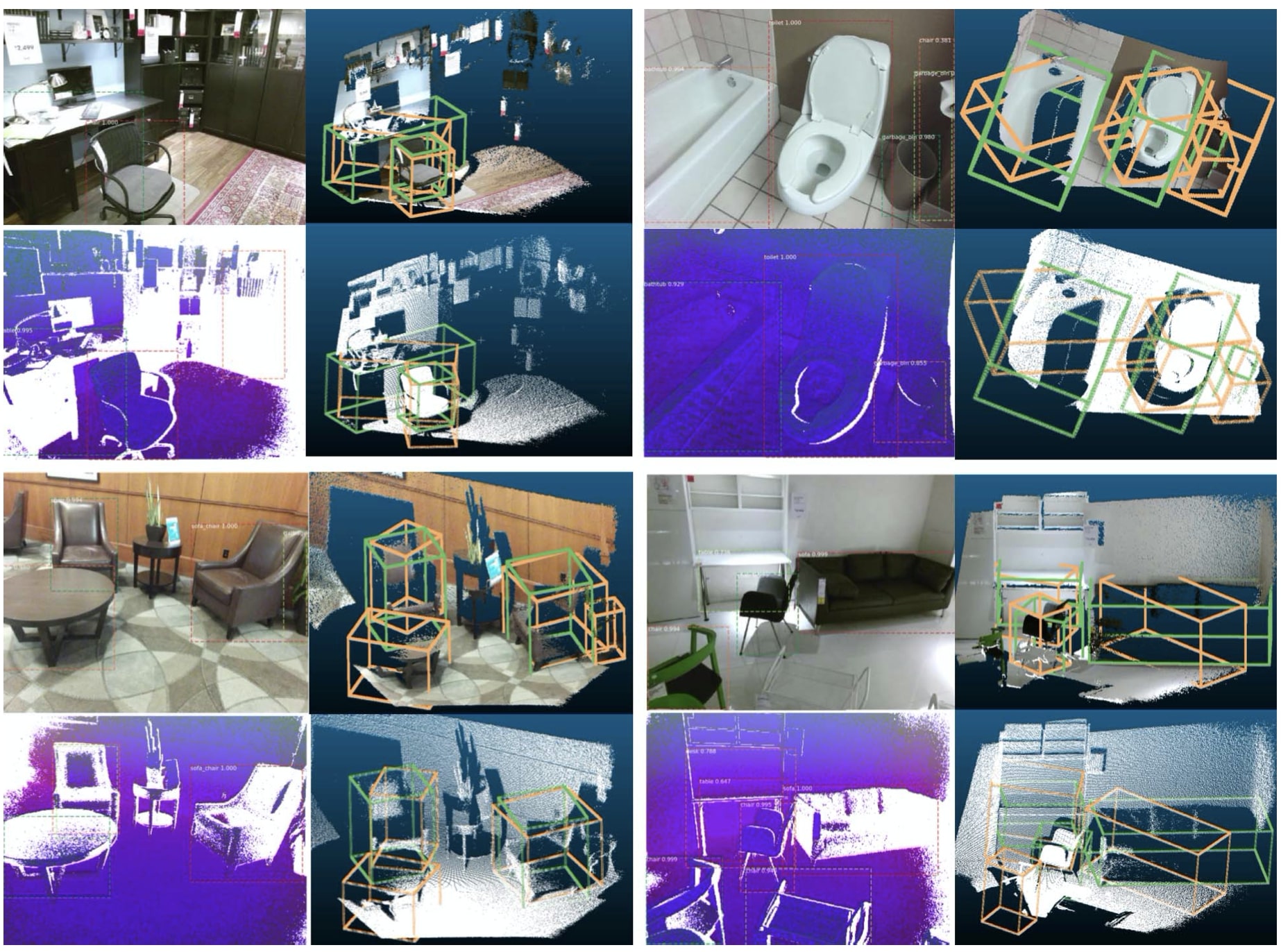}
\end{center}
    \caption{Visualizations of 2D and 3D detection results. This visualization contains four images. For each image, upper left shows 2D detection based on RGB image. Upper right shows the corresponding 3D detection results (light green ones are the 3D ground truth boxes and orange-colored boxes are predictions) based on frustums generated from RGB image 2D detections (to have a better visualization, RGB colors are projected back to the cloud points). Figure and Caption are from \cite{Shen_2020_WACV}.}
        \label{lucky}
\end{figure}

The v1 system overview can be found in Figure \ref{fig:9999}. For the Frustum v2 system, besides the 3D object detection, 3D instance segmentation are also achieved. As this is a 2D driven 3D object detection system, and for the 3D part, only 6 layers of Fully Convolutional Networks are used, the whole system has a fast inference speed. The 3D instance segmentation network can be found in Figure \ref{3DCNNnetwork_seg}. Visualizations of the 2D and 3D detection results based on Frustum VoxNet v1 can be found in Figure \ref{lucky}.\\

\subsubsection{Outdoor}
KITTI\cite{Geiger2012CVPR} is an important outdoor dataset. For the outdoor scenario, the algorithms used to do the outdoor object detection will be studied by the leaderboard of the 3D object detection for the KITTI data set by April, 2018.\\

\textbf{The leaderboard of 3D object detection for the KITTI dataset}

\begin{table}[H]
\scriptsize
\begin{center}
\begin{tabular}{|c|c|c|c|c|c|}
\hline
 \textbf{Method}	   	 	 &	\textbf{Moderate}	 	       &   \textbf{Easy}	 	         &     \textbf{Hard}	 	         &          \textbf{Runtime}	 &	\textbf{Environment}	 	\\
\hline
AVOD-FPN\cite{2017arXiv171202294K}	 	       &	71.88 $\%$	 &81.94 $\%$	 &66.38 $\%$	 &0.1 s	      &\makecell{Titan X \\(Pascal)}	\\
\hline
F-PointNet\cite{DBLP:journals/corr/abs-1711-08488}	        &	70.39 $\%$	 &81.20 $\%$	 &62.19 $\%$	 &0.17 s        &	\makecell{GPU@3.0 Ghz\\ (Python)}	\\
\hline
DF-PC$\_$CNN\cite{2017arXiv171202294K}  	 &	66.22 $\%$	 &80.28 $\%$	 &58.94 $\%$	 &0.5 s	     &  \makecell{GPU@3.0 Ghz\\(Matlab + C/C++)}	\\
\hline
AVOD\cite{2017arXiv171202294K}	        	       &       65.78 $\%$	 &73.59 $\%$	 &58.38 $\%$	 &0.08 s	     &\makecell{Titan X\\(pascal)}	\\
\hline
VoxelNet\cite{DBLP:journals/corr/abs-1711-06396}  	 &	65.11 $\%$	 &77.47 $\%$	 &57.73 $\%$	 &0.03 s	     &\makecell{GPU@2.5 Ghz\\(Python + C/C++)}	\\
\hline
MV3D\cite{DBLP:journals/corr/ChenMWLX16}	              &	62.35 $\%$	 &71.09 $\%$	 &55.12 $\%$	 &0.36 s	     &\makecell{GPU@2.5 Ghz\\(Python + C/C++)}	\\
\hline
\makecell{MV3D\\(LiDAR)}\cite{DBLP:journals/corr/ChenMWLX16}	       &52.73 $\%$        &66.77 $\%$	 &51.31 $\%$	 &0.24 s	     &\makecell{GPU@2.5 Ghz\\(Python + C/C++)}	\\
\hline
F-PC$\_$CNN\cite{2017arXiv171202294K}	   	       &	48.07 $\%$	 &60.06 $\%$    &45.22 $\%$	 &0.5 s	     &    \makecell{  GPU@3.0 Ghz\\(Matlab + C/C++)}\\
\hline
\end{tabular}
\end{center}
\caption{3D object detection evaluation result for \textbf{Car} based on the KITTI evaluation server.  All methods are ranked based on the moderately difficult results. The table is adjusted from \cite{kittileaderboard}. The top 8 are listed here. }
\label{car}
\end{table}

\begin{table}[H]
\scriptsize% or footnotesize, scriptsize, tiny, etc.
\begin{center}
\begin{tabular}{|c|c|c|c|c|c|c|}
\hline
	\textbf{Method}	   	 	 &	\textbf{Moderate}	 	       &   \textbf{Easy}	 	         &     \textbf{Hard}	 	         &          \textbf{Runtime}	 &	\textbf{Environment}	 	\\
\hline
F-PointNet\cite{DBLP:journals/corr/abs-1711-08488}	 	       &	44.89 $\%$	 &51.21 $\%$	 &40.23 $\%$	 &0.17 s        &	\makecell{GPU@3.0 Ghz\\ (Python)}	\\
\hline
AVOD-FPN\cite{2017arXiv171202294K}		       &	39.00 $\%$	 &46.35 $\%$	 &36.58 $\%$	 &0.1 s	      &\makecell{Titan X \\(Pascal)}	\\
\hline
 VoxelNet\cite{DBLP:journals/corr/abs-1711-06396}  	 &	33.69 $\%$	 &39.48 $\%$	 &31.51 $\%$	 &0.03 s	     &  \makecell{GPU@2.5 Ghz\\(Python + C/C++)}	\\
\hline
AVOD\cite{2017arXiv171202294K}	              &       31.51 $\%$	 &38.28 $\%$	 &26.98 $\%$	 &0.08 s	     &\makecell{Titan X\\(pascal)}	\\
\hline
\end{tabular}
\end{center}
\caption{3D object detection evaluation result for \textbf{Pedestrian} based on the KITTI evaluation server.  All methods are ranked based on the moderately difficult results. The table is adjusted from \cite{kittileaderboard}. The top 4 are listed here.}
\label{pedestrian}
\end{table}

\begin{table}[H]
\scriptsize
\begin{center}
\begin{tabular}{|c|c|c|c|c|c|c|}
\hline
\textbf{Method}	   	 	 &	\textbf{Moderate}	 	       &   \textbf{Easy}	 	         &     \textbf{Hard}	 	         &          \textbf{Runtime}	 &	\textbf{Environment}	 	\\
\hline
F-PointNet\cite{DBLP:journals/corr/abs-1711-08488}		      &	56.77 $\%$	 &71.96 $\%$	 &50.39 $\%$	 &0.17 s        &	\makecell{GPU@3.0 Ghz\\ (Python)}	\\
\hline
VoxelNet(LiDAR)\cite{DBLP:journals/corr/abs-1711-06396}  	 &	48.36 $\%$	 &61.22 $\%$	 &44.37 $\%$	 &0.03 s	     &  \makecell{GPU@2.5 Ghz\\(Python + C/C++)}	\\
\hline
AVOD-FPN\cite{2017arXiv171202294K}		       &	46.12 $\%$	 &59.97 $\%$	 &42.36 $\%$	 &0.1 s	      &\makecell{Titan X \\(Pascal)}	\\
\hline
AVOD\cite{2017arXiv171202294K}	            &       44.90 $\%$	 &60.11 $\%$	 &38.80 $\%$	 &0.08 s	     &\makecell{Titan X\\(pascal)}	\\
\hline
\end{tabular}
\end{center}
\caption{3D object detection evaluation result for \textbf{Cyclist} based on the KITTI evaluation server.  All methods are ranked based on the moderately difficult results. The table is adjusted from \cite{kittileaderboard}. The top 4 are listed here.}
\label{cyclist}
\end{table}

%According to the types of input features used in the algorithm, the current algorithms for 3D object detection can be categorized into four major groups: (1) mono image based, (2) stereo image, (3) LiDAR (Light Detection and Ranging) only, and (4) fusion between mono image and LiDAR.

\textbf{MV3D\cite{DBLP:journals/corr/ChenMWLX16} and AVOD\cite{2017arXiv171202294K}}\\

The comparison of the input data and feature encoding for the RPN and detection part of MV3D\cite{DBLP:journals/corr/ChenMWLX16} and AVOD \cite{2017arXiv171202294K} can be found in Table \ref{by_input_data}. Since MV3D is using the BEV only to propose the region candidates, it performances good on big objects such as cars. However, it will not have a good performance for small object detection such as pedestrians and also for the indoor scenario. AVOD\cite{2017arXiv171202294K} further improves the MV3D\cite{DBLP:journals/corr/ChenMWLX16} by using BEV and RGB image to propose the region candidates.\\

\textit{Framework of MV3D}\\

\begin{figure}[H]
\begin{center}
%\fbox{\rule{0pt}{2in} \rule{.9\linewidth}{0pt}}
\includegraphics[width=1.0\linewidth]{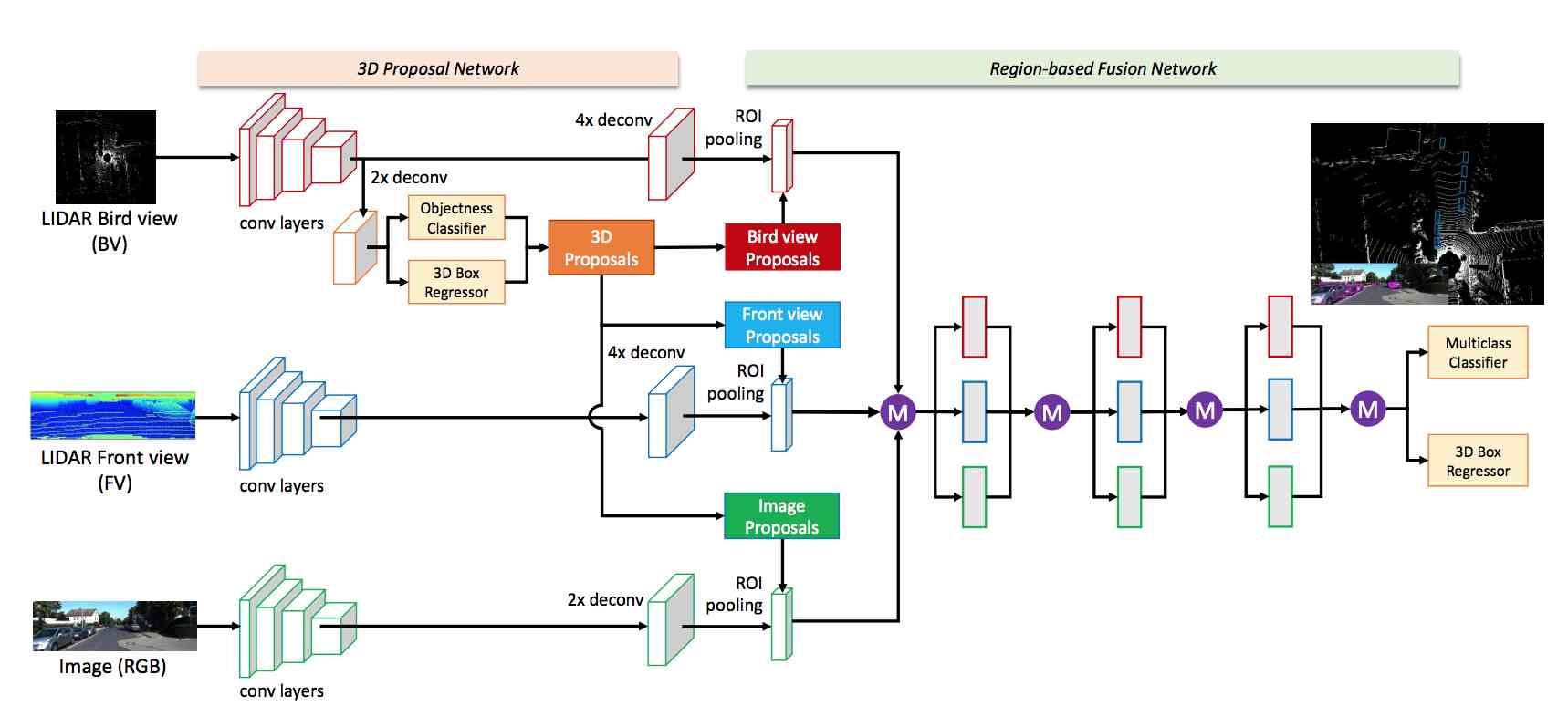}

\end{center}
   \caption{Multi-View 3D object detection network (MV3D)\cite{DBLP:journals/corr/ChenMWLX16} .}
\label{fig:311}
\end{figure}

The two-stage detection framework of the MV3D is shown in Figure \ref{fig:311} including the RPN and detection two stages. The network takes the bird's eye view and front view of LiDAR point cloud as well as an image as input. It first generates 3D object proposals from bird's eye view map and projects them to three views: BEV, FV from LiDAR and Image plane view. A deep fusion network is used to combine region-wise features obtained via ROI pooling for each view. The fused features are used to jointly predict object class and do oriented 3D box regression\cite{DBLP:journals/corr/ChenMWLX16}.\\

\textit{Input features overview for MV3D}\\

\begin{figure}[H]
\begin{center}
%\fbox{\rule{0pt}{2in} \rule{.9\linewidth}{0pt}}
\includegraphics[width=1.0\linewidth]{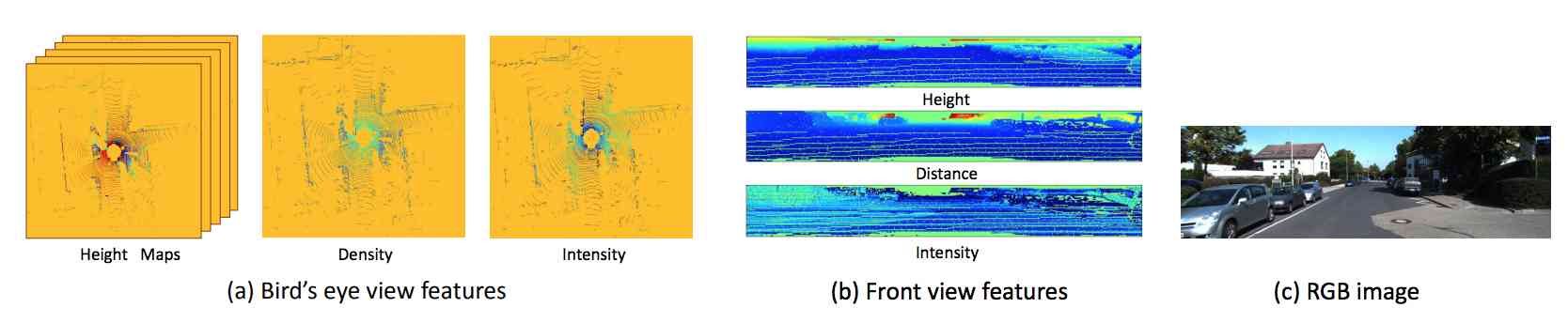}

\end{center}
   \caption{Input features of the MV3D network: BEV, FV and RGB. Figure is from original paper.}
   \label{fig:322}
\end{figure}
MV3D projects 3D point cloud to the bird's eye view and the front view. Figure \ref{fig:322} visualizes the point cloud representation and input image. The detail of the BEV and FV representation is given next.\\

\textit{BEV features  for MV3D}\\

The bird's eye view representation is encoded by height, intensity and density. 
The point cloud is projected into a 2D grid with resolution of $0.1m$. For each cell, the height feature is computed as the maximum height of the points in the cell. To encode more detailed height information, the point cloud is divided equally into $M$ slices. A height map is computed for each slice. The intensity
feature is the reflectance value of the point which has the maximum height in each cell. The point cloud density indicates the number of points in each cell. So it has $(M +2)$ channel features\cite{DBLP:journals/corr/ChenMWLX16} .  MV3D uses point cloud in the range of $[0,70.4] \times [-40, 40]$ meters in the $X$ and $Y$ dimensions. The size of the input features is $704\times800\times(M+2)$. The value of $M$ is not provided in \cite{DBLP:journals/corr/ChenMWLX16}. The length of $Z$ dimension is also not provided. If we suppose that for the length of $Z$ dimension is 2.5 meters as in AVOD\cite{2017arXiv171202294K} and the resolution of the  $Z$ dimension  is also 0.1 meters, then the size of the input feature for the BEV will be $704\times800\times 27$.\\ 

\textit{FV features  for MV3D}\\

MV3D projects the FV into a cylinder plane to generate
a dense front view map as in VeloFCN\cite{DBLP:journals/corr/LiZX16}. The front view map is encoded with three-channel features, which are height, distance and intensity as shown in Figure \ref{fig:322}.  Since KITTI uses a 64-beam Velodyne laser scanner, the size of map for the front view
 is $64 \times 512$.\\
 
  \textit{Performance of MV3D}\\
   
   \begin{figure}[H]
\begin{center}
%\fbox{\rule{0pt}{2in} \rule{.9\linewidth}{0pt}}
\includegraphics[width=1.0\linewidth]{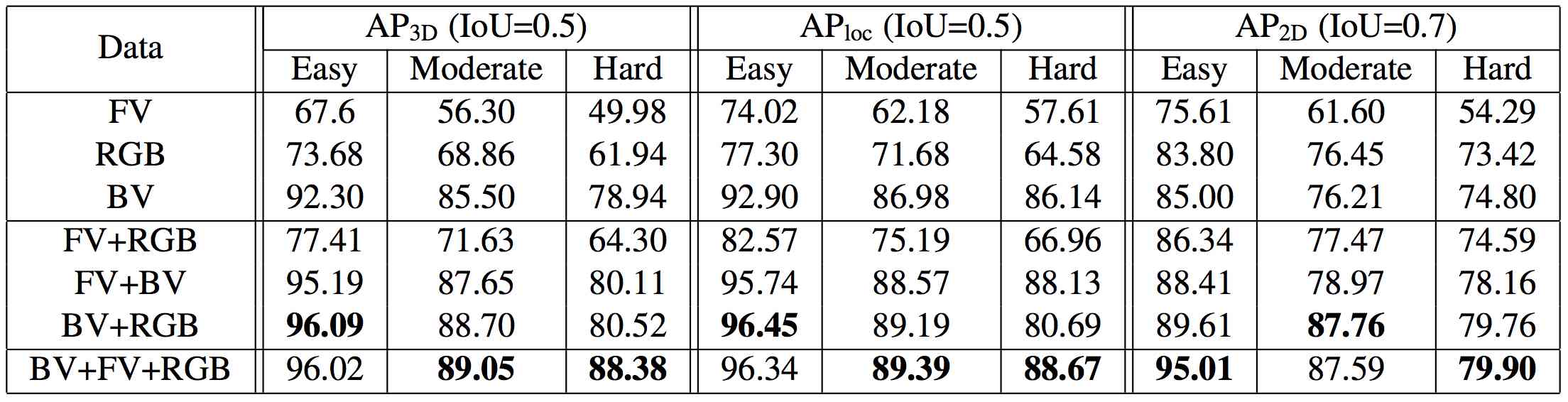}

\end{center}
   \caption{An ablation study of multi-view features: Performance are evaluated on KITTI validation set. Figure is from \cite{DBLP:journals/corr/ChenMWLX16}}
\label{perfromance_input_Fe_mv3d}
\end{figure}

   \begin{figure}[H]
\begin{center}
%\fbox{\rule{0pt}{2in} \rule{.9\linewidth}{0pt}}
\includegraphics[width=1.0\linewidth]{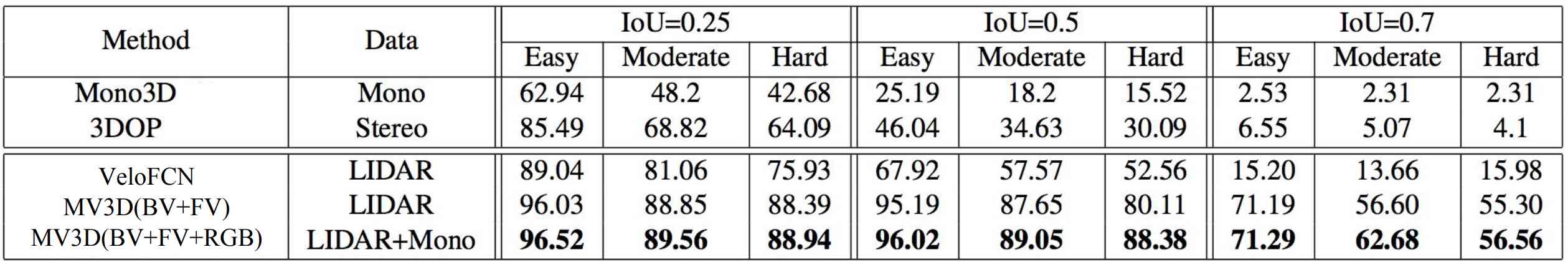}

\end{center}
   \caption{The comparison of the 3D detection performance based on the KITTI validation set. The comparison is provided by \cite{DBLP:journals/corr/ChenMWLX16}}
\label{fig:mv3d_vs_2d}
\end{figure}

  The performance of MV3D is evaluated based on the outdoor KITTI dataset. The performance of 3D object detection based on the test set can be found from the leaderboard. The performance of 3D object detection based on validation dataset is shown in Figures \ref{perfromance_input_Fe_mv3d} and \ref{fig:mv3d_vs_2d}. It only provides the car detection results. Detection results for the pedestrians and cyclists are not provided.\\
  
 \begin{figure}[H]
\begin{center}
%\fbox{\rule{0pt}{2in} \rule{.9\linewidth}{0pt}}
\includegraphics[width=1.0\linewidth]{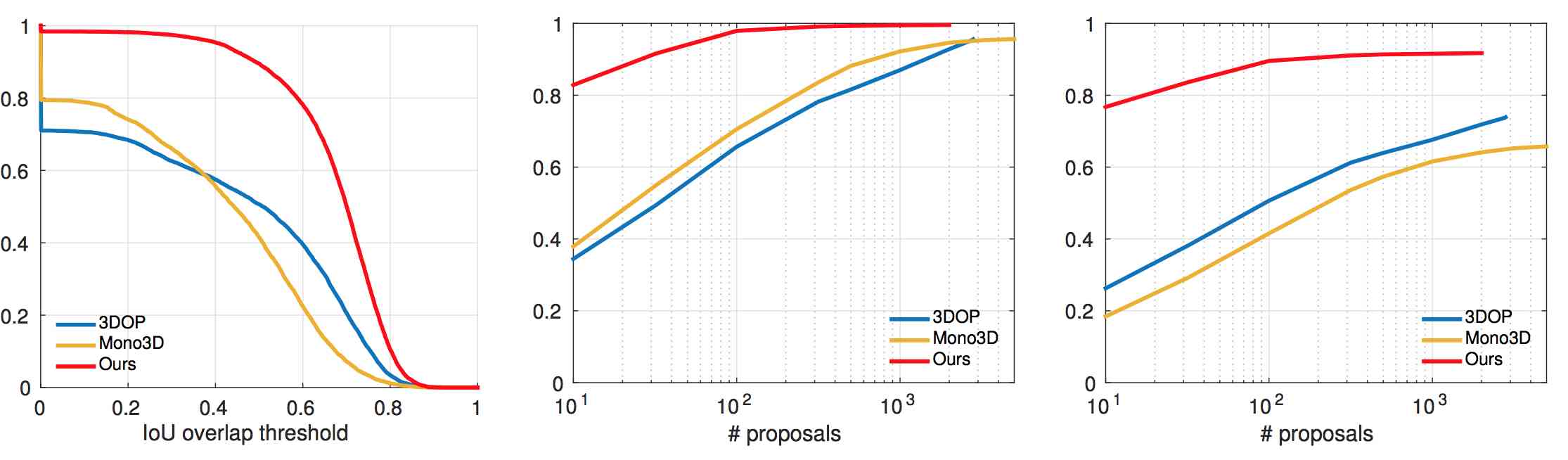}

\end{center}
   \caption{  3D bounding box Recall: From left to right: Recall vs IoU using 300 proposals, Recall vs $\#$Proposals at IoU
threshold of 0.25 and 0.5 respectively. Recall are evaluated on moderate data of KITTI validation set. Figure and Caption are from \cite{DBLP:journals/corr/ChenMWLX16}.
}
  \label{mv3drpn}
\end{figure}

  The performance of the RPN is shown in Figure \ref{mv3drpn}
   
\textit{Framework of AVOD\cite{2017arXiv171202294K}}

%\subsubsection{AVOD\cite{2017arXiv171202294K}}
\begin{figure}[H]
\begin{center}
%\fbox{\rule{0pt}{2in} \rule{.9\linewidth}{0pt}}
\includegraphics[width=1.0\linewidth]{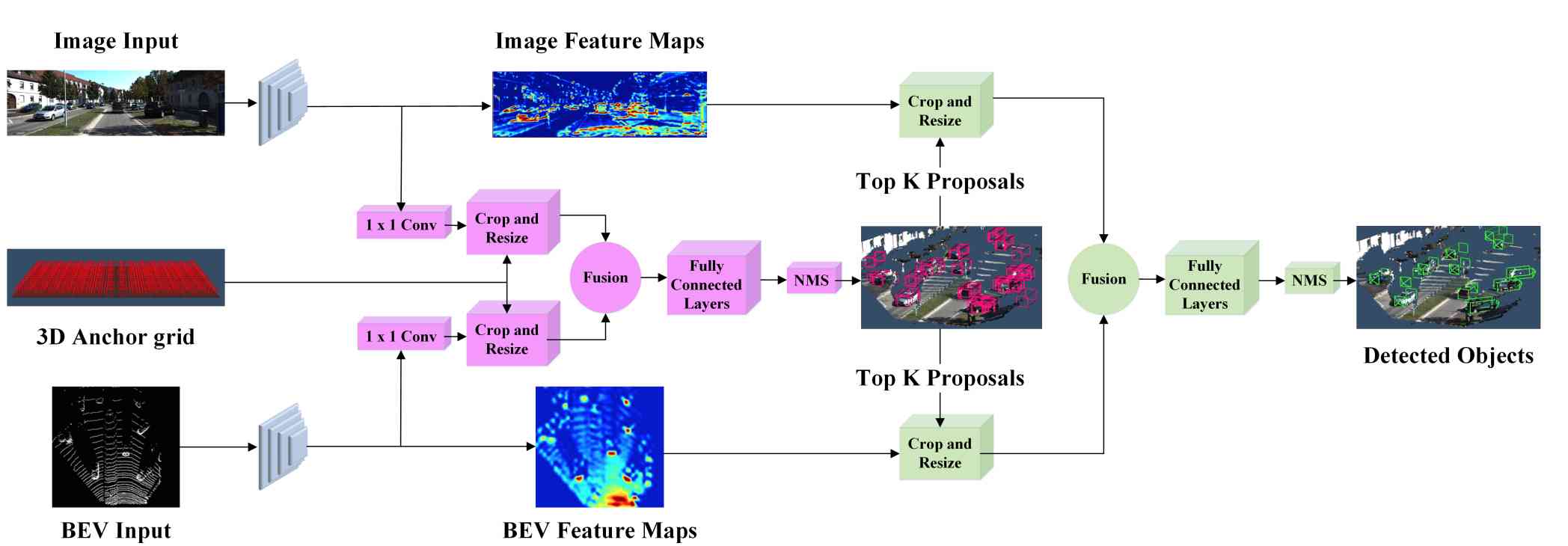}

\end{center}
   \caption{AVOD's architectural diagram. The feature extractors are shown in blue, the region proposal network
in pink, and the second stage detection network in green. Figure and Caption are from \cite{2017arXiv171202294K}.}
\label{fig:211}
\end{figure}

The framework of AVOD is shown in Figure \ref{fig:211}. AVOD is using the same encoding method as MV3D for the BEV. In AVOD, the value of $M$ is set as 5 and the range of the LiDAR is $[0,70] \times [-40, 40] \times [0, 2.5] $ meters. So the size of the input feature for the BEV is $700\times800\times 7$. AVOD is using both the BEV and image to do the region proposals which is the main difference to the MV3D work.\\

\textit{Performance of AVOD\cite{2017arXiv171202294K}}

\begin{figure}[H]
\begin{center}
%\fbox{\rule{0pt}{2in} \rule{.9\linewidth}{0pt}}
\includegraphics[width=1.0\linewidth]{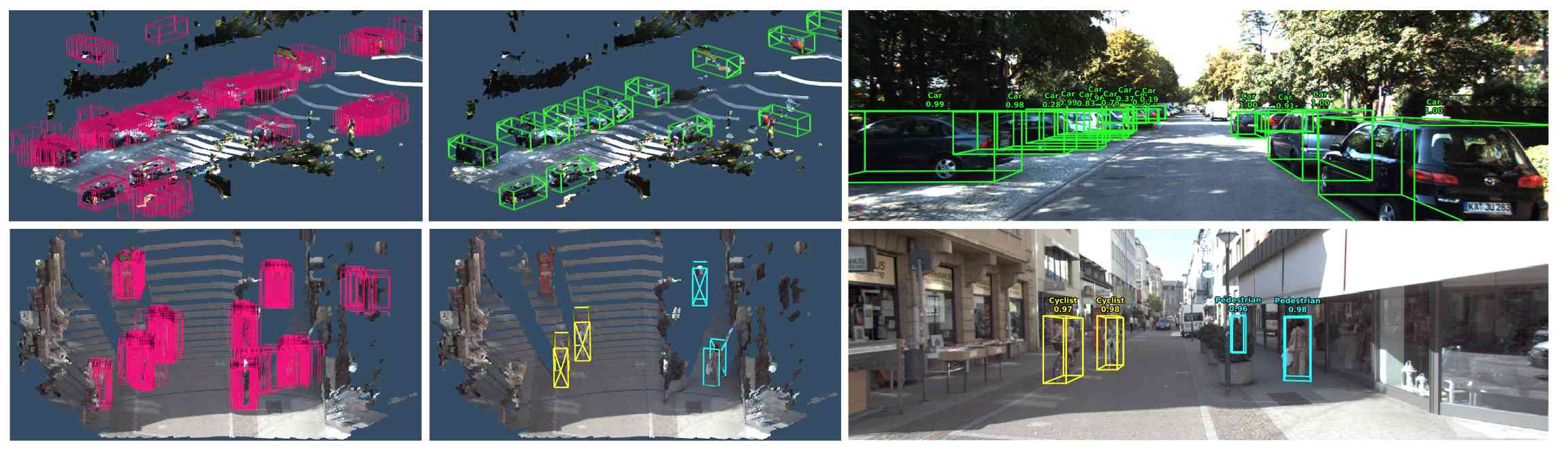}

\end{center}
   \caption{Qualitative results of AVOD\cite{2017arXiv171202294K} for cars (top) and pedestrians/cyclists (bottom). Left: 3D region proposal network output,
Middle: 3D detection output, and Right: the projection of the detection output onto image space for all three classes. The
3D LiDAR point cloud has been colorized and interpolated for better visualization. Figure and Caption are from \cite{2017arXiv171202294K} }
\label{fig:222}
\end{figure}

\begin{figure}[H]
\begin{center}
%\fbox{\rule{0pt}{2in} \rule{.9\linewidth}{0pt}}
\includegraphics[width=1.0\linewidth]{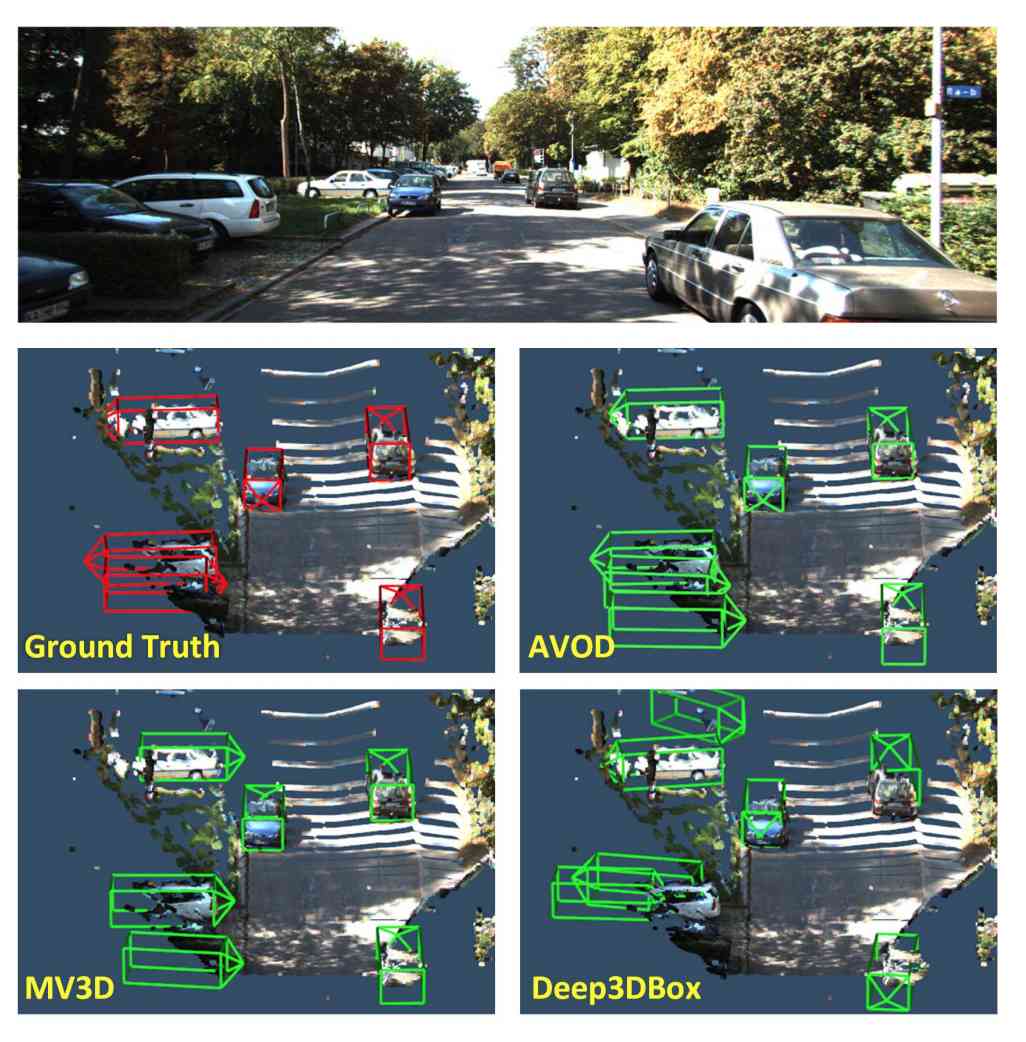}

\end{center}
   \caption{A qualitative comparison between MV3D,
Deep3DBox\cite{DBLP:journals/corr/MousavianAFK16}, and AVOD architecture relative to KITTI's
ground truth on a sample in the validation set. Figure is from \cite{2017arXiv171202294K}. We did not cover Deep3DBox in this survey.}
\label{fig:233}
\end{figure}

The comparison of the AVOD and other methods is shown in the leaderboard. Some visualization results are shown in Figures \ref{fig:222} and \ref{fig:233}.\\

\textbf{VoxelNet\cite{DBLP:journals/corr/abs-1711-06396}}

\textit{Architecture of VoxelNet\cite{DBLP:journals/corr/abs-1711-06396}}\\

%\begin{figure}[H]
%\begin{center}
%\fbox{\rule{0pt}{2in} \rule{.9\linewidth}{0pt}}
%\includegraphics[width=1.0\linewidth]{9911}

%\end{center}
 %  \caption{VoxelNet directly operates on the raw point cloud (no need for feature engineering) and produces the 3D detection results using a single end-to-end trainable network.}
%\label{fig:9911}
%\end{figure}

\begin{figure}[H]
\begin{center}
%\fbox{\rule{0pt}{2in} \rule{.9\linewidth}{0pt}}
\includegraphics[width=1.0\linewidth]{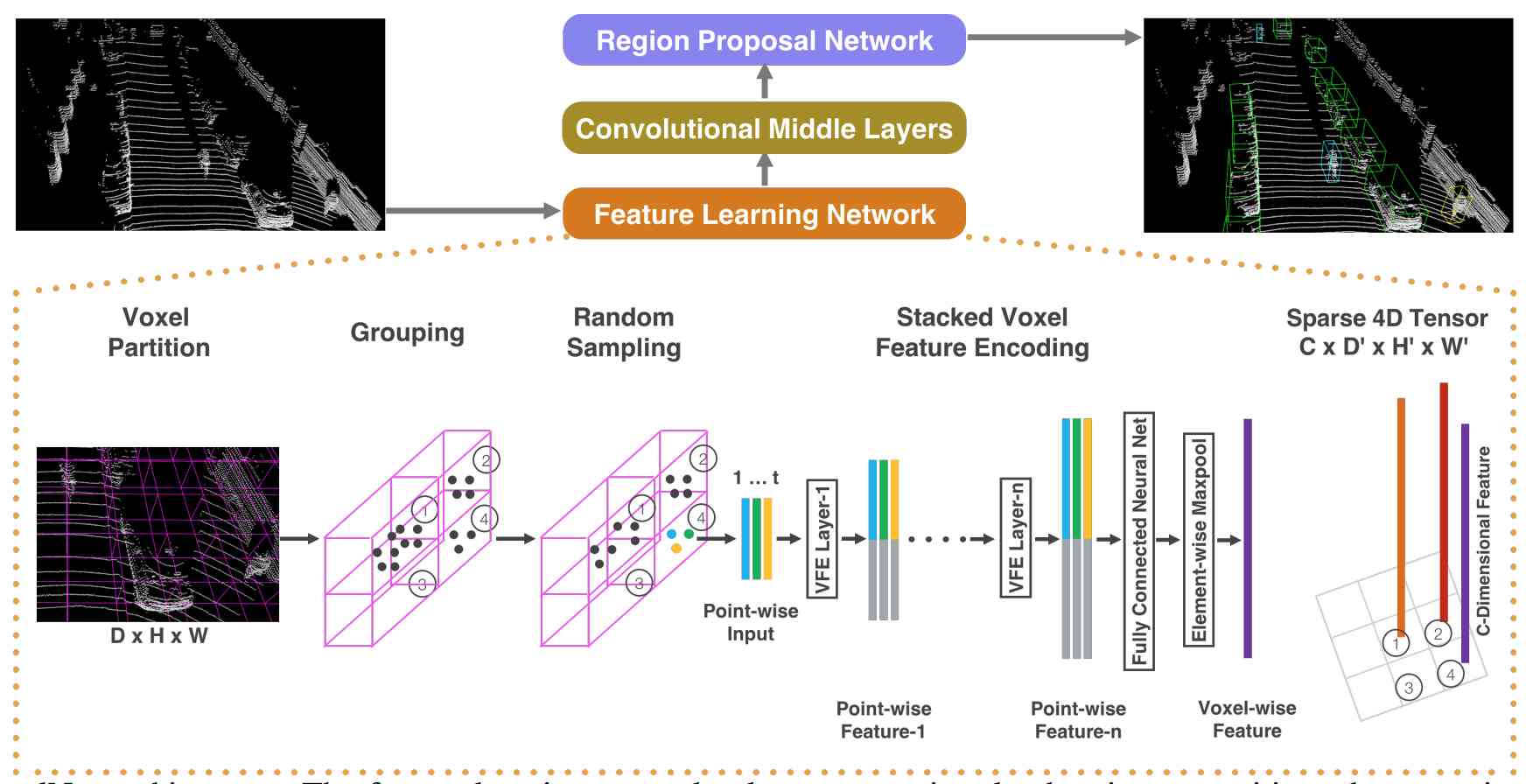}

\end{center}
   \caption{VoxelNet\cite{DBLP:journals/corr/abs-1711-06396} architecture. Figure is from original paper.}
\label{fig:9922}
\end{figure}

VoxelNet architecture is shown in Figure \ref{fig:9922}. The feature learning network takes a raw point cloud as input, partitions the space into voxels, and
transforms points within each voxel to a vector representation characterizing the shape information. The space is represented as a sparse
4D tensor. The convolutional middle layers processes the 4D tensor to aggregate spatial context. Finally, a RPN generates the 3D detection.\\

\textit{Feature Learning Network}\\

\begin{figure}[H]
\begin{center}
%\fbox{\rule{0pt}{2in} \rule{.9\linewidth}{0pt}}
\includegraphics[width=1.0\linewidth]{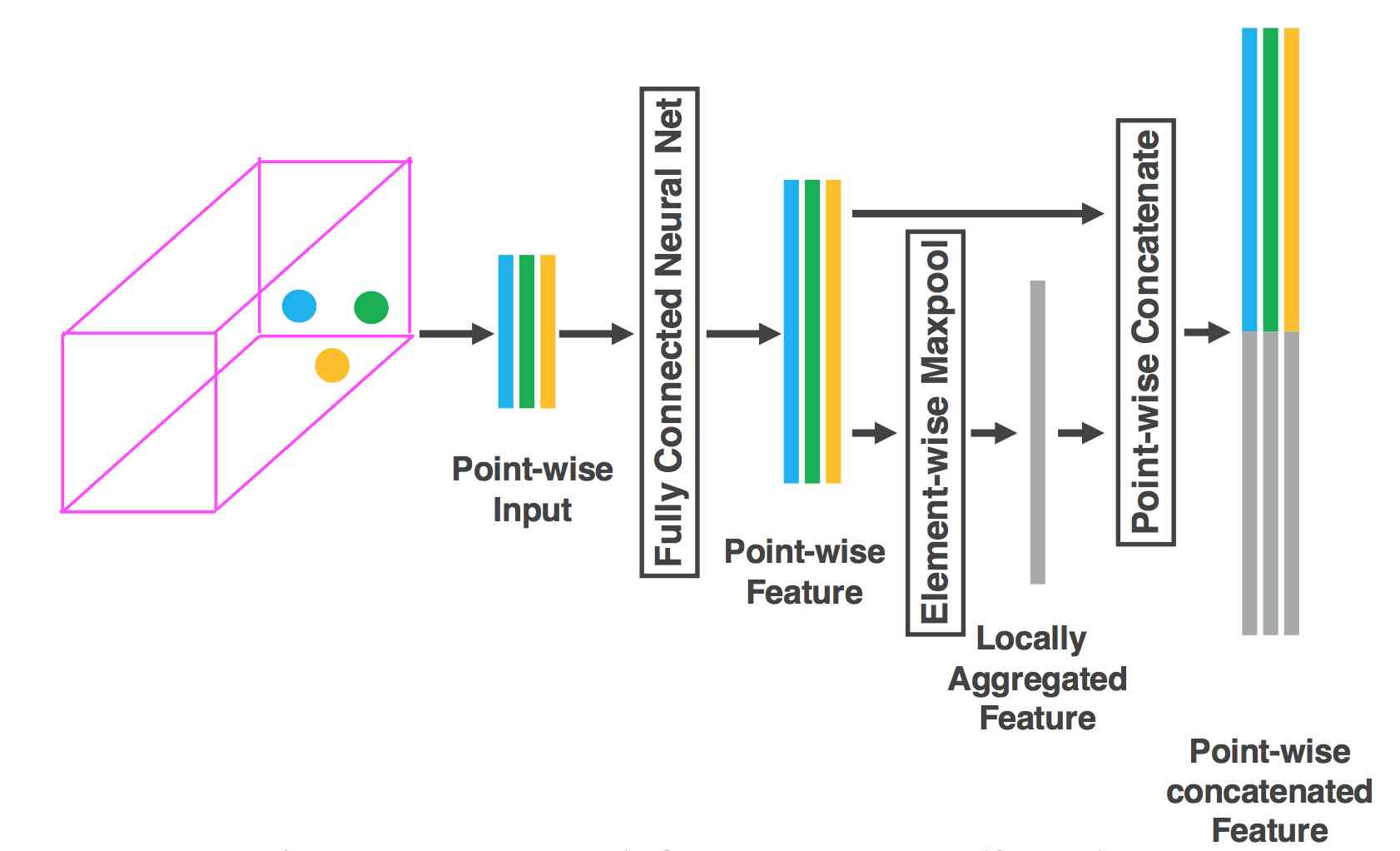}

\end{center}
   \caption{Voxel feature encoding layer. Figure is from \cite{DBLP:journals/corr/abs-1711-06396}}
\label{fig:9933}
\end{figure}

The point cloud is subdivided into equally spaced voxels with size $D^ {\prime}, H^ {\prime}, W^ {\prime}$ along the $Z, Y, X$ axes respectively. The $D^ {\prime}, H^ {\prime}$ and $W^ {\prime}$ for car detection is set as $D^ {\prime} = 10, H^ {\prime} = 400, W^ {\prime} = 352$ with the consideration of the range of LiDAR data as $[-3, 1] \times [-40, 40] \times [0, 70.4]$ meters and resolution of $0.4 \times 0.2  \times0.2$ meters along $Z, Y, X$ axis respectively.\\

A fixed number, $T$, of points from voxels containing more than $T$ points are randomly sampled. For each point, a 7-feature is used which is $(x, y, z,r, x-v_x, y-v_y, z-v_z)$ where $x, y, z$ are the $XYZ$ coordinates for each point.  $r$
is the received reflectance and $(v_x, v_y, v_z)$ is the centroid of points in the voxel.\\
Voxel Feature Encoding is proposed in VoxelNet. The 7-feature for each point is fed into the Voxel feature encoding layer as shown in Figure \ref{fig:9933}. Fully connected networks are used in the VFE network with element-wise MaxPooling for each point and concatenation between each point and the element-wise MaxPooling output.  The input of the VFE is $T\times 7$ and the output will be $T \times C$ where $C$ depends on the FC layers of the VFE itself and depends on the whole VFE layers network used. Finally, an element-wise MaxPooling is used again and change the dimension of the output to $1\times C$. Then for each voxel we have a one vector with $C$ elements as shown in Figure \ref{fig:9922}. For the whole framework, we will have an input data with shape of $C \times D^ {\prime} \times  H^ {\prime} \times W^ {\prime}$.\\

\textit{Convolutional Middle Layers}\\

For car detection,  $T = 35, C= 128$. A 3D CNN is used to further extract features. The input of this 3D CNN is $C \times D^ {\prime} \times  H^ {\prime} \times W^ {\prime} = 128 \times 10 \times 400 \times 352$. The output is $64  \times  2  \times  400  \times  352$. After reshaping, the input
to RPN is a feature map of size $128  \times  400  \times  352$.\\

\textit{Region Proposal Network}\\

\begin{figure}[H]
\begin{center}
%\fbox{\rule{0pt}{2in} \rule{.9\linewidth}{0pt}}
\includegraphics[width=1.0\linewidth]{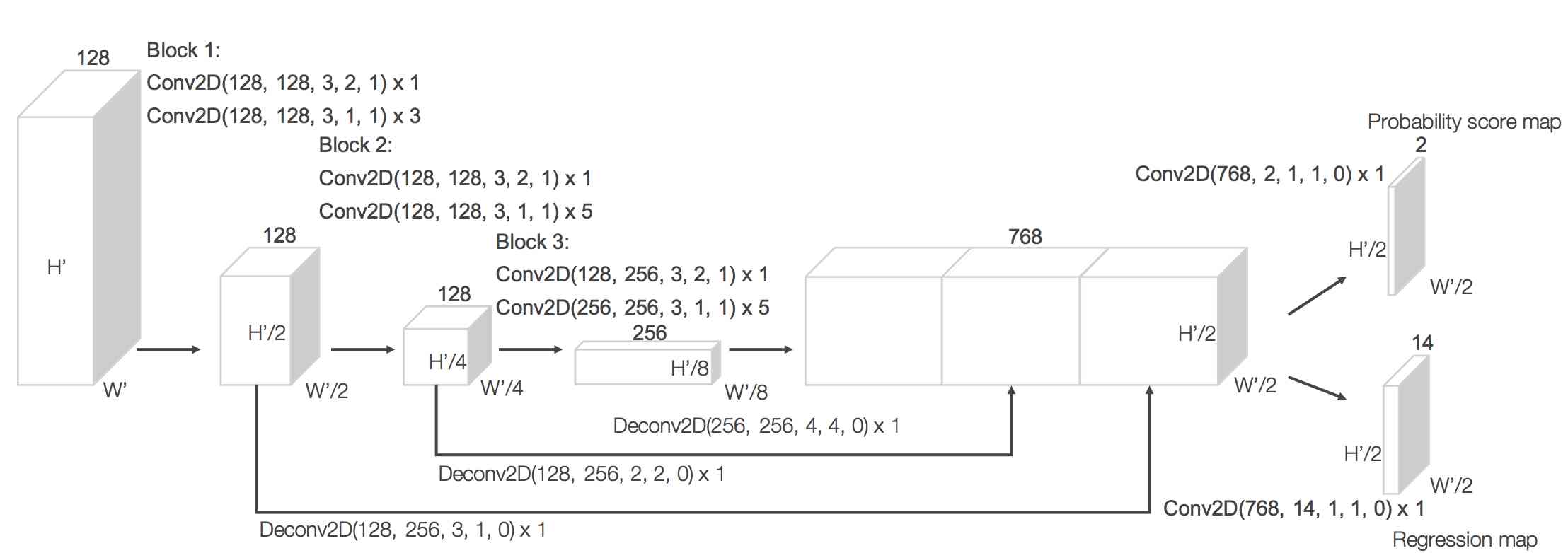}

\end{center}
   \caption{Region proposal network architecture. Figure is from \cite{DBLP:journals/corr/abs-1711-06396}.}
\label{fig:9944}
\end{figure}

Only one anchor size, $length = 3.9, width = 1.6, height = 1.56$ meters is used. It centered at $z = -1.0$ meters with two rotations, 0 and 90 degrees.
The input to RPN is the feature map provided by the convolutional middle layers. The architecture of this network is illustrated in Figure \ref{fig:9944}.\\

\textit{Performance of VoxelNet}\\

The comparison of the VoxelNet and other methods is shown in the leaderboard. 
\subsubsection{Outdoor $\&$ Indoor}
Some 3D object detection systems can work well for both the indoor and outdoor scenarios are introduced in this part.\\

\textbf{F-PointNet\cite{DBLP:journals/corr/abs-1711-08488}	}\\

\textit{Framework of F-PointNet}\\

\begin{figure}[H]
\begin{center}
%\fbox{\rule{0pt}{2in} \rule{.9\linewidth}{0pt}}
\includegraphics[width=1.0\linewidth]{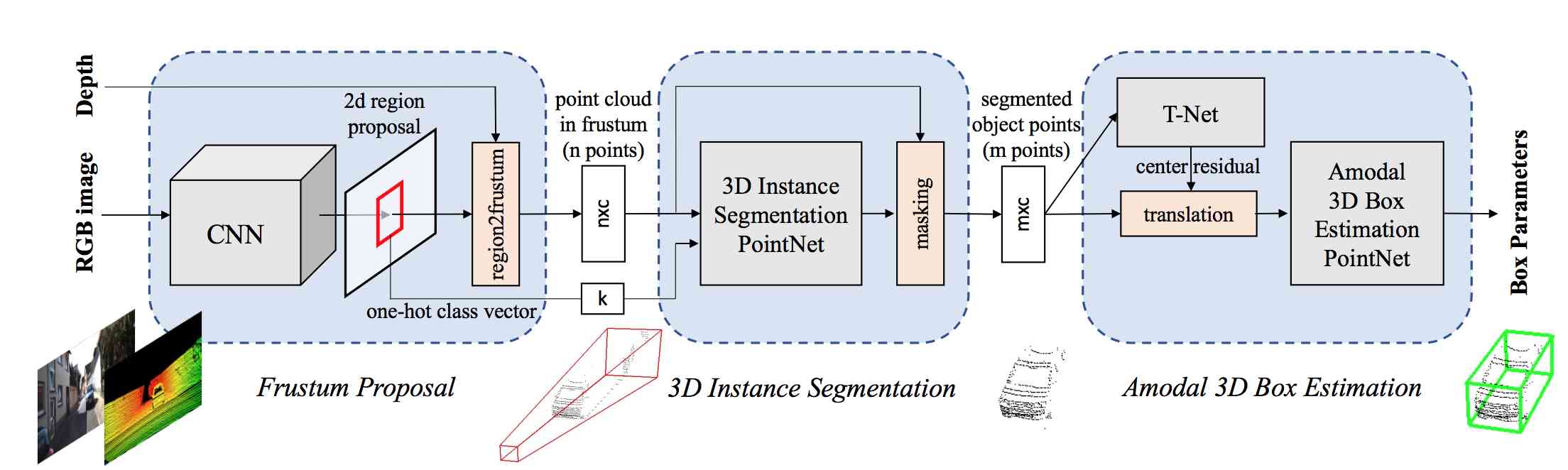}

\end{center}
   \caption{Frustum PointNets\cite{DBLP:journals/corr/abs-1711-08488} for 3D object detection. Figure is from original paper. }
\label{fig:111}
\end{figure}
The framework of the Frustum PointNets is given in Figure \ref{fig:111}. It first leverages a 2D CNN object detector to propose 2D regions and classify
their content. 2D regions are then lifted to 3D and thus become frustum proposals. Given a point cloud in a frustum ($n \times c$ with n points
and c channels of XYZ, intensity etc. for each point), the object instance is segmented by binary classification of each point. Based on the
segmented object point cloud ($m\times c$), a light-weight regression PointNet (T-Net) tries to align points by translation such that their centroid
is close to amodal box center. At last the box estimation net estimates the amodal 3D bounding box for the object.\\

\textit{Performance of F-PointNet}\\

\begin{figure}[H]
\begin{center}
%\fbox{\rule{0pt}{2in} \rule{.9\linewidth}{0pt}}
\includegraphics[width=1.0\linewidth]{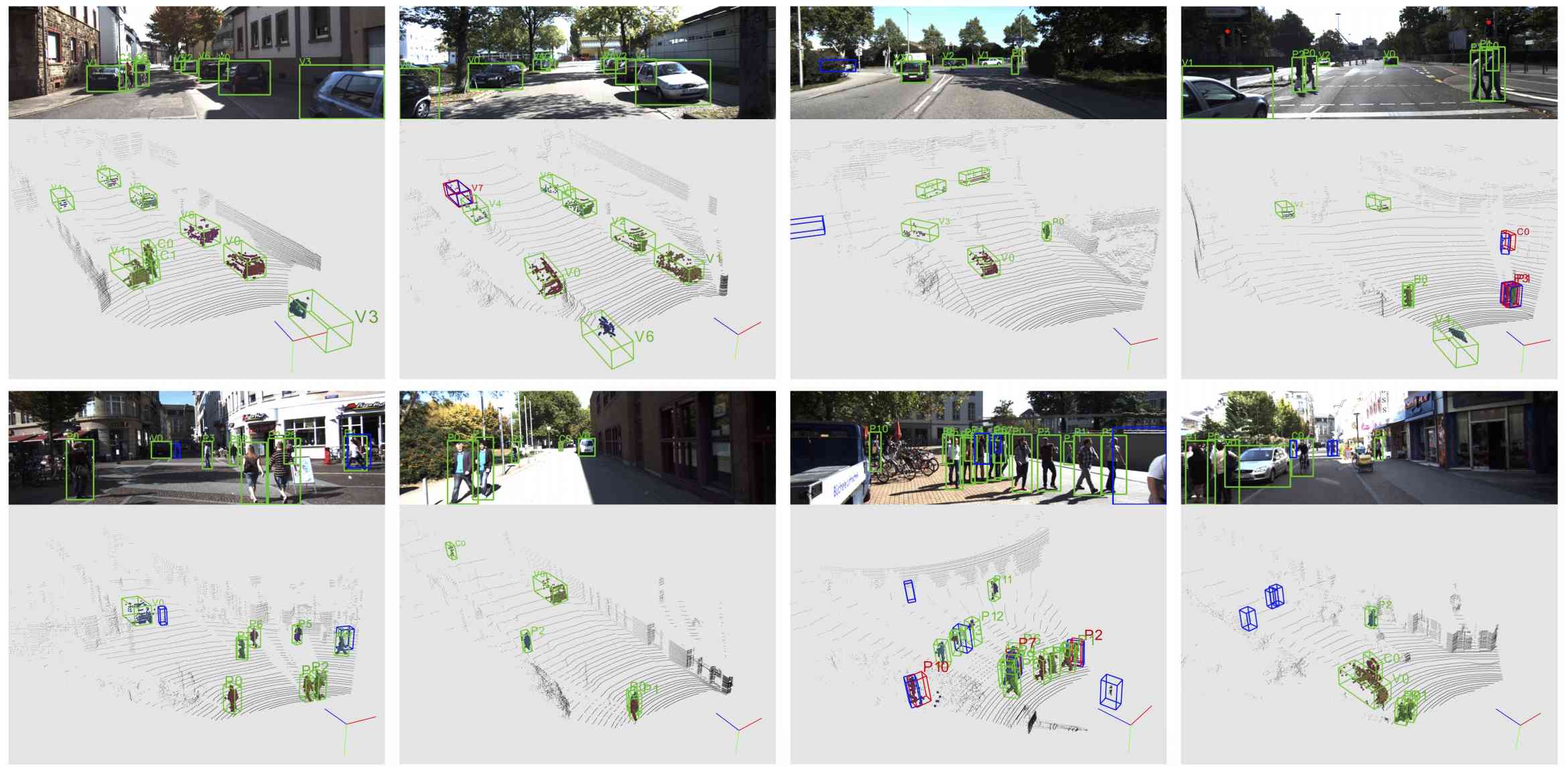}

\end{center}
   \caption{Visualizations of Frustum PointNet results on KITTI val set (best viewed in color with zoom in). These results are based
on PointNet++\cite{DBLP:journals/corr/QiYSG17} models , running at 5 fps and achieving test set 3D AP of 70.39, 44.89 and 56.77 for car, pedestrian and cyclist,
respectively. 3D instance masks on point cloud are shown in color. True positive detection boxes are in green, while false positive boxes
are in red and ground truth boxes in blue are shown for false positive and false negative cases. Digit and letter beside each box denote
instance id and semantic class, with $v$ for cars, $p$ for pedestrian and $c$ for cyclist. Figure and Caption are from \cite{DBLP:journals/corr/abs-1711-08488}.}
\label{fig:122}
\end{figure}

The comparison of the F-PointNet and other methods is shown in the leaderboard.  Some detection results are demonstrated in Figure \ref{fig:122}.\\

\subsection{Data representation methods summary for 3D system}
From the surveyed 3D systems, we can see the importance of the data representation to the performance. Here the pros and cons of different data representation methods to the classification and detection tasks is summarized:\\
\begin{itemize}
\item using \textbf{Projected multiple view RGB images}
\begin{itemize}
\item \textbf{Pros} : It is a similar to the human being's recognization process for a 3D object by looking from different views. Since it can encode the multiple view info into 2D RGB image, it can take the advantage of well developed 2D image recognization system such as 2D CNN to further classify or detect 3D objects.
\item \textbf{Cons} : Sometime, not all the desired multiple-view RGB images are available. In some papers which use multiple views such as RotationNet\cite{DBLP:journals/corr/QiSMG16} and MVCNN\cite{su15mvcnn}, views from different angles are used to do the classification. This is possible when the whole object's CAD model is available. However, in the real application such as the autonomous cars scenario, from the self-driving cars' perspective, it can only take multiple-view of objects from one side during the driving process. The other side of the objects cannot be observed due to self-occultations. The partial availability of all views will reduce the performance of the algorithms based on all views.
\end {itemize}
\item using \textbf{voxel}
\begin{itemize}
\item \textbf{Pros}:  It is a natural way to represent a 3D shape into voxel.
\item \textbf{Cons}: Corresponding to voxel representation, 3D CNN is commonly used based on this representation. The computation complexity of $\mathcal{O}(n^3)$ makes the system which uses 3D CNN based on voxel representation can only afford low resolution voxels. Low resolution will decrease the performance as not all the shape information will be preserved. Furthermore, the voxel representation will suffer from occultations.
\end{itemize}
\item using \textbf{projected 2D similar images from depth image}
\begin{itemize}
\item \textbf{Pros}:  By projecting depth image to 2D similar images such as the 3 channels including the $x$, $y$ and $z$ values collecting from depth image can take the benefit of well developed 2D image system by using the state of the art technology such as 2D CNN.
\item \textbf{Cons}: Projecting depth info to 2D similar image will lose the geometry information and further decrease the classification or detection performance.
\end{itemize}
\item using \textbf{raw point cloud}
\begin{itemize}
\item \textbf{Pros}:  properly using the point cloud info can well preserve the 3D geometry information. At the same time, the complexity of using point cloud is $\mathcal{O}(n^2)$ which is less expensive than using the voxel.
\item \textbf{Cons}: The techniques of using point cloud is under development. Same to the voxel representation, point cloud will suffer from occultations.
\end{itemize}
\end{itemize}

\section{Conclusion}
In this survey, the main works in the 2D/3D image based object classification/detection system are introduced to help the researcher have an understanding about this area. Comprehensive comparisons are given to different methods based on the RPN network, detection network, feature encoding methods, bounding box encoding methods and also the application scenarios. Also the performance of different methods are compared based on some commonly used datasets.\\
 
%Multi-View 3D Object Detection Network for Autonomous Driving

%\cite{Ge_2017_CVPR} 3D Convolutional Neural Networks for Efficient and Robust Hand Pose Estimation From Single Depth Images

%\cite{Geiger2013IJRR} KITTI 

%Deep MANTA: A Coarse-To-Fine Many-Task Network for Joint 2D and 3D Vehicle Analysis From Monocular Image

% FCN

%DeepLab: Semantic Image Segmentation with Deep Convolutional Nets, Atrous Convolution, and Fully Connected CRFs

%\cite{DBLP:journals/corr/ChenPK0Y16}

%\cite{DBLP:journals/corr/LiQDJW16} FCIS

%Mask R-CNN

%\cite{DBLP:journals/corr/HeGDG17}

 %%%%%%%%%%%%%%%%%%%%%%%%%%
 
% deep mask

%\cite{DBLP:journals/corr/PinheiroCD15}

%crf RNN \cite{DBLP:journals/corr/ZhengJRVSDHT15}

%hypper column
%\cite{DBLP:journals/corr/HariharanAGM14a}\cite{DBLP:journals/corr/PinheiroCD15}

\bibliography{jimmy_shen}
\bibliographystyle{ieeetr}

 \end{document}